\documentclass{article}

\usepackage{microtype}
\usepackage{booktabs} 

\usepackage{hyperref}
\usepackage{url}
\usepackage{multirow}
\usepackage{subcaption}

\usepackage[pdftex]{graphicx} 
\DeclareGraphicsExtensions{.pdf,.jpeg,.png,.jpg}
\usepackage[pdftex]{color}
\usepackage{threeparttable}
\usepackage{here}

\usepackage{bm}

\usepackage{amsmath,amsfonts,bm}

\def\eqref#1{equation~\ref{#1}}

\def\1{\bm{1}}

\def\met<#1>{\langle #1 \rangle}

\newcommand{\dif}{\mathrm{d}}

\def\vb{{\bm{b}}}

\def\ve{{\bm{e}}}

\def\vp{{\bm{p}}}
\def\vq{{\bm{q}}}

\def\vs{{\bm{s}}}

\def\vu{{\bm{u}}}
\def\vv{{\bm{v}}}
\def\vw{{\bm{w}}}
\def\vx{{\bm{x}}}
\def\vy{{\bm{y}}}
\def\vz{{\bm{z}}}

\def\mA{{\bm{A}}}
\def\mB{{\bm{B}}}
\def\mC{{\bm{C}}}
\def\mD{{\bm{D}}}

\def\mG{{\bm{G}}}

\def\mI{{\bm{I}}}
\def\mJ{{\bm{J}}}

\def\mO{{\bm{O}}}

\def\mS{{\bm{S}}}

\def\mU{{\bm{U}}}
\def\mV{{\bm{V}}}
\def\mW{{\bm{W}}}

\DeclareMathAlphabet{\mathsfit}{\encodingdefault}{\sfdefault}{m}{sl}
\SetMathAlphabet{\mathsfit}{bold}{\encodingdefault}{\sfdefault}{bx}{n}

\def\gM{{\mathcal{M}}}
\def\gN{{\mathcal{N}}}

\def\sR{{\mathbb{R}}}

\newcommand{\R}{\mathbb{R}}

\newcommand{\argmin}{\mathop{\rm arg~min}\limits}

\usepackage{multicol}
\newcommand{\mblk}{\color{black}}
\newcommand{\mmmag}{\color{black}}
\newcommand{\mmred}{\color{black}}
\newcommand{\mmblu}{\color{black}}

\newcommand{\mmmblu}{\color{black}}
\newcommand{\mred}{}
\newcommand{\mblu}{}
\newcommand{\mgre}{}
\newcommand{\mmag}{}

\usepackage{hyperref}

\usepackage{mathrsfs}
\usepackage{mathtools,xparse}
\usepackage{amsmath,xparse}

\makeatletter
\newcommand{\absdetaux}[3]{\mathpalette\absdetaux@i{{#1}{#2}{#3}}}
\newcommand{\absdetaux@i}[2]{\absdetaux@ii#1#2}
\newcommand{\absdetaux@ii}[4]{%
  \sbox\z@{$\m@th#1#2#4#3$}%
  \sbox\tw@{$\m@th\|$}%
  \mathopen{\hbox to\wd\tw@{\hss\vrule height \ht\z@ depth \dp\z@ width .3\wd\tw@\hss}}%
  #4
  \mathclose{\hbox to\wd\tw@{\hss\vrule height \ht\z@ depth \dp\z@ width .3\wd\tw@\hss}}%
}
\makeatother

\usepackage{theorem}  
{\theorembodyfont{\upshape}  
} 

\usepackage[accepted]{icml2020}

\icmltitlerunning{Rate-Distortion Optimization Guided Autoencoder for Isometric Embedding in Euclidean Latent Space}

\begin{document}

\twocolumn[
\icmltitle{Rate-Distortion Optimization Guided Autoencoder for Isometric Embedding in Euclidean Latent Space}




\begin{icmlauthorlist}
\icmlauthor{Keizo Kato}{jp}
\icmlauthor{Jing Zhou}{ch}
\icmlauthor{Tomotake Sasaki}{jp}
\icmlauthor{Akira Nakagawa}{jp}
\end{icmlauthorlist}

\icmlaffiliation{jp}{FUJITSU LABORATORIES LTD.}
\icmlaffiliation{ch}{Fujitsu R\&D Center Co., Ltd.}

\icmlcorrespondingauthor{kato.keizo, anaka}{@fujitsu.com}
\icmlkeywords{Machine Learning, ICML}

\vskip 0.3in
]



\printAffiliationsAndNotice{}  

\begin{abstract}
To analyze high-dimensional and complex data in the real world, deep generative models, such as variational autoencoder (VAE) embed data in a low-dimensional space (latent space) and learn a probabilistic model in the latent space. However, they struggle to accurately reproduce the probability distribution function (PDF) in the input space from that in the latent space. If the embedding were isometric, this issue can be solved, because the relation of PDFs can become tractable. To achieve isometric property, we propose Rate-Distortion Optimization guided autoencoder inspired by orthonormal transform coding. We show our method has the following properties: (i) the Jacobian matrix between the input space and a Euclidean latent space forms a constantly-scaled orthonormal system and enables isometric data embedding; (ii) the relation of PDFs in both spaces can become tractable one such as proportional relation. Furthermore, our method outperforms state-of-the-art methods in unsupervised anomaly detection with four public datasets. 
\end{abstract}
\section{Introduction}
Capturing the inherent features of a dataset from high-dimensional and complex data is an essential issue in machine learning. Generative model approach learns the probability distribution of data, aiming at data generation, unsupervised learning, disentanglement,  etc.~\cite{AESurvey}. 
It is generally difficult to directly estimate a probability density function (PDF) $P_{\vx}(\vx)$ of high-dimensional data $\vx \in \sR^{M}$. 
Instead, one promising approach is to 
map $\vx$ to a low-dimensional latent variable $\vz \in \sR^{N}$ ($N < M$), and capture PDF $P_{\vz}(\vz)$. 
Variational autoencoder (VAE) is a widely used generative model to capture $\vz$ as a probabilistic model with univariate Gaussian priors \citep{VAE}. 
For a more flexible estimation of $P_{\vz}(\vz)$, successor models have been proposed, such as using Gaussian mixture model (GMM)~\cite{DAGMM}, combining univariate Gaussian model and GMM~\citep{GMVAE}, etc. 


In tasks where the quantitative analysis is vital, $P_{\vx}(\vx)$ should be reproduced from $P_{\vz}(\vz)$. For instance, in anomaly detection, the anomaly likelihood is calculated based on PDF value of data sample~\cite{anomaly_survey}. 
However, the embedding of VAEs is not isometric; that is, the distance between data points $\vx^{(1)}$ and $\vx^{(2)}$ is inconsistent to the distance of corresponding latent variables $\vz^{(1)}$ and $\vz^{(2)}$~\cite{geo1, geo2, geo3}. 
Obviously mere estimation of  $P_{\vz}(\vz)$ cannot be the substitution of the estimation for $P_{\vx}(\vx)$ under such situation. 
As \citet{nearly} mentioned, for a reliable data analysis, the isometric embedding in low-dimensional space is necessary. 
In addition, to utilize the standard PDF estimation techniques, the latent space is preferred to be a Euclidean space. 
Despite of its importance, 
this point is not considered even in methods developed for the quantitative analysis of PDF \cite{SAVE,DAGMM, GMVAE, ALAD, inclusive}. 

According to the Nash embedding theorem, an arbitrary smooth and compact Riemannian manifold $\gM $ can be embedded in a Euclidean space $\sR^{N}$ ($N \geq \dim \gM +1$, sufficiently large) isometrically \citep{NashThm}. 
On the other hand, the manifold hypothesis argues that real-world data presented in a high-dimensional space concentrate in the vicinity of a much lower dimensional manifold $\gM_{\vx} \subset \sR^{M}$ \citep{mani}. 
Based on these theories, it is expected that the input data $\vx$ can be embedded isometrically in a low-dimensional Euclidean space $\sR^{N}$ when $ \dim \gM_{ \vx } < N \ll M$. 
Although the existence of the isometric embedding was proven, the method to achieve it has been challenging. Some previous works have proposed algorithms to do that \citep{nearly,bernstein2000graph}. Yet, they do not deal with high-dimensional input data, such as images. 
Another thing to consider is the distance on $\gM_{\vx}$ may be defined by the data tendency with an appropriate metric function. For instance, we can choose the binary cross entropy (BCE) for 
binary data
and structured similarity (SSIM) for image.
As a whole, our challenge is to develop a deep generative model that guarantees the isometric embedding even for the high-dimensional data observed around $\gM_{\vx}$ endowed with a variety of metric function.

Mathematically, the condition of isometric embedding to Euclidean space is equivalent to that the columns of the Jacobian matrix between two spaces form an orthonormal system. 
When we turn our sight to conventional image compression area, 
orthonormal transform is necessary for an efficient compression. This is proven by rate-distortion (RD) theory~\citep{RDTheory} . 
Furthermore, the empirical method for optimal compression with orthonormal transform coding is established as rate-distortion optimization (RDO)~\citep{RDOVideo}. 
It is intuitive to regard data embedding to a low-dimensional latent space as an analogy of efficient data compression. 
Actually, deep learning based image compression (DIC) methods with RDO \citep{Balle, Zhou} have been proposed and they have achieved good compression performance. 
Although it is not discussed in \citet{Balle, Zhou}, we guess that behind the success of DIC, there should be theoretical relation to RDO of convetional transform coding. 

Hence, in this study, we investigate the theoretical property and dig out the proof that RDO guides deep-autoencoders to have the orthonormal property. 
Based on this \mred finding\mblk, we propose a method that enables isometric data embedding and allows a comprehensive data analysis, named Rate-Distortion Optimization Guided Autoencoder for Generative Analysis (RaDOGAGA). 
We show the validity of RaDOGAGA in the following steps. 

(1) We show that RaDOGAGA has the following properties both theoretically and experimentally.
\vspace{-0.5\baselineskip}
\begin{itemize}
   \setlength{\leftskip}{-0.4cm}
	\setlength{\itemsep}{3pt}
	\setlength{\parskip}{0pt}      
	\setlength{\itemindent}{0pt}   
	\setlength{\labelsep}{3pt}     
      \item The Jacobian matrix between the 
      data observation space 
      (inner product space endowed with a metric tensor) 
      and latent space forms a constantly-scaled orthonormal system. Thus, data can be embedded in a Euclidean latent space isometrically. 
      \item Thanks to the property above, the relation of $P_{\vz}(\vz)$ and $P_{\vx}(\vx)$ can become 
      tractable one (e.g., proportional relation).
      Thus, PDF of $\vx$ in the data observation space can be estimated by maximizing log-likelihood of parametric PDF $P_{\vz, \psi}(\vz)$ in the low-dimensional Euclidean space.
      \end{itemize}
\vspace{-0.5\baselineskip}
(2) Thanks to (1), RaDOGAGA outperforms the current state-of-the-art method in unsupervised anomaly detection task with four public datasets.
\noindent
{\bf Isometric Map and Notions of Differential Geometry}

Here, we explain notions of differential geometry 
adopted to our context.
Given two Riemannian manifolds 
$\gM \subset \sR^{M}$ and $\gN \subset \sR^{N}$,  
a map $g:\gM \rightarrow \gN$ is called isometric if  
\begin{equation}
\met< \vv , \vw >_{\vp} = \met< \dif g(\vv), \dif g(\vw) >_{g(\vp)}
\label{def_iso}
\end{equation}
holds. Here, $\vv$ and $\vw$ are tangent vectors in $T_{\vp}{\gM}$ (tangent space of $\gM$ at $\vp \in \gM$) represented as elements of $\sR^{M}$ and $\dif g$ is the differential of $g$ (this can be written as a Jacobian matrix). $\met< \vv, \vw >_{\vp} = \vv^\top \mA_{\gM}(\vp) \vw$, where $\mA_{\gM}(\vp) \in  \sR^{M \times M}$ is a metric tensor 
represented as a positive define matrix. 
The inner product in the right side is also defined by another metric tensor $\mA_{\gN}(\vq) \in \sR^{N \times N}$. 
$\mA_{\gM}(\vp)$ or $\mA_{\gN}(\vq)$ is an identity matrix for a Euclidean case and the inner product becomes the standard one (the dot product). 

We slightly abuse the terminology and call 
a map $g$ isometric if the following relation holds for some constant $C > 0$: 
\begin{align}
\met< \vv, \vw >_{\vp} =C \met< \dif g(\vv), \dif g(\vw) >_{g(\vp)}, \label{iso_vx}
\end{align}
since Eq.~(\ref{def_iso}) 
is achieved by replacing $g$ with $\tilde{g} = (1 / \sqrt{C}) g$.
%

\section{Related Work}
\textbf{Flow-based model: }Flow-based generative models generate images with astonishing quality \cite{GLOW, NICE}. Flow mechanism explicitly takes the Jacobian of $\vx$ and $\vz$ into account. The transformation function $\vz=f(\vx)$ is learned, calculating and storing the Jacobian of $\vx$ and $\vz$. Unlike ordinary autoencoders, which reverse $\vz$ to $\vx$ with function $g(\cdot)$ different from $f(\cdot)$, inverse function transforms $\vz$ to $\vx$ as $\vx=f^{-1}(\vz)$. Since the model stores Jacobian, $P_{\vx}(\vx)$ can be estimated from $P_{\vz}(\vz)$.
However, in these approaches, the form of $f(\cdot)$ is limited so that the explicit calculation of Jacobian is manageable, such as $f(\cdot)$ cannot reduce the dimension of $\vx$. 

\textbf{Data interpolation with autoencoders: }
For a smooth data interpolation, in \citet{geo1, geo2}, a function learns to map latent variables to a geodesic (shortest path in a manifold) space, in which the distance corresponds to the metric of the data space. 
In \citet{geo3, deep_iso}, a penalty for the anisometricity of a map is added to training loss. 
Although these approaches may remedy scale inconsistency, they do not deal with PDF estimation. Furthermore, the distance for the input data is assumed to be a  Euclidean distance and the cases for other distances are not considered. 

\textbf{Deep image compression (DIC) with RDO: } 
RD theory is a part of Shannon's information theory for lossy compression which formulates the optimal condition between information rate and distortion. 
The signal is  decorrelated by orthonormal transformation such as Karhunen-Lo\`eve transform (KLT) \cite{KLTBook} and discrete cosine transform (DCT). 
In RDO, a cost $L= R + \lambda D$ is minimized at given Lagrange parameter $\lambda$. 
%
Recently, DIC methods with RDO \cite{Balle, Zhou} have been proposed. 
In these works, instead of orthonormal transform in the conventional lossy compression method, a deep autoencoder is trained for RDO. 
In the next section, we explain the idea of RDO guided autoencoder and its relationship with VAE. 

\section{Overview of RDO Guided Approach} 
\subsection{Derivation from RDO in Transform Coding}
\begin{figure}[h]
  \begin{center}
   \includegraphics[width=70mm]{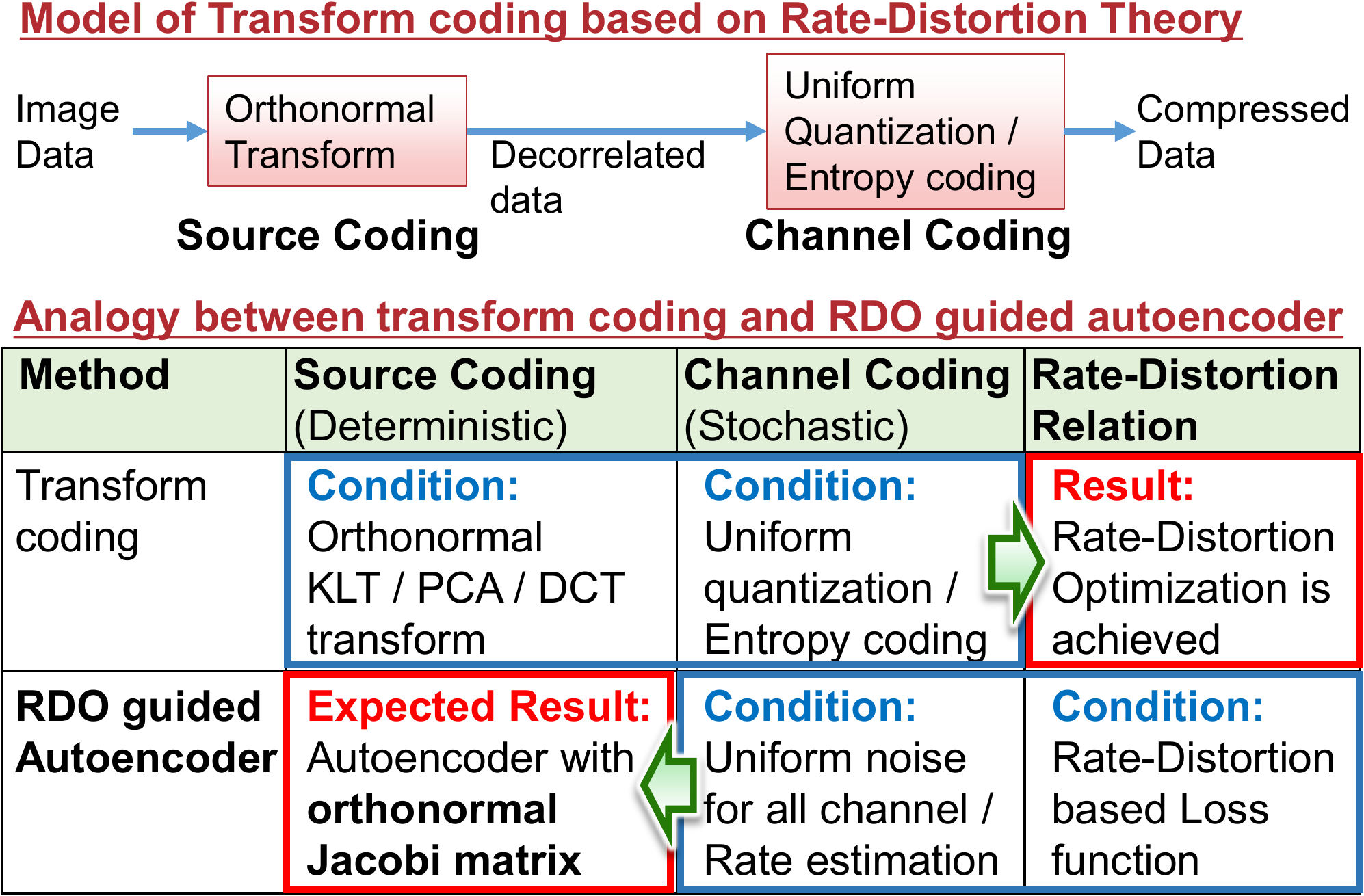}
  \end{center}
  \caption{Overview of our idea. Orthonormal transformation and uniform quantization noise result in an RDO. Our idea is that uniform quantization noise and RDO make an antoencoder to be orthonormal. \\}
  \label{fig:one}
\end{figure}

\mblu
\mmag Figure \ref{fig:one} shows the overview of our idea based on the RDO inspired by transform coding. In the transform coding, the optimal method to encode data with Gaussian distribution is as follows \citep{TransformCoding}. \mblk 
First, the data are transformed deterministically to decorrelated data using  orthonormal transforms such as Karhunen-Lo\`eve transform (KLT) and discrete cosine transform (DCT). Then these decorrelated data are quantized stochastically with uniform quantizer for all channels such that the quantization noise for each channel is equal. Lastly optimal entropy encoding is applied to quantized data where the rate can be calculated by the logarithm of symbol's estimated probability. 
From this fact, we have an intuition that the columns of the Jacobian matrix of the autoencoder forms an orthonormal system if the data were compressed based on RDO with a uniform quantized noise and parametric distribution of latent variables. Inspired by this, we propose autoencoder which scales latent variables according to the definition of \mmag metrics \mblu of data. 
\mblk
\subsection{Relationship with VAE}
\mblu
\mgre There is a number of VAE studies taking RD trade-off into account.
In VAEs, 
it is common to maximize ELBO instead of maximizing log-likelihood of $P_{\vx}(\vx)$ directly. 
In beta-VAE \citep{betaVAE}, the objective function $L_{VAE}$ is described as $L_{VAE} = L_{rec} - \beta L_{kl}$. 
Here, $L_{kl}$ is the KL divergence between the encoder output and prior distribution, usually a Gaussian distribution. By changing $\beta$, the rate-distortion trade-off at desirable rate can be realized as discussed in \citet{BELBO}.

\begin{figure}[h]
  \begin{center}
   \includegraphics[width=70mm]{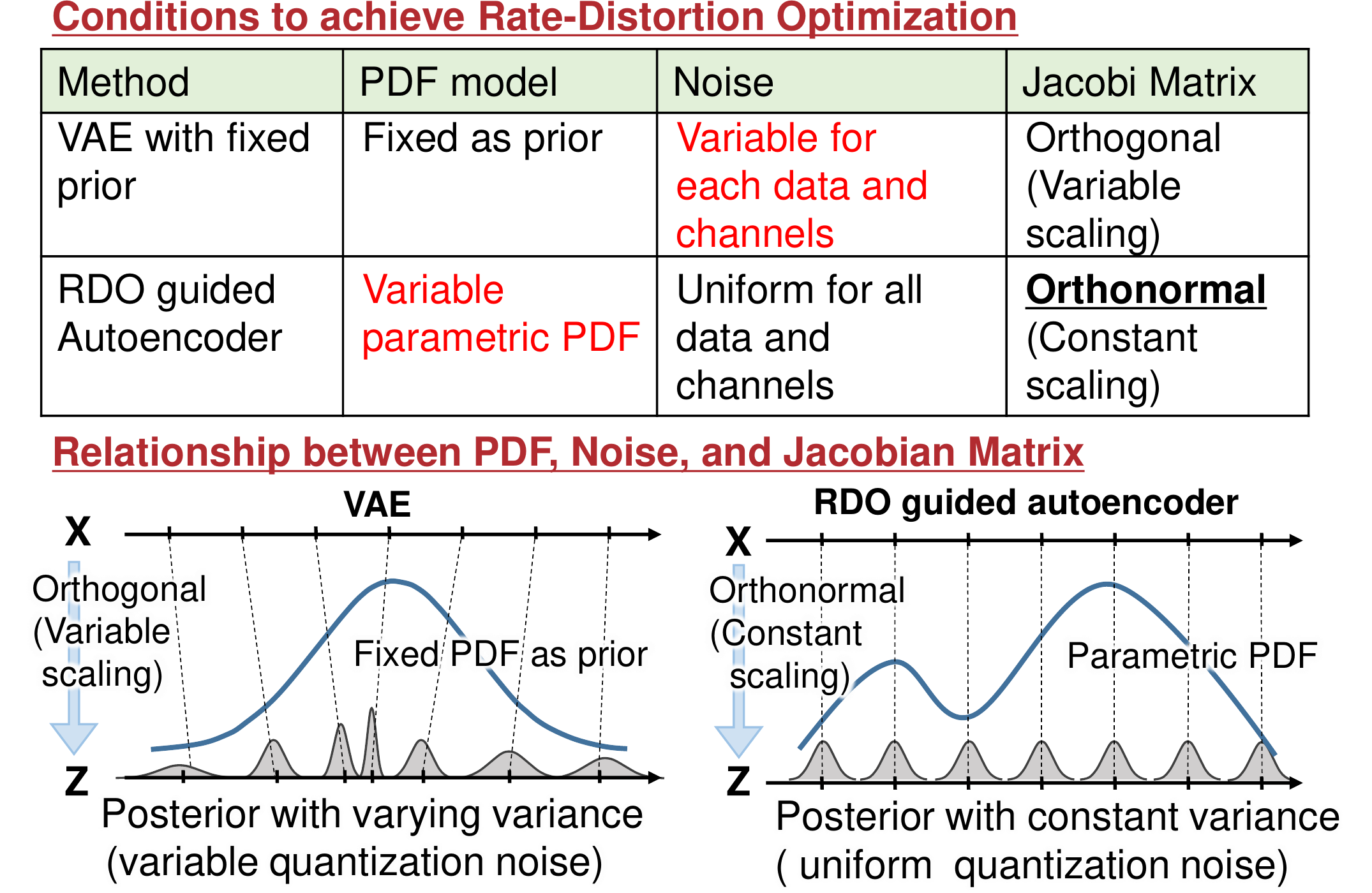}
  \end{center}
  \caption{The condition of RDO in VAE and our method. In VAE, to fit the fixed prior (blue line),  data are transformed anisometrically with precisely controlled noise as a posterior variance (gray area). 
  A wider distribution of noise makes the PDF of transformed data smaller. 
In our method, a parametric prior distribution is estimated, and data is transformed isometrically with uniform noise.}
  \label{fig:two}
\end{figure}

Note that the beta-VAE and the RDO in image compression are analogous to each other. 
That is, ${\beta}^{-1}$, $-L_{kl}$, and $L_{rec}$ correspond to $\lambda$, a rate $R$, and a distortion $D$ respectively. 
However, the probability models of latent variables are quite different. VAE uses a fixed prior distribution. 
This causes a nonlinear scaling relationship between real data and latent variables. 
Figure \ref{fig:two} shows the conditions to achieve RDO in both VAE and RaDOGAGA. 
\mmmag
In VAE, for RDO condition, a nonlinear scaling of the data distribution is necessary to fit prior. 
To achieve it, \citet{EchoNoise} suggested to precisely control noise as a posterior variance for each channel. 

As proven in \citet{VAEPCA}, in the optimal condition, the Jacobian matrix of VAE forms an orthogonal system, but the norm is not constant. 
In RaDOGAGA, uniform quantization noises are added to all channels. 
Instead, a parametric probability distribution should be estimated as a prior. 
As a result, the Jacobian matrix forms an orthonormal system because both orthogonality and scaling normalization are simultaneously achieved. 
As discussed above, the precise noise control in VAE and parametric prior optimization
in RaDOGAGA are essentially the same. 
Accordingly, complexities in methods are estimated to be at the same degree. 
\section{METHOD AND THEORETICAL PROPERTIES}
\label{sec:theory}
\subsection{Method}
Our method is based on the RDO of the autoencoder for the image compression proposed in \citet{Balle} with some modifications. 
In \citet{Balle}, the cost function 
\begin{align}
\label{cost0}
L=R+\lambda D
\end{align}
consists of (i) reconstruction error $D$ between input data and decoder output with noise to latent variable and (ii) rate $R$ of latent variable. \mgre This is analogous to beta-VAE where \mmag $\lambda = \beta ^ {-1}$. 

\begin{figure}[H]
	\centering
	\includegraphics[width=7.5cm]{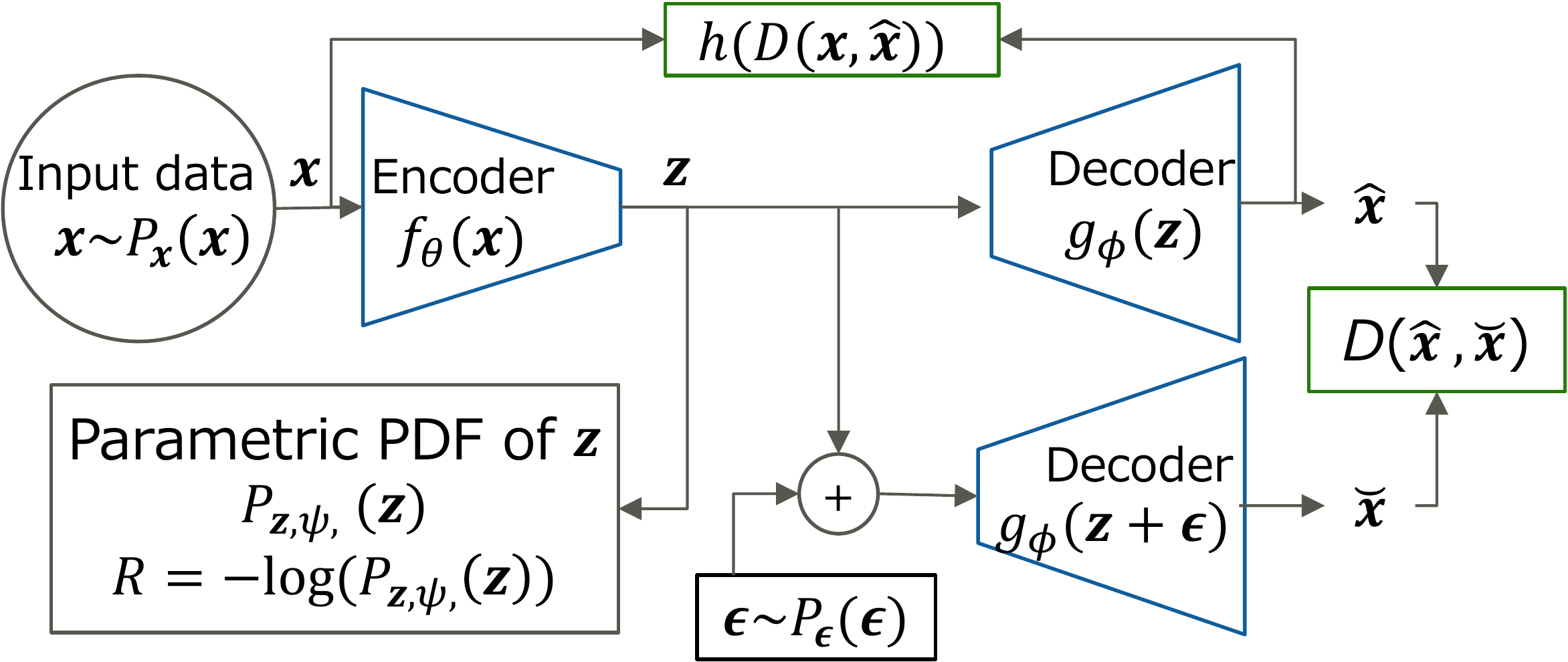}
	\caption{Architecture of RaDOGAGA}
	\label{figArc}
\end{figure}

Figure \ref{figArc} depicts the architecture of our method. 
The details are given in the following. 
Let $\bm x$ be an $M$-dimensional input data, $\sR^M$ be 
a data observation space
endowed with a metric function 
$D(\cdot \ , \ \cdot)$,
and $P_{\vx}({\vx})$ be the PDF of $\bm x$. 
Let $f_\theta({\bm x})$, $g_\phi({\bm z})$, and $P_{\vz, \psi}(\vz)$ be the parametric encoder, decoder, and PDF of the latent variable with parameters $\theta$, $\phi$, and $\psi$.
Note that both of the encoder and decoder are deterministic, while the encoder of VAE is stochastic.  

First, the encoder converts the input data $\bm x$ to an  $N$-dimensional latent variable $\vz$ in a Euclidean latent space $\sR^N$, and then the decoder converts $\vz$ to the decoded data $\hat {\vx}  \in \sR^M$: 
\begin{align}
\vz =f_\theta(\vx), \quad 
\hat {\vx} = g_\phi(\vz). \label{dfn_hatx} 
\end{align}

Let ${\bm \epsilon} \in \sR^N$ be a noise vector to emulate uniform quantization, where each component is independent from others and has an equal 
mean 0 and an equal variance  ${\sigma}^2$: 
\begin{align}
\label{addnoise}
{\bm \epsilon} = (\epsilon_1, \epsilon_2,\ ..\epsilon_N),\ E\left[ \epsilon_i \right] = 0,\ E\left[ \epsilon_i \epsilon_j\right] = {\delta}_{ij} {\sigma}^2. 
\end{align}
Here, $\delta_{ij}$ denotes the Kronecker's delta. 
Given the sum of latent variable $\bm z $ and noise $\bm \epsilon$, another  decoder output $\breve {\bm x}  \in \sR^M$ is obtained as 
\begin{equation}
{\breve {\bm x}} = g_\phi({\bm z} + {\bm \epsilon}) \label{dfn_brevex} 
\end{equation}
with the same parameter $\phi$ used to obtain $\hat {\vx}$. 
This is analogous to the stochastic sampling and decoding procedure in VAE. 

The cost function is defined based on Eq.~(\ref{cost0}) with some modifications as follows:
\begin{eqnarray}
\label{cost1}
L = - \log ({P_{\vz, \psi}({\bm z})})+ \lambda_1 h \left(D \left({\bm x},{\hat {\bm x}}\right)\right) + \lambda_2  D \left({\hat {\bm x}}, {\breve {\bm x}}\right).
\end{eqnarray} 
The first term 
corresponds to the estimated rate of the latent variable. 
We can use arbitrary probabilistic model as $P_{\vz, \psi}(\vz)$. 
For example, 
\citet{Balle} uses univariate independent (factorized) model $P_{\vz,\psi}(\bm z) = \prod_{i=1}^N P_{z_i, \psi}(z_i)$. In this work, a parametric function $c_{\psi}(z_{i})$ outputs cumulative distribution function of $z_{i}$. 
A rate for quantized symbol is calculated by $c_{\psi}(z+\frac{1}{2})-c_{\psi}(z-\frac{1}{2})$, assuming the symbol is quantized with the side length of 1. 
A model based on GMM like \citet{DAGMM}
is another instance. 

The second and the third term in Eq.~(\ref{cost1}) is based on the decomposition 
$D \left({{\bm x}}, {\breve {\bm x}}\right) \simeq D \left({{\bm x}}, {\hat {\bm x}}\right) + D \left({\hat {\bm x}}, {\breve {\bm x}}\right)$ 
shown 
in \citet{VAEPCA}. 
The second term in \mred Eq.~(\ref{cost1}) \mblk purely calculate reconstruction loss as an autoencoder. 
In the RDO, the consideration is trade-off between rate (the first term) and the distortion by the quantization noise (the third term). 
By this decomposition, we can avoid the interference between better reconstruction and RDO trade-off durning the training. 
The weight $\lambda_1$ controls the degree of reconstruction, and $\lambda_2$ ($\simeq \beta^{-1}$ of beta-VAE) controls a scaling between  data and latent spaces respectively.  

The function $h(\cdot)$ in the second term of Eq.~(\ref{cost1}) is a monotonically increasing function. In experiments in this paper, we use $h(d)=\log(d)$. In the theory shown in Appendix \ref{OrthoJacobian}, better reconstruction provide much rigid orthogonality. We find $h(d)=\log(d)$ is much more appropriate for this purpose than $h(d)=d$ as detailed in Appendix \ref{sec:hd}. 
 
The properties of our method shown in the rest of this paper hold 
for a variety of metric function $D(\cdot \ , \ \cdot)$, 
as long as it can be approximated by the following quadratic form in the neighborhood of $\vx$: 
\begin{align}
\label{SecondaryForm}
D(\bm x, \bm x + \bm{\Delta x}) \simeq {\bm{\Delta x}}^\top \ \bm A(\bm x) \ \bm{\Delta x}.
\end{align}
Here, $\bm{\Delta x}$ is an arbitrary infinitesimal variation of $\vx$, and $\mA(\vx)$ is an $M \times M$ positive definite matrix depending on $\bm x$ that corresponds to  
a metric tensor. 
When $D(\cdot \ , \ \cdot)$ is the square of the Euclidean distance, 
$\bm A(\bm x)$ is the identity matrix.
For another instance, a cost with structure similarity (SSIM~\cite{SSIM})  and binary cross entropy (BCE) can also be approximated 
by a quadratic form 
as explained in Appendix \ref{SSIM_EXP}.
By deriving parameters that minimize the average of Eq.~(\ref{cost1}) according to $\vx \sim P_{\vx}(\vx)$ and ${\bm \epsilon} \sim P_{ {\bm \epsilon} }({\bm \epsilon})$, the encoder, decoder, and probability distribution of the latent space are trained as
\begin{align}
\label{getparams}
\theta, \phi, \psi = \argmin_{\theta, \phi, \psi} (E_{\vx \sim P_{\vx}(\vx),\ {\bm \epsilon} \sim P_{{\bm \epsilon}} ({\bm \epsilon})}[\ L\ ]).
\end{align} 
\subsection{Theoretical Properties} \label{TheoreticalProps} 
\mblu
In this section, we explain the theoretical properties of the method. 
To show the essence in a simple form, we first (formally) consider the case $M = N$. 
The theory for $M > N$ is then outlined. All details are given in Appendices. 

We begin with examining the condition to minimize the loss function analytically, assuming that the reconstruction part is trained enough so that $\vx \simeq \hat{\vx}$. In this case, the second term
in  Eq.~(\ref{cost1}) can be ignored. 
Let $\mJ(\vz) = \partial \vx / \partial  \vz =  \partial g_{\phi}(\vz) / \partial  \vz \in \sR^{N \times N}$ be the Jacobian matrix between the data space and latent space, 
which is assumed to be full-rank at every point. 
Then, $\breve{\vx}-\hat{\vx}$ can be approximated as 
$\acute{\bm \epsilon} =  \sum_{i=1}^N   \epsilon_i (\partial \vx/\partial z_{i})$  
through the Taylor expansion. 
By applying $E[\epsilon _i \epsilon _j] = \sigma ^2 \delta_{ij}$ and Eq.~(\ref{SecondaryForm}), 
the expectation of the third term in Eq.~(\ref{cost1}) is rewritten as
\begin{eqnarray}
\label{cost2_2}
\underset{{}^{{{\bm \epsilon} \sim P_{\bm \epsilon}(\bm \epsilon)}}}{E} 
\left[\acute{{\bm \epsilon}}^{\top}  \mA (\bm x)  \acute{{\bm \epsilon}}\right]
= \sigma^2  \sum_{j=1}^{N} \left( \frac{\partial \vx}{\partial z_{j}} \right)^{\top} \mA (\bm x) \left(\frac{\partial \vx}{\partial z_{j}} \right). 
\end{eqnarray}
As is well known, the relation between 
$P_{\vz}(\vz)$ and $P_{\vx}(\vx)$
in such case is described 
as $P_{\vz}(\vz) = |\det(\mJ(\vz))| P_{\vx}(\vx)$. 
The expectation of $L$ in Eq.~(\ref{cost1}) is thus approximated as 
\begin{align}
\label{cost2_f}
\underset{{}^{{{\bm \epsilon} \sim P_{\bm \epsilon}(\bm \epsilon)}}}{E} 
 \left[L \right]
& \simeq - \log( | \det( \mJ(\vz))| )-\log(P_{\vx}(\vx))  \nonumber \\
&+  \lambda_2 \sigma ^2 \sum_{j=1}^{N}   \left( \frac{\partial \vx}{\partial z_{j}} \right)^\top  \mA (\vx) \left( \frac{\partial \vx}{\partial z_{j}} \right). 
\end{align}
By differentiating Eq.~(\ref{cost2_f}) by $\partial \vx / \partial z_{j}$, the following equation is derived as a condition to minimize the expected  loss:
\begin{align}
\label{difcost_s}
2 \lambda_2 {\sigma}^2 \mA(\vx)  \left( \frac{\partial \vx}{\partial z_{j}} \right)  
=
\frac{1}{\det(\mJ(\vz))} \tilde{\mJ}(\vz)_{:, j}, 
\end{align}
where $\tilde{\mJ}(\vz)_{:, j} \in \sR^{N}$ is the $j$-th column vector of the cofactor matrix of $\mJ(\vz)$.  Due to the trait of cofactor matrix, $(\partial \vx/ \partial z_{i} )^\top \tilde{\mJ}(\vz)_{:,j} = \delta_{ij} \det(\mJ(\vz))$ holds. 
Thus, the following relationship is obtained by multiplying Eq.~(\ref{difcost_s}) by $(\partial \vx/ \partial z_{i} )^\top$  from the left and rearranging the results: 
\begin{align}
\label{Ortho}
\left({\frac{\partial \vx}{\partial z_i}}\right)^\top \bm A(\bm x) \left({\frac{\partial \bm x}{\partial z_j}}\right)
=
\frac{1}{2 \lambda_2 {\sigma}^2} \delta_{ij}. 
\end{align}
This means that \emph{the columns of the Jacobian matrix of two spaces form a constantly-scaled orthonormal system  with respect to the inner product defined by $\mA(\vx)$ for all $\vz$}.
\mblu
%

Given tangent vectors $\vv_{z}$ and $\vw_{z}$ in the tangent space of $\sR^{N}$ at $\vz$ represented as elements of $\sR^{N}$, 
let $\vv_{x}$ and $\vw_{x}$ be the corresponding tangent vectors 
represented as elements of $\sR^{M} = \sR^{N}$.
The following relation holds due to Eq.~(\ref{Ortho}), which means that 
\emph{the map is isometric 
in the sense of Eq.~(\ref{iso_vx})}: 
\begin{align}
& \vv_{x}\mA(\vx)\vw_{x} =\sum_{i=0}^N \sum_{j=0}^N \left(\frac{\partial \vx}{\partial z_{i}}v_{z{i}} \right)^\top \mA(\vx) \left(\frac{\partial \vx}{\partial z_{j}}w_{z{j}} \right) \nonumber \\
& =\frac{1}{2 \lambda_2 {\sigma}^2} \sum_{i=0}^{N}v_{z_{i}}w_{z_{i}} 
 =\frac{1}{2 \lambda_2 {\sigma}^2} \vv_{z} \cdot \vw_{z}. \label{isometricity}
\end{align}
Since $f_{\theta}(\cdot)$ and $g_{\phi}(\cdot)$ acts like the inverse functions of each other when restricted on the input data, isometric property holds for both.

Even for the case $M>N$,  
equations in the same form as Eqs.~(\ref{Ortho}) and (\ref{isometricity}) can be derived essentially  in the same manner (Appendix \ref{OrthoJacobian}); that is, \emph{RaDOGAGA achieves isometric data embedding for the case $M>N$ as well}.


Now let us proceed to PDF estimation.
First, we (formally) consider the case $M = N$ as before. 
Note that Eq.~(\ref{Ortho}) can 
be expressed as follows: $\mJ(\vz)^\top \mA(\vx) \mJ(\vz) = (1/2 \lambda_2 {\sigma}^2) \mI_{N}$ ($\mI_{N}$ is the $N \times N$ identity matrix). 
We have the following equation by taking the determinants of both sides of this 
and using the properties of the determinant: 
$| \det(\mJ(\vz)) | = (1/2 \lambda_2 {\sigma}^2)^{N/2} \det(\mA(\vx))^{-1/2}$. 
Note that $\det(\mA(\vx)) = \prod_{j=1}^{N} \alpha_{j}(\mA(\vx)) $, where $0<  \alpha_{1}(\mA(\vx)) \le \cdots \le  \alpha_{N}(\mA(\vx))$ are the eigenvalues of $\mA(\vx)$.   
Thus, $P_{\vz}(\vz)$ and $P_{\vx}(\vx)$ 
are related in the following form:
\begin{align}
P_{\vx}(\vx) = \left( \frac{1}{2 \lambda_{2} \sigma^{2}} \right)^{- \frac{N}{2}}  \biggl( \prod_{j=1}^{N} \alpha_{j}(\mA(\vx)) \biggr)^{\frac{1}{2}} P_{\vz}(\vz) . 
\label{relation_pdf}
\end{align}

To consider the relationship of $P_{\vz}(\vz)$ and $P_{\vx}(\vx)$ for $M > N$, we follow the manifold hypothesis and  assume 
the situation where 
the data $\vx$ substantially exist in the vicinity of a low-dimensional manifold $\gM_{\vx}$, 
and 
$\vz \in \sR^{N}$ can sufficiently capture its feature. 
In such case, 
we can regard that 
the distribution of $\vx$ away from $\gM_{\vx}$ is negligible 
and the ratio of $P_{\vz}(\vz)$ and $P_{\vx}(\vx)$
is equivalent to  
that of the volumes of corresponding regions in $\sR^{N}$ and $\sR^{M}$. 
This ratio is shown to be $J_{sv}(\vz)$,  
the product of the singular values of $\mJ(\vz)$, 
and we get the relation $P_{\vz}(\vz) = J_{sv}(\vz) P_{\vx}(\vx)$.
We can further show that $J_{sv}(\vz) $ is also $ (1/2 \lambda_2 {\sigma}^2) ^{N/2} ( \prod_{j=1}^{N} \alpha_{j}(\mA(\vx)) )^{-1/2}$ under a certain condition that includes the case $\mA(\vx) = \mI_{M}$ (see Appendix \ref{app_prob}). 
Consequently, Eq.~(\ref{relation_pdf}) holds even for the case $M>N$. 
In such case, $P_{\vz}(\vz)$ and  
$(\prod_{j=1}^{N} \alpha_{j}(\mA(\vx)) )^{-1/2} P_{\vx}(\vx)$, the probability distribution function of $\vx$ modified by a metric depending scaling,  becomes proportional. 
As a result, when we obtain a parameter $\psi$ 
attaining $P_{\vz, \psi}(\vz) \simeq P_{\vz}(\vz)$ by training, 
$P_{\vx}(\vx)$ is proportional to $P_{\vz, \psi}(\vz)$ with a metric depending scaling $( \prod_{j=1}^{N} \alpha_{j}(\mA(\vx)) )^{1/2}$ as:
\begin{align}
P_{\vx}(\vx) \propto \biggl( \prod_{j=1}^{N} \alpha_{j}(\mA(\vx)) \biggr)^{\frac{1}{2}} P_{\vz, \psi}(\vz).
\end{align}
In the case of $\mA(\vx) = \mI_{M}$ (or more generally $\kappa \mI_{M}$ for a constant $\kappa > 0$ ), $P_{\vx}(\vx)$ is simply proportional to  $P_{\vz, \psi}(\vz)$:
\begin{align}
P_{\vx}(\vx) \propto P_{\vz, \psi}(\vz).
\end{align}

\section{Experimental Validations}
Here, we show the properties of our method experimentally. In Section \ref{exp_iso}, we examine the isometricity of the map  as in Eq.~(\ref{iso_vx}) with real data. In Section \ref{exp_toy}, we confirm the proportionality of PDFs as in Eq.~(\ref{relation_pdf}) with toy data. In Section \ref{exp_ano}, the usefulness is validated in anomaly detection. 
\subsection{Isometric Embedding}\label{exp_iso}
In this section, we confirm that our method can embed data in the latent space isometrically.
First, a randomly picked data point $\vx$ is mapped to $\vz(=f_{\theta}(\vx))$. Then, let $\vv_z$ be a small tangent vector in the latent space. The corresponding tangent vector in the data space $\vv_x$ is approximated by $g(\vz + \vv_z) - g(\vz)$.  
Given randomly generated two different tangent vectors $\vv_z$ and $\vw_z$, $\vv_z \cdot \vw_z$ is compared with ${\vv_x}^\top \bm{A}(\vx) \vw_x$. 
We use the CelebA dataset~\citep{CelebA}\footnote[1]{\url{http://mmlab.ie.cuhk.edu.hk/projects/CelebA.html}} that consists of 202,599 celebrity images. Images are center-cropped with a size of 64 x 64.
\subsubsection{CONFIGURATION}
In this experiment, factorized distributions \citep{Balle} are used to estimate $P_{\vz, \psi}(\vz)$ \footnote[2]{Implementation is done with a library for TensorFlow  provided at \url{https://github.com/tensorflow/compression}  with default parameters.}. 
The encoder part is constructed with four convolution (CNN) layers and two fully connected (FC) layers. For CNN layers, the kernel size is 9$\times$9 for the first one and 5$\times$5 for the rest. The dimension is 64, stride size is 2, and activation function is the generalized divisive normalization (GDN)~\citep{GDN}, which is suitable for image compression, for all layers. The dimensions of FC layers are 8192 and 256. For the first one, $softplus$ function is attached. The decoder part is the inverse form of the encoder.
For comparison, we evaluate beta-VAE with the same form of autoencoder with 256-dimensional $\vz$. 
In this experiment, we test two different metrics; $MSE$, where $\mA(\vx) = \frac{1}{M} \mI_{M}$, 
and $1-SSIM$, where $\mA(\vx) = \left( \frac{1}{2{\mu_x}^2} \mW_{m} + \frac{1}{2{\sigma_x}^2} \mW_{v} \right)$. 
$\mW_{m} \in \sR^{M\times M}$ is a matrix such that all elements are $\frac{1}{M^2}$ and 
$\mW_{v} = \frac{1}{M} \mI_{M} - \mW_{m}$.
Note that, in practice, $1-SSIM$ for an image is calculated with a small window. In this experiment the window size is 11$\times$11, and this local calculation is performed for the entire image with the stride size of 1. The cost is the average of local values. For the second term in Eq.~(\ref{cost1}), $h(d)$ is $\log(d)$ and $\mA(\vx) = \frac{1}{M} \mI_{M}$. 
For beta-VAE, we set $\beta^{-1}$ as $1\times10^{5}$ and $1\times10^{4}$ regarding to the training with $MSE$ and $1-SSIM$ respectively. 
For RaDOGAGA, $(\lambda_{1}, \lambda_{2})$ is (0.1, 0.1) and (0.2, 0.1). Optimization is done by Adam optimizer \citep{adam} with learning rate $1\times10^{-4}$. All models are trained so that the $1-SSIM$ between the input and reconstructed images is approximately 0.05.
\subsubsection{Results}
\begin{figure}[]
 \begin{minipage}[b]{0.49\linewidth}
  \centering
  \includegraphics[keepaspectratio, scale=0.17]
  {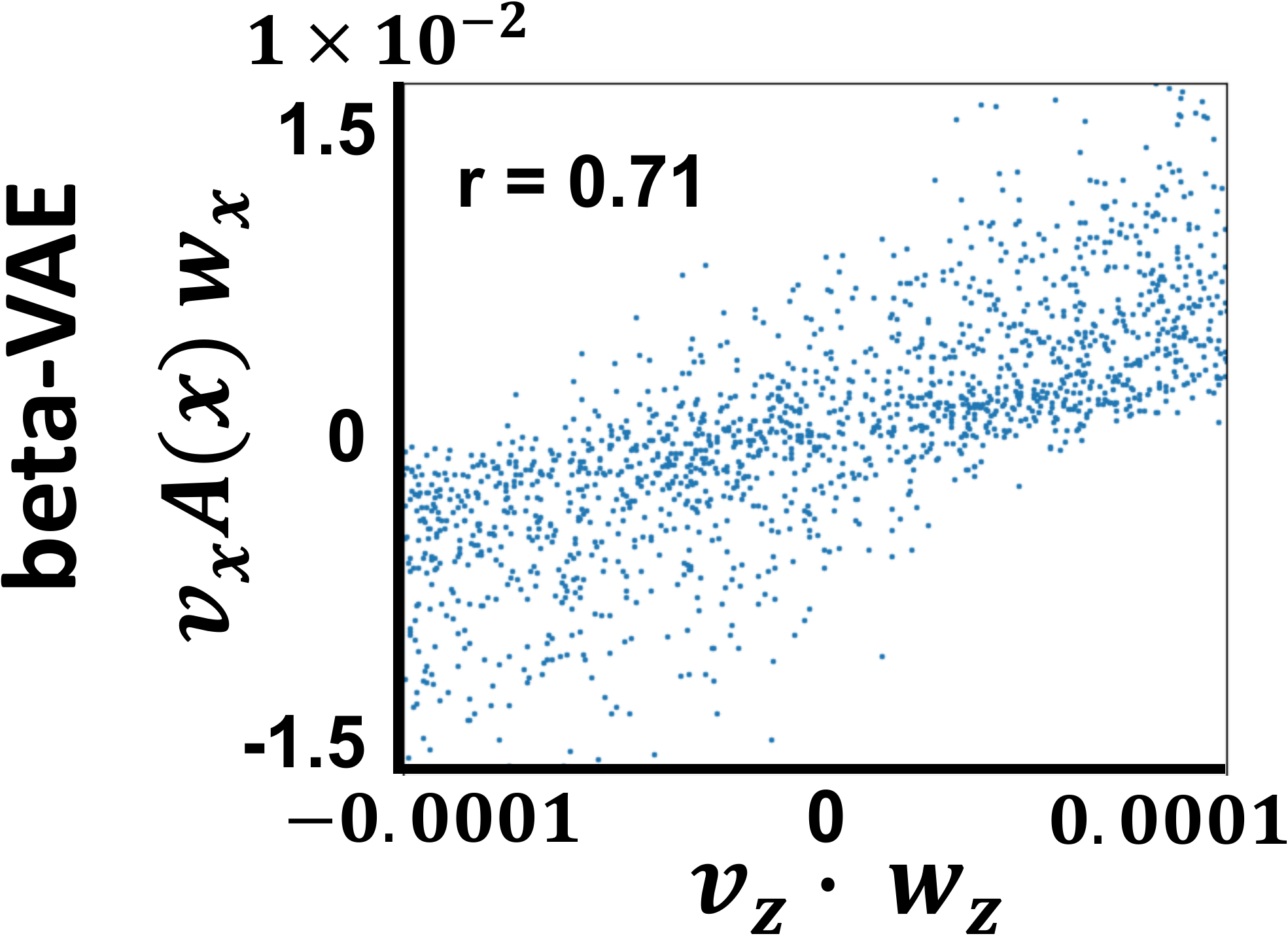}
 \end{minipage}
 \begin{minipage}[b]{0.49\linewidth}
  \centering
  \includegraphics[keepaspectratio, scale=0.17]
  {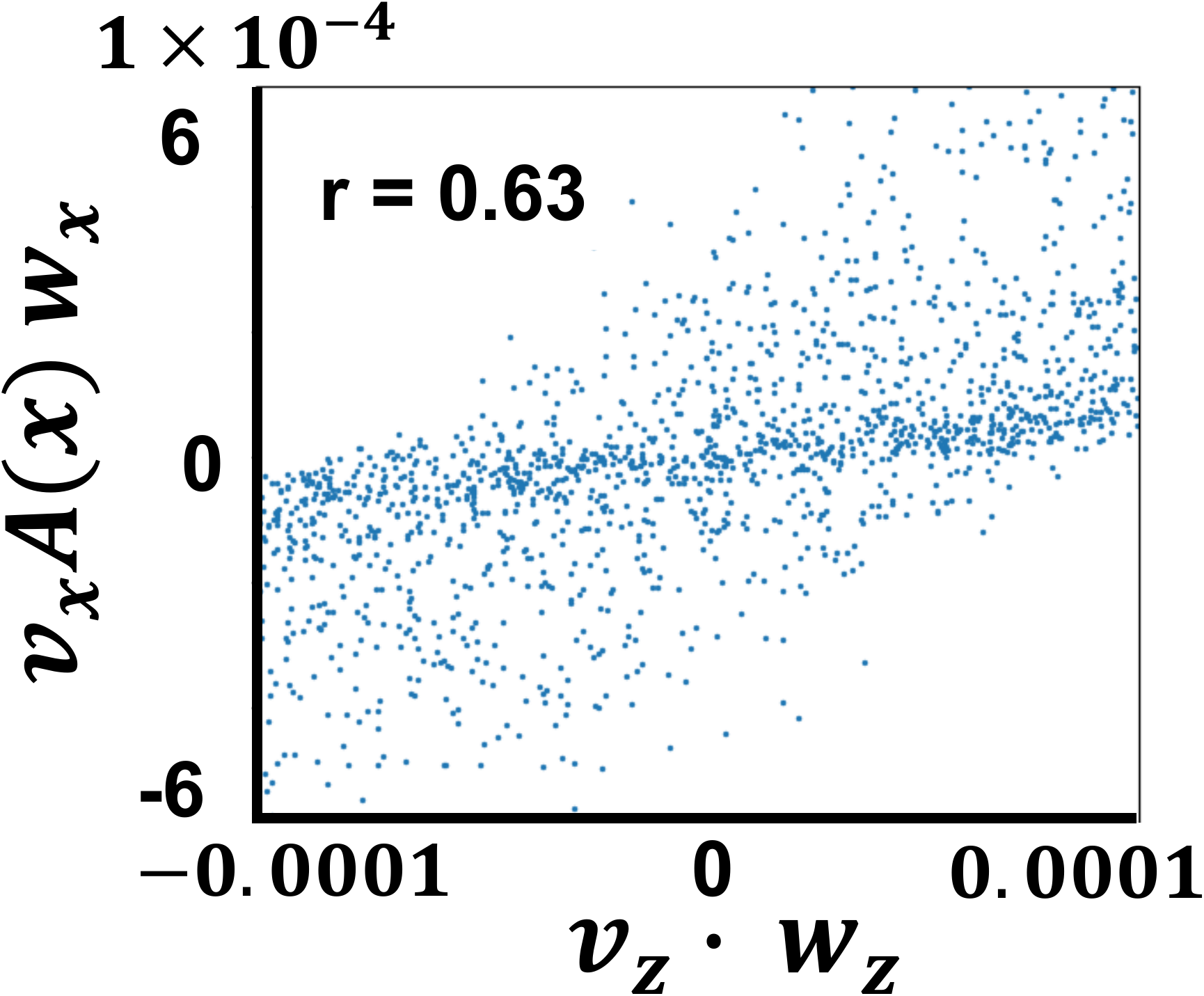}
 \end{minipage}
 \begin{minipage}[b]{0.49\linewidth}
  \centering
  \includegraphics[keepaspectratio, scale=0.17]
  {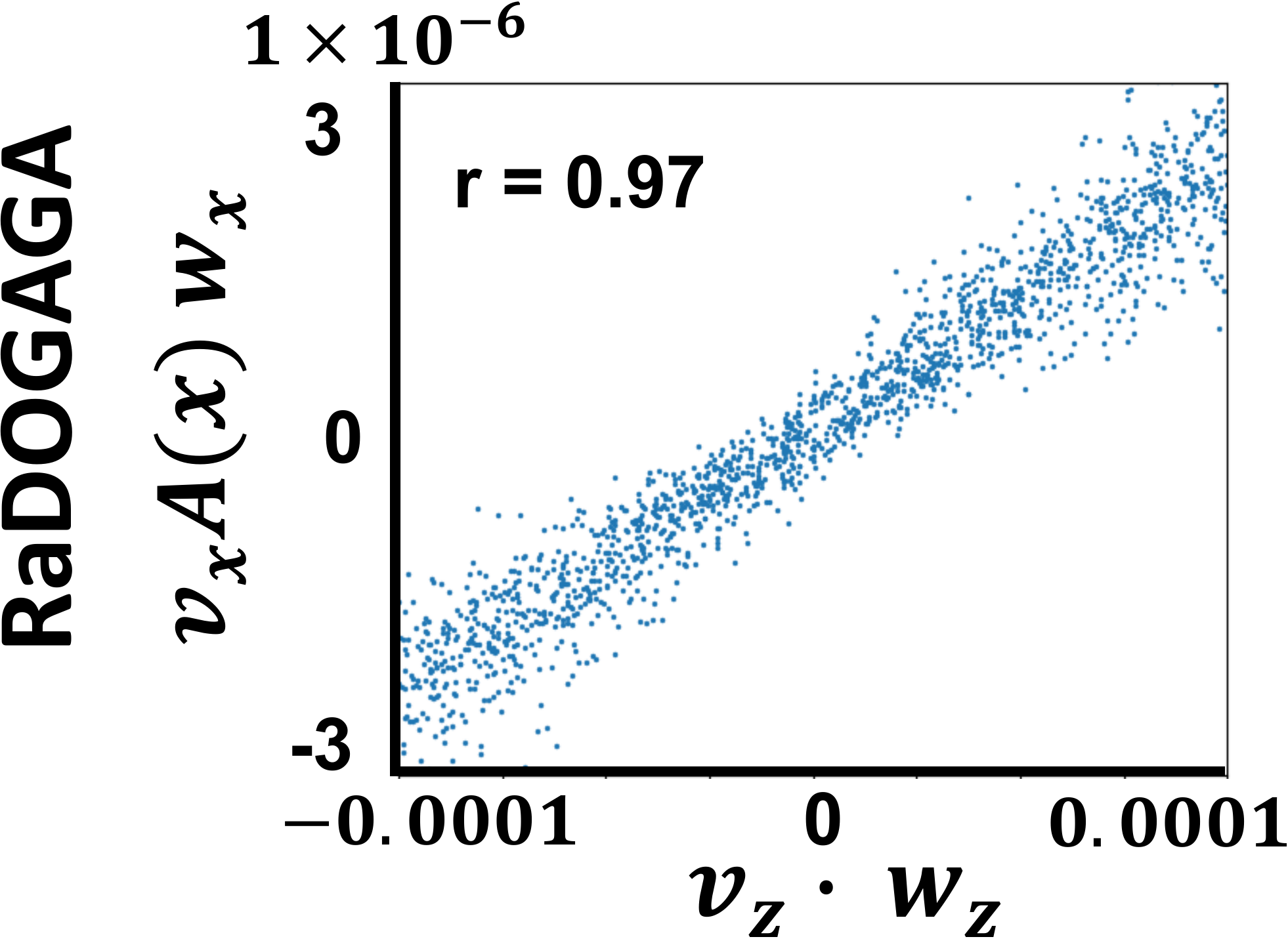}
  \subcaption{$MSE$}
 \end{minipage}
 \begin{minipage}[b]{0.49\linewidth}
  \centering
  \includegraphics[keepaspectratio, scale=0.17]
  {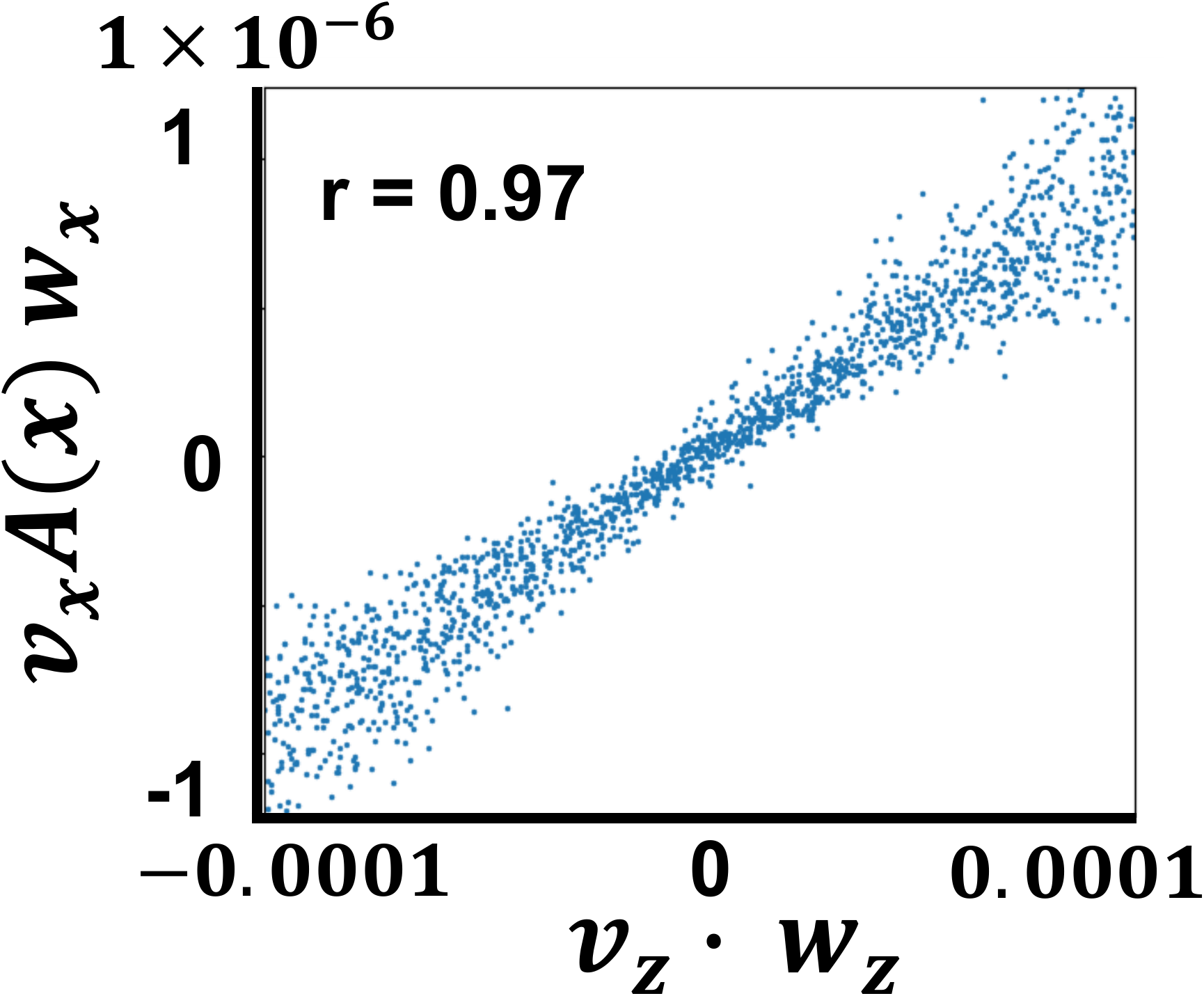}
  \subcaption{$1-SSIM$}
 \end{minipage}
 \caption{Plot of $\vv_z \cdot \vw_z$ (horizontal axis) and ${\vv_x}^\top \bm{A}(\vx) \vw_x$ (vertical axis). In beta-VAE (top row), the correlation is weak 
 whereas in our method (bottom row) we can observe proportionality. }\label{iso}
\end{figure}

Figure \ref{iso} depicts $\vv_z \cdot \vw_z$ (horizontal axis) and ${\vv_x}^\top \bm{A}(\vx) \vw_x$ (vertical axis). The top row is the result of beta-VAE and the bottom row shows that of our method. In our method, $\vv_z \cdot \vw_z$ and ${\vv_x}^\top \bm{A}(\vx) \vw_x$ are almost proportional regardless of the metric function. The correlation coefficients $r$ reach 
0.97, 
whereas that of beta-VAE are around 0.7. 
It can be seen that our method enables isometric embedding to a Euclidean space even with this large scale real dataset. 
For interested readers, we provide the experimental results with the MNIST dataset in Appendix \ref{app_iso}.
\subsubsection{Consistency to Nash Embedding Theorem}
\begin{figure}[t]
 \begin{minipage}[]{0.49\linewidth}
  \centering
  \includegraphics[keepaspectratio, scale=0.2]
  {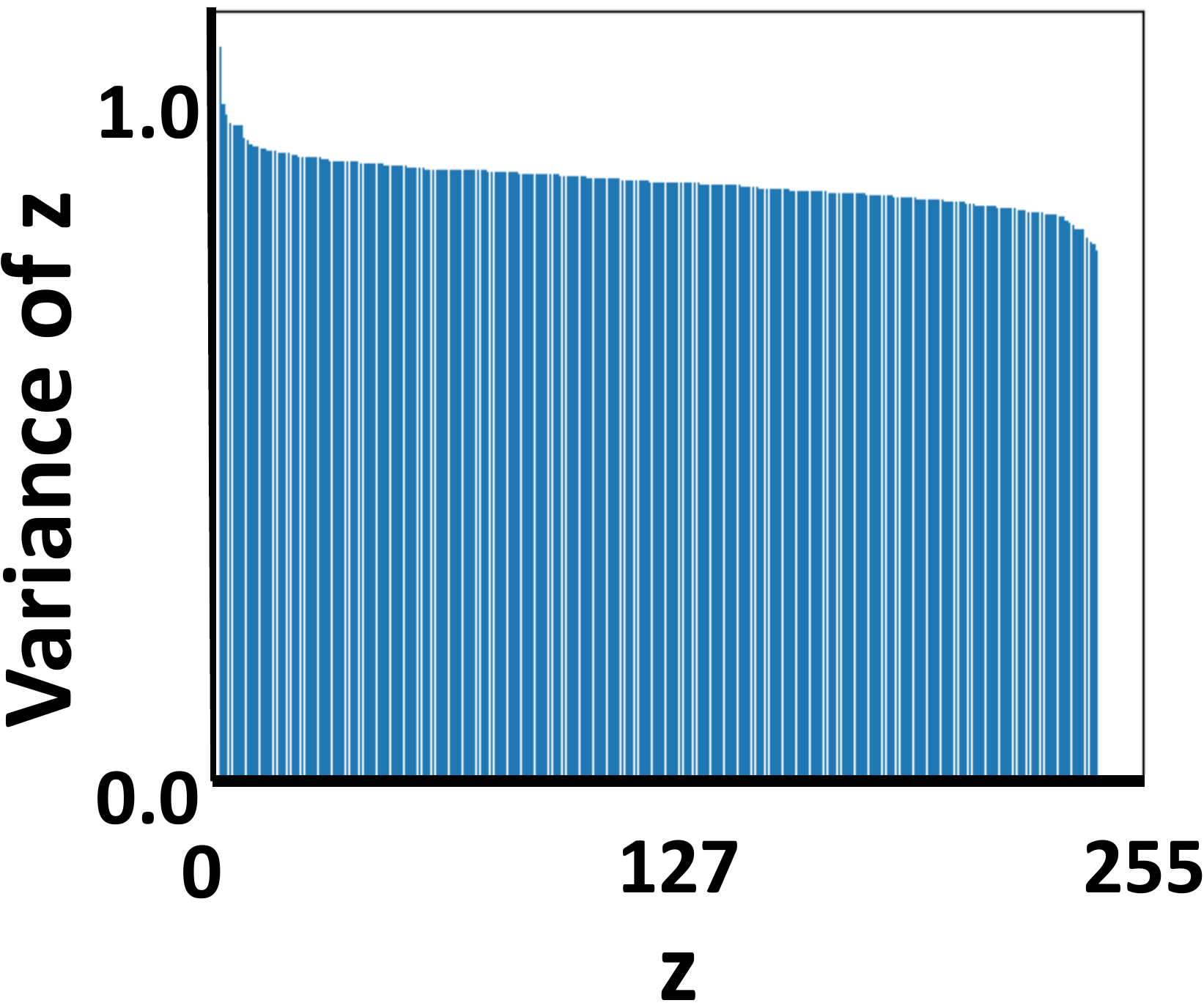}
  \subcaption{beta-VAE}
	\label{fig:var_vae_mse}
 \end{minipage}
  \begin{minipage}[]{0.49\linewidth}
  \centering
  \includegraphics[keepaspectratio, scale=0.2]
  {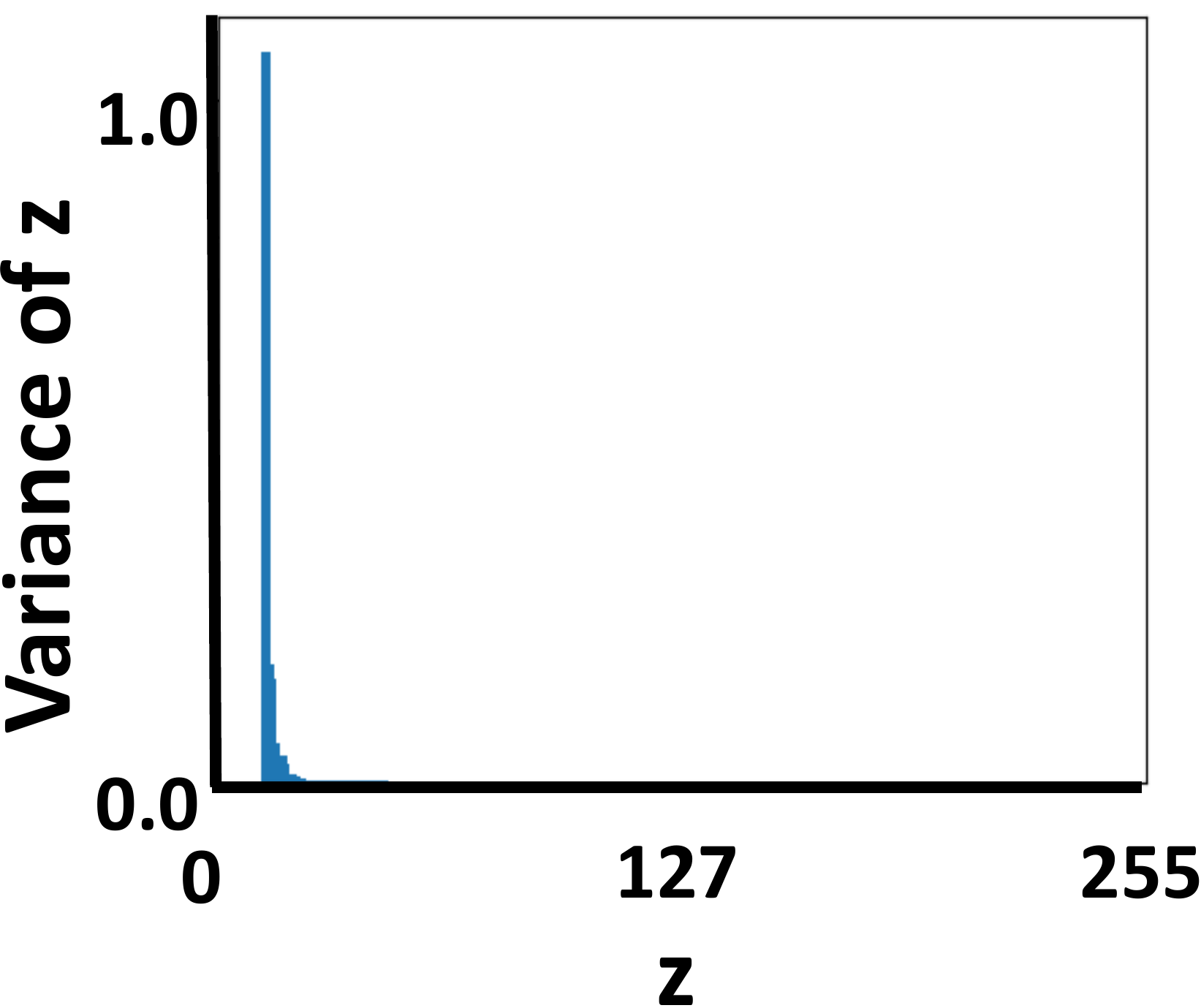}
  \subcaption{RaDOGAGA}
	\label{fig:var_rd_mse}
 \end{minipage}
 \caption{Variance of $\vz$. In beta-VAE, variances of all dimensions are trained to be 1. In RaDOGAGA, the energy is concentrated in few dimensions.}\label{fig:variance}
\end{figure}
\begin{figure*}[t]
 \begin{minipage}[b]{0.49\linewidth}
  \centering
  \includegraphics[keepaspectratio, scale=0.35]
  {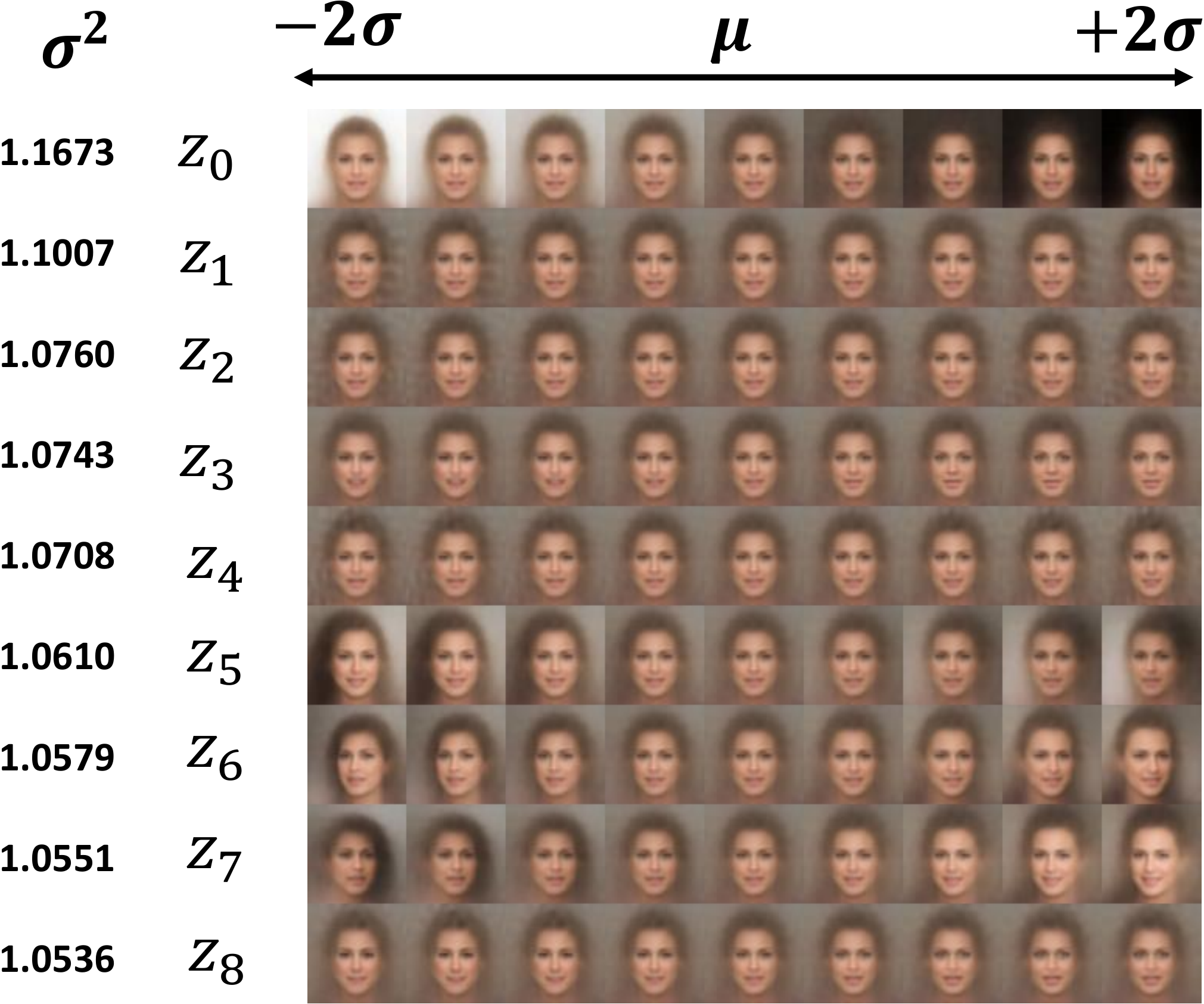}
  \subcaption{beta-VAE}\label{vae_sample}
 \end{minipage}
\begin{minipage}[b]{0.49\linewidth}
  \centering
  \includegraphics[keepaspectratio, scale=0.35]
  {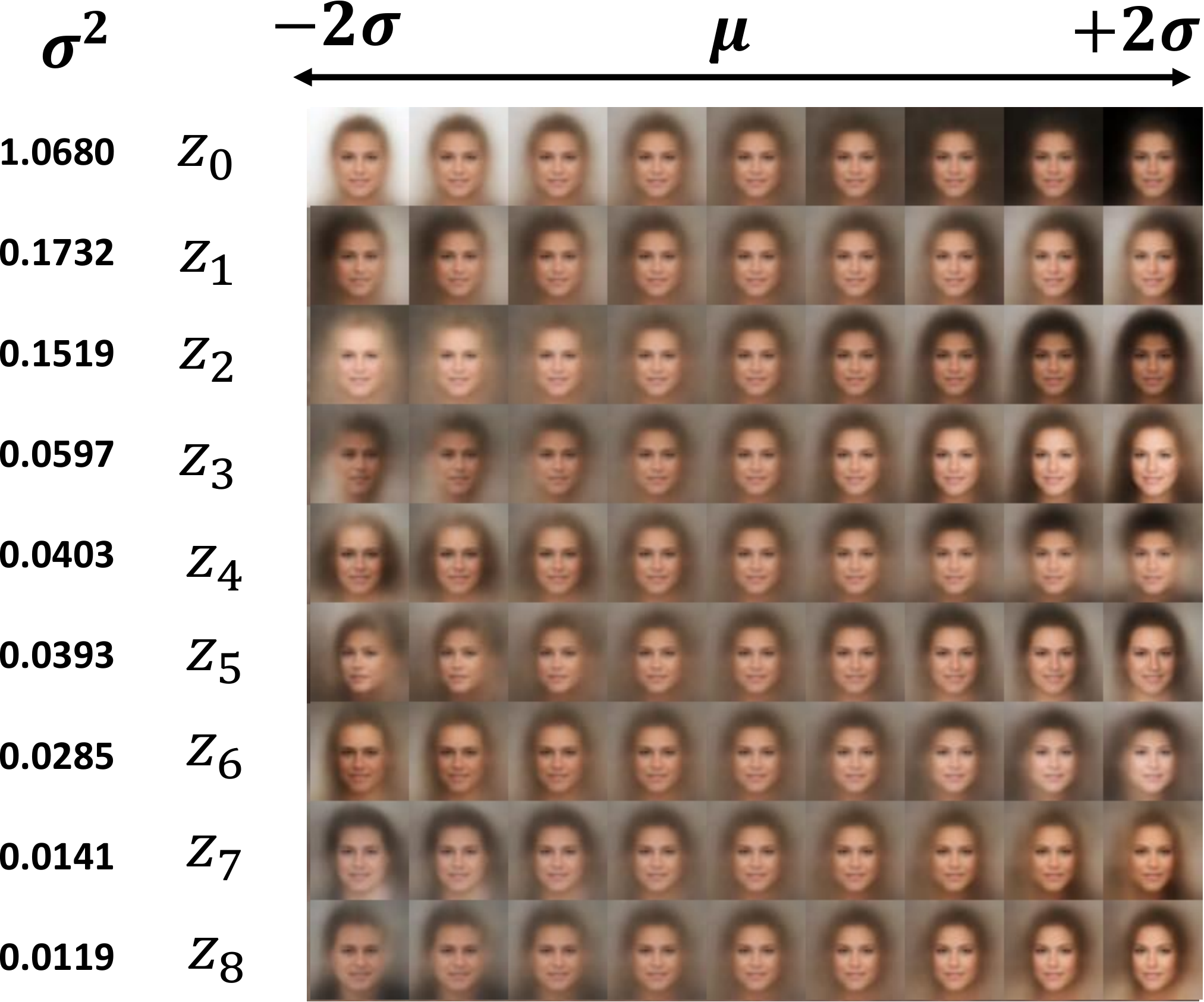}
  \subcaption{RaDOGAGA}\label{rdae_sample}
 \end{minipage}
\caption{Latent space traversal of variables with top-9 variance. In beta-VAE, some latent variables do not influence 
the visual so much even though they have almost the same variance. In RaDOGAGA, all latent variables with large variance have important information for image.}\label{fig:sample}
\end{figure*}
As explained in Introduction, the Nash embedding theorem and manifold hypothesis are behind our exploration of the isometric embedding of input data. 
Here, the question is whether the trained model satisfied the condition that $\dim \gM_{ \vx } < N$. 
With RaDOGAGA, we can confirm it by observing the variance of each latent variable. 
Because the Jacobian matrix forms an orthonormal system, RaDOGAGA can work like principal component analysis (PCA) and evaluates the importance of each latent variable. The theoretical proof for this property is described in Appendix \ref{EX_PCA}. 
Figure \ref{fig:variance} shows the variance of each dimension of the model trained with $MSE$. The variance concentrates on the few dimensions. This means that $\sR^{N}$ is large enough to represent the data. 
Figure \ref{fig:sample} shows the decoder outputs when each component $z_{i}$ is traversed from $-2\sigma$ to $2\sigma$, fixing the rest of $\vz$ as the mean. 
Note the index $i$ is arranged in a descending order of $\sigma^{2}$. 
Here, $\sigma^2$ and $\mu$ for the $i$-th dimension of $\vz(=f_{\theta}(\vx))$ are $Var[z_i]$ and $E[z_i]$ respectively with all data samples. 
From the top, each row corresponds to $z_{0} $, $z_{1} $, $z_{2}$  ..., and the center column is mean. In Fig.~\ref{rdae_sample}, the image changes visually in any dimension of $\vz$, whereas in Fig.~\ref{vae_sample}, depending on the dimension $i$, there are cases where no significant changes can be seen (such as $z_{1} $, $z_{2}$, $z_{3}$, and so on).
In summary, we can qualitatively observe that $Var[z_i]$ corresponds to the eigenvalue of PCA; that is, a latent variable with a larger $\sigma$ have bigger impact on image. 

These results suggest that the important information to express data are concentrated in the lower few dimensions and the dimension number of 256 is large enough to satisfy $ \dim \gM_{ \vx } < N$. 
To confirm the sufficiency of the dimension is difficult in beta-VAE because 
$\sigma^2$ should be 1 for all dimensions because it is trained to fit to the prior. 
However, some dimensions have a large impact on the image, meaning that $\sigma$ does not work as the measure of importance. 

We believe that this PCA-like trait is very useful for the interpretation of latent variables. 
For instance, if the metric function were designed 
so as to 
reflect semantics, important variables for a semantics are easily found. 
Furthermore, 
we argue that this is a promising way to capture the minimal feature to express data, which is one of the goals of machine learning. 
\subsection{PDF Estimation with Toy Data}
\label{exp_toy}
In this section, we describe our experiment using toy data to demonstrate whether the probability density function of the input data $P_{\vx}(\vx)$ and that of the latent variable estimated in the latent space $P_{\vz, \psi}(\vz)$ are proportional to each other as in theory.
First, we sample data points 
$\vs=(\vs_{1}, \vs_{2}, ..., \vs_{n} ,..., \vs_{10,000}) \in \sR^{3 \times 10,000}$
from three-dimensional GMM consists of three mixture-components with mixture weight $\bm{\pi} = (1/3, 1/3, 1/3)$, 
mean $\bm{\mu}_{1} = (0, 0, 0)$, $\bm{\mu}_{2} = (15, 0, 0)$, $\bm{\mu}_{3} = (15, 15, 15)$, 
and covariance $\bm{\Sigma}_{k} =  \mathrm{diag} (1, 2, 3)$. 
$k$ is the index for the mixture-component.
Then, we scatter $\vs$ with uniform random noise $\vu \in \sR^{3\times 16}$, $u_{dm}\sim U_{d}\left(-\frac{1}{2}, \frac{1}{2}\right)$, where $d$ and $m$ are index for dimension of sampled data and scattered data. The $\vu_{d}$s are uncorrelated with each other.
We produce 16-dimensional input data as $\vx_{n}=\sum_{d=1}^{3}\vu_{d} s_{nd}$ with a sample number of 10,000 in the \mmred end. \mblk
The appearance probability of input data $P_{\vx}(\vx)$ is equals to a generation probability of $\vs$.
\subsubsection{CONFIGURATION}
In the experiment, we estimate 
$P_{\vz, \psi}(\vz)$ using GMM with parameter $\psi$ as in DAGMM \citep{DAGMM}.
Instead of  
the EM algorithm ,
GMM parameters are learned using Estimation Network (EN), which consists of a multi-layer neural network.
When the GMM consists of $N$-dimensional Gaussian distribution $\mathscr{N}_{N}(\vz; \bm{\mu}, \bm{\Sigma})$ with $K$ mixture-components, and $L$ is the size of batch samples, EN outputs the mixture-components membership prediction as a $K$-dimensional vector $\widehat{\bm{\gamma}}$ as follows:
\begin{align}
\vp = EN(\vz;\psi), \widehat{\bm{ \gamma}} = softmax(\vp).
\end{align}
Then, $k$-th mixture weight $\widehat{\pi}_{k}$, mean $\widehat{\bm{ \mu}}_{k}$, covariance $\widehat{\bm{\Sigma}}_{k}$, and entropy $R$ of $\vz$ are further calculated as follows:\\ 
$\widehat{\pi}_{k}=\sum_{l=1}^L {\widehat{\gamma}_{lk}}/{L}$,  
$\quad \widehat{\bm{\mu}}_{k}={\sum_{l=1}^L\widehat{\gamma}_{lk}\vz_{l}}/{\sum_{l=1}^L\widehat{\gamma}_{lk}}$, \\
$\widehat{\bm{\Sigma}}_{k}={\sum_{l=1}^L\widehat{\gamma}_{lk}(\vz_{l}-\widehat{\bm{\mu}}_k)(\vz_{l}-\widehat{\bm{\mu}}_k)^{\top}}/{\sum_{l=1}^L\widehat{\gamma}_{lk}}$, \\
$R=-\log \left( \sum ^{K}_{k=1} \mathscr{N}_{N}(\vz; \widehat{\bm{\mu}}_{k}, \widehat{\bm{\Sigma}}_{k}) \right)$.

The overall network is trained by Eqs.~(\ref{cost1}) and (\ref{getparams}). In this experiment, we set $D(\cdot \ , \ \cdot)$ as the square of the Euclidean distance because the input data is generated obeying the PDF in the Euclidean space. 
We test two types of $h(\cdot)$, $h(d)=d$ and $h(d)=\log(d)$, and denote models trained with these $h(\cdot)$ as RaDOGAGA(d) and RaDOGAGA(log(d)) respectively. 
We used DAGMM as a baseline method. DAGMM also consists of an encoder, decoder, and EN. In DAGMM, to avoid falling into the trivial solution that the entropy is minimized when the diagonal component of the covariance matrix is 0,
the inverse of the diagonal component $P(\widehat{\bm{\Sigma}})=\sum_{k=1}^K\sum_{i=1}^N \widehat{\bm{\Sigma}}_{ki}^{-1}$ is added to the cost: 
\begin{align}
\label{loss_dagmm}
L = \|{\bm x} - {\hat {\bm x}}\|_2^2 + \lambda_1 (- \log ({P_{\vz, \psi}({\bm z})}) ) + \lambda_{2}P(\widehat{\bm{\Sigma}}).
\end{align} 
The only differences between our model and DAGMM is 
that the regulation term $P(\widehat{\bm{\Sigma}})$ is replaced by $D(\hat {\bm x}, \breve{\bm x})$. 
The model complexity such as the number of parameters is the same. 
For both RaDOGAGA and DAGMM, the autoencoder part is constructed with FC layers with sizes of 64, 32, 16, 3, 16, 32, and 64. For all FC layers except for the last of the encoder and the decoder, we 
use $tanh$ as the activation function.
The EN part is also constructed with FC layers with sizes of 10 and 3. For the first layer, we 
use $tanh$ as the activation function and dropout (ratio=0.5). For the last one, $softmax$ is 
used. ($\lambda_{1}$, $\lambda_{2}$) is set as ($1\times10^{-4}$, $1\times10^{-9}$), ($1\times10^{6}$, $1\times10^{3}$) and ($1\times10^{3}$, $1\times10^{3}$) for DAGMM, RaDOGAGA(d) and RaDOGAGA(log(d)) respectively. We optimize all models by 
Adam 
optimizer 
with a learning rate of $1\times10^{-4}$. We set $\sigma^{2}$ as $1/12$.
\subsubsection{Results}
\begin{figure}[t]
 \begin{minipage}[b]{0.49\hsize}
  \begin{center}
  \includegraphics[keepaspectratio, scale=0.17]
  {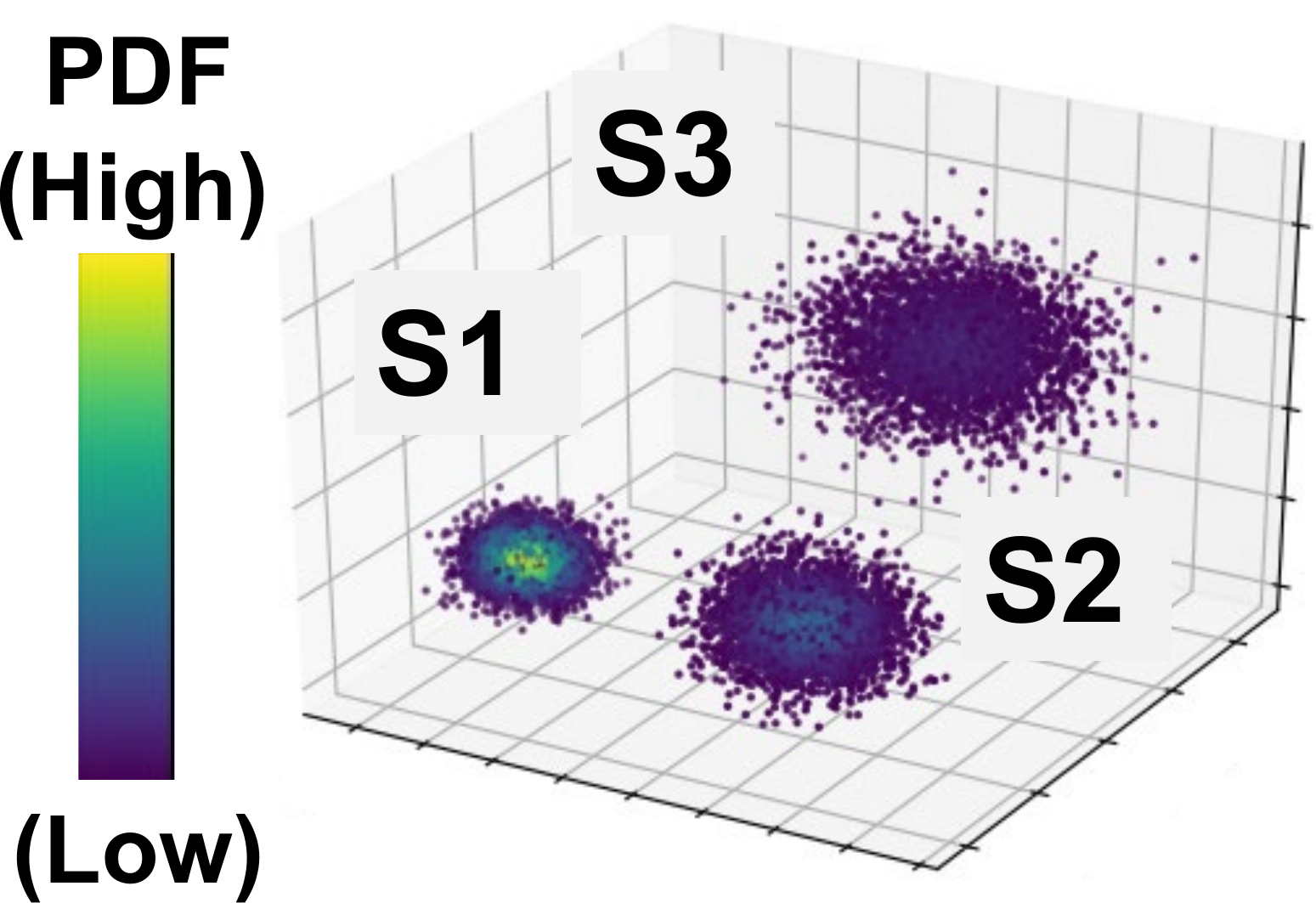}
  \end{center}
  \subcaption{Input source $\vs$}  
  \label{fig:vs}
 \end{minipage}
 \begin{minipage}[b]{0.49\linewidth}
  \centering
  \includegraphics[keepaspectratio, scale=0.15]
  {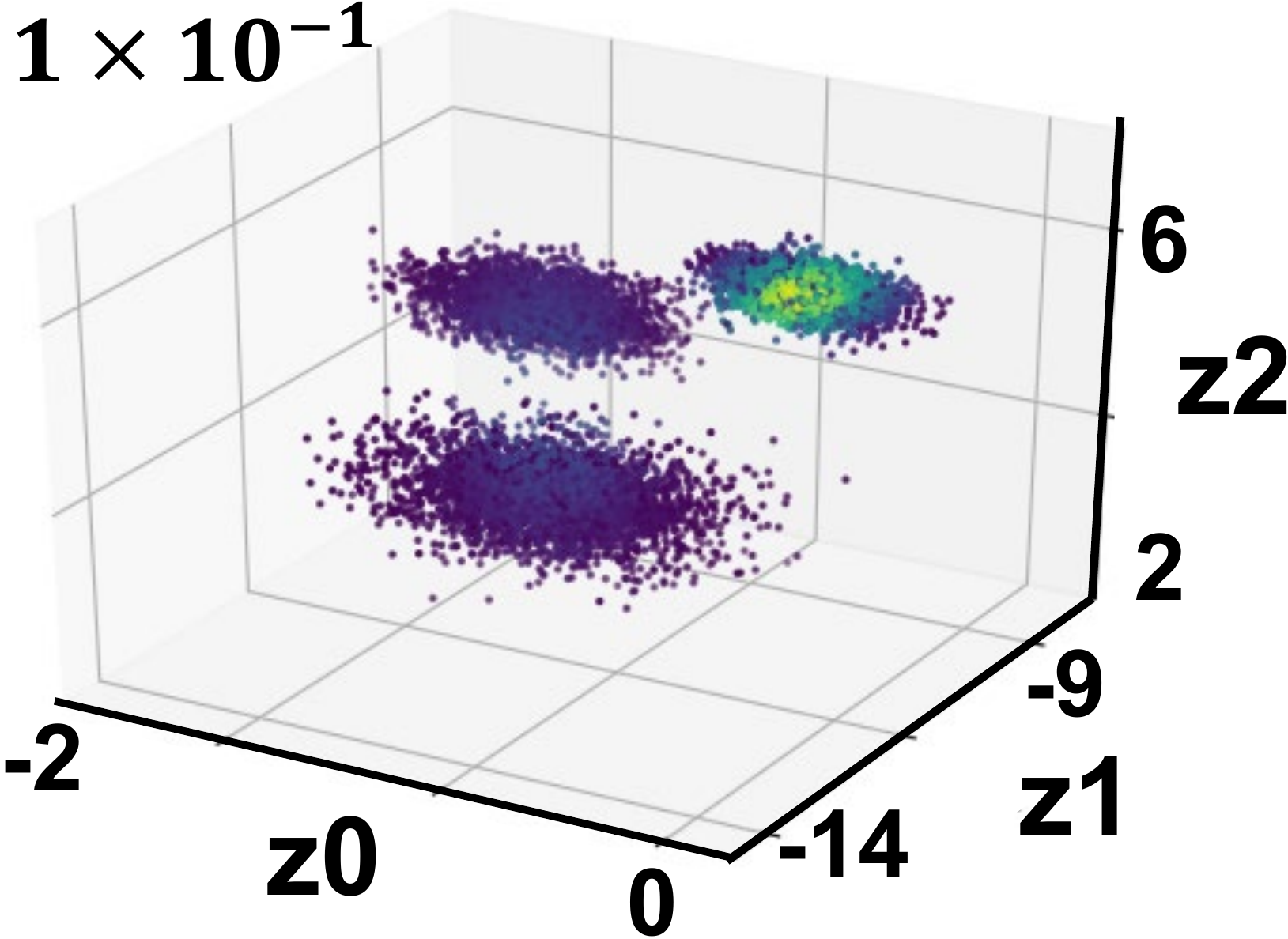}
  \subcaption{DAGMM}\label{z_dagmm}
 \end{minipage}
 \begin{minipage}[b]{0.49\linewidth}
  \centering
  \includegraphics[keepaspectratio, scale=0.15]
  {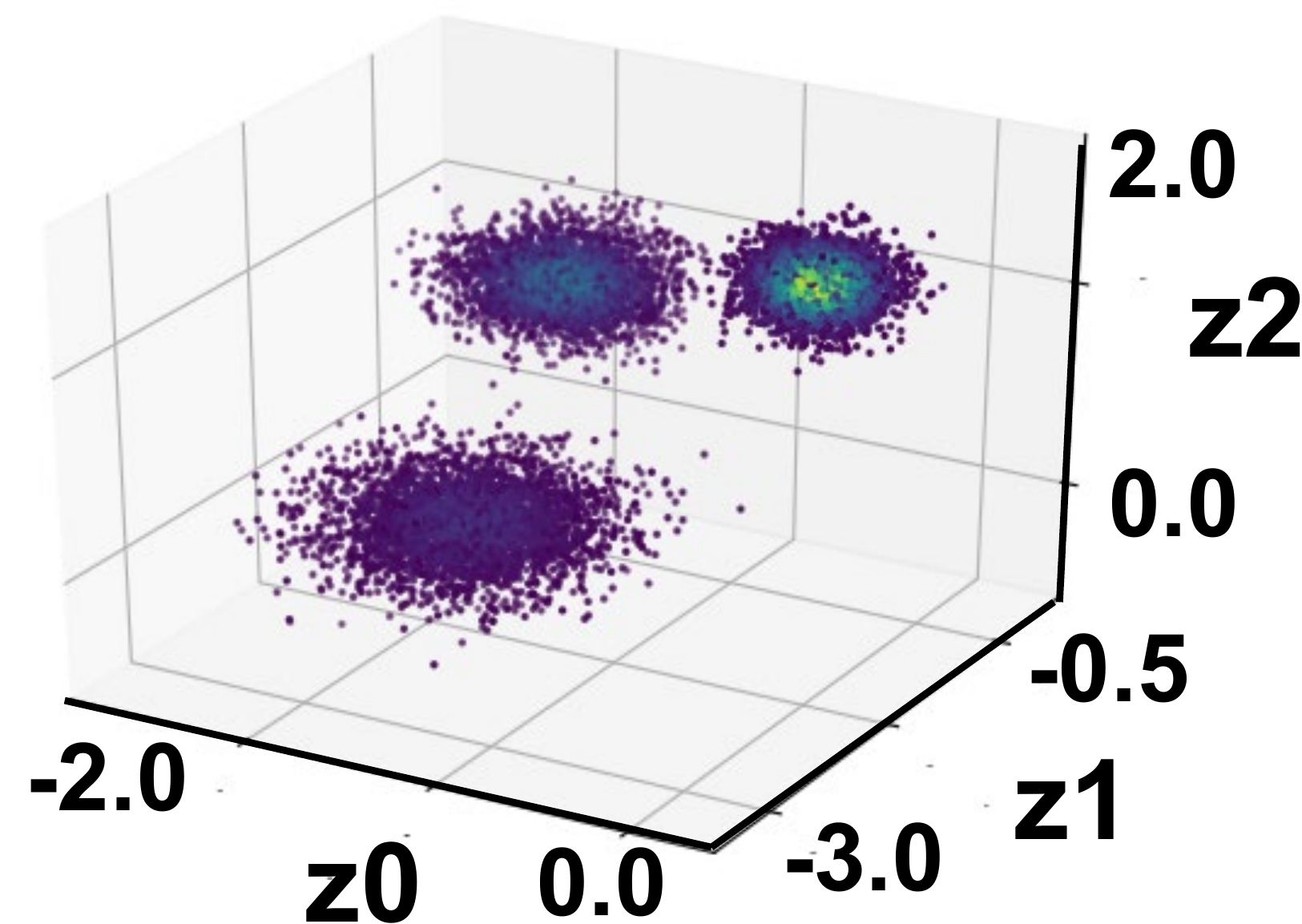}
  \subcaption{RaDOGAGA(d)}\label{z_RaDOGAGA_l2}
 \end{minipage}
 \begin{minipage}[b]{0.49\linewidth}
  \centering
  \includegraphics[keepaspectratio, scale=0.15]
  {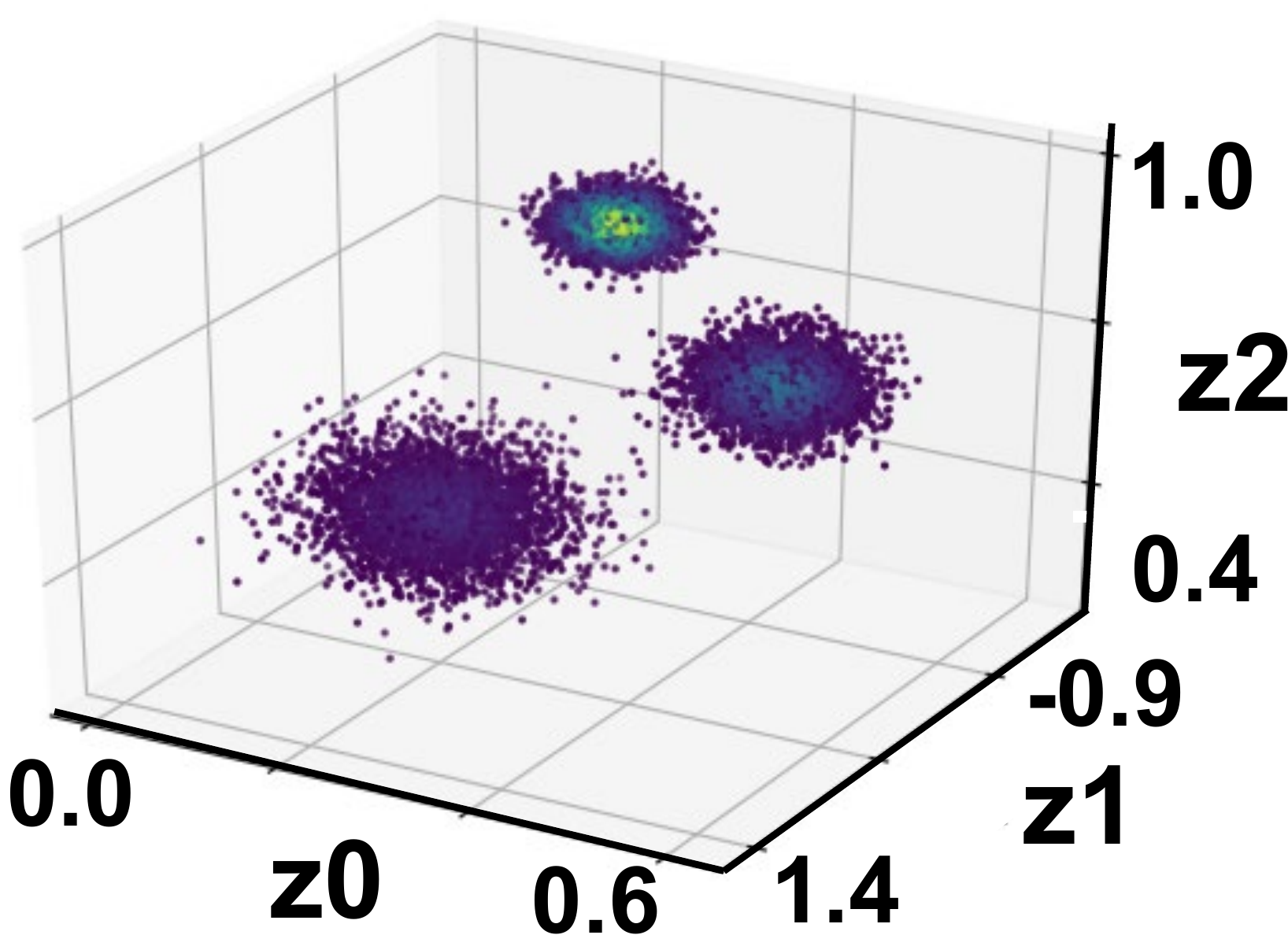}
  \subcaption{RaDOGAGA(log(d))}\label{z_RaDOGAGA_log}
 \end{minipage}
 \caption{Plot of the source of input data $\vs$ and latent variables $\vz$. The color bar located left of (a) corresponds to the normalized PDF.  Both DAGMM and RaDOGAGA capture three mixture-components, but the PDF in DAGMM looks different from the input data source. Points with high PDF do not concentrate on the center of the cluster especially in the upper right one. }\label{z_plot}
\end{figure}

\begin{figure}[h]
 \begin{minipage}[b]{0.32\linewidth}
  \centering
  \includegraphics[keepaspectratio, scale=0.13]
  {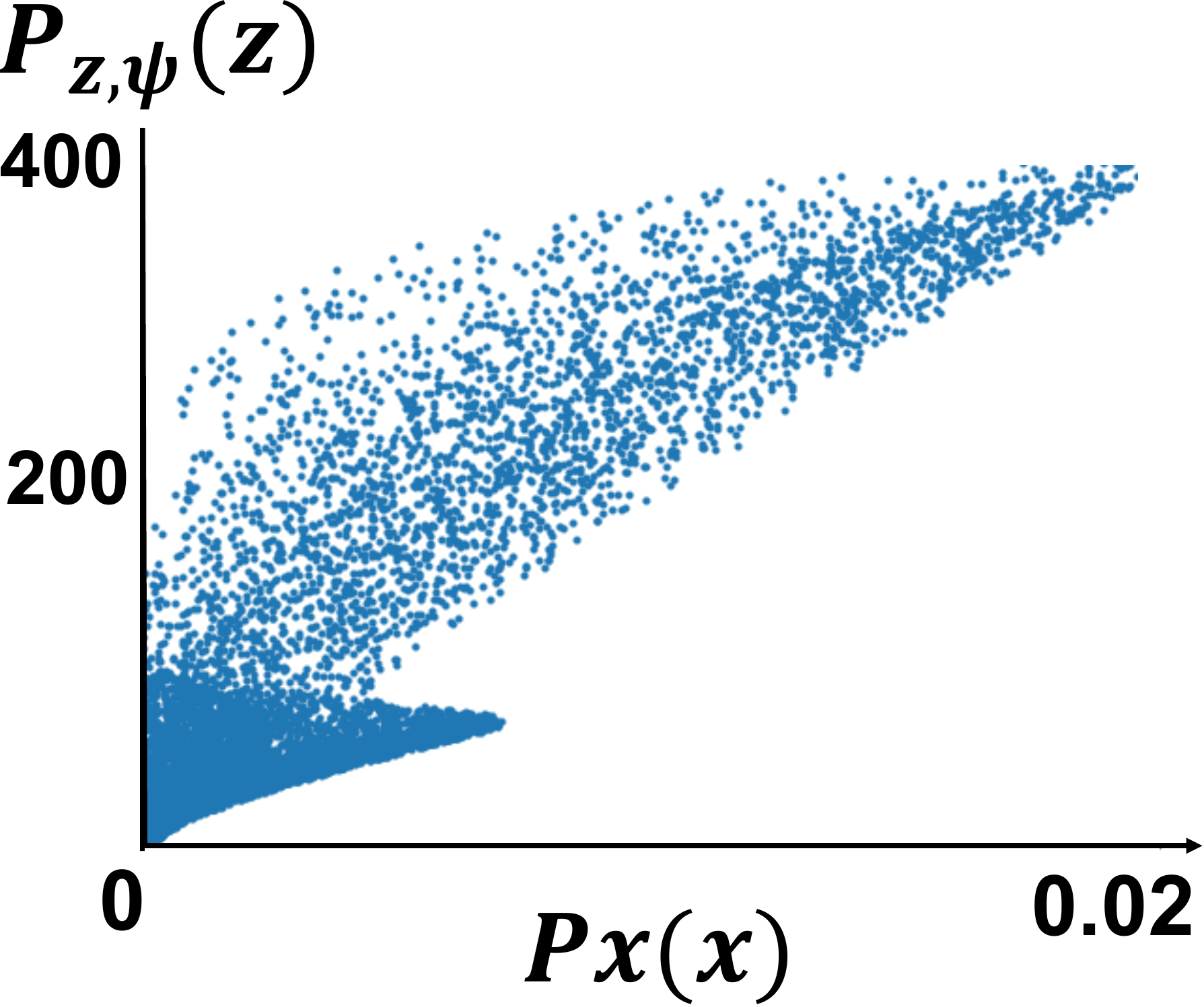}
  \subcaption{DAGMM}\label{pdf_dagmm}
 \end{minipage}
 \begin{minipage}[b]{0.32\linewidth}
  \centering
  \includegraphics[keepaspectratio, scale=0.13]
  {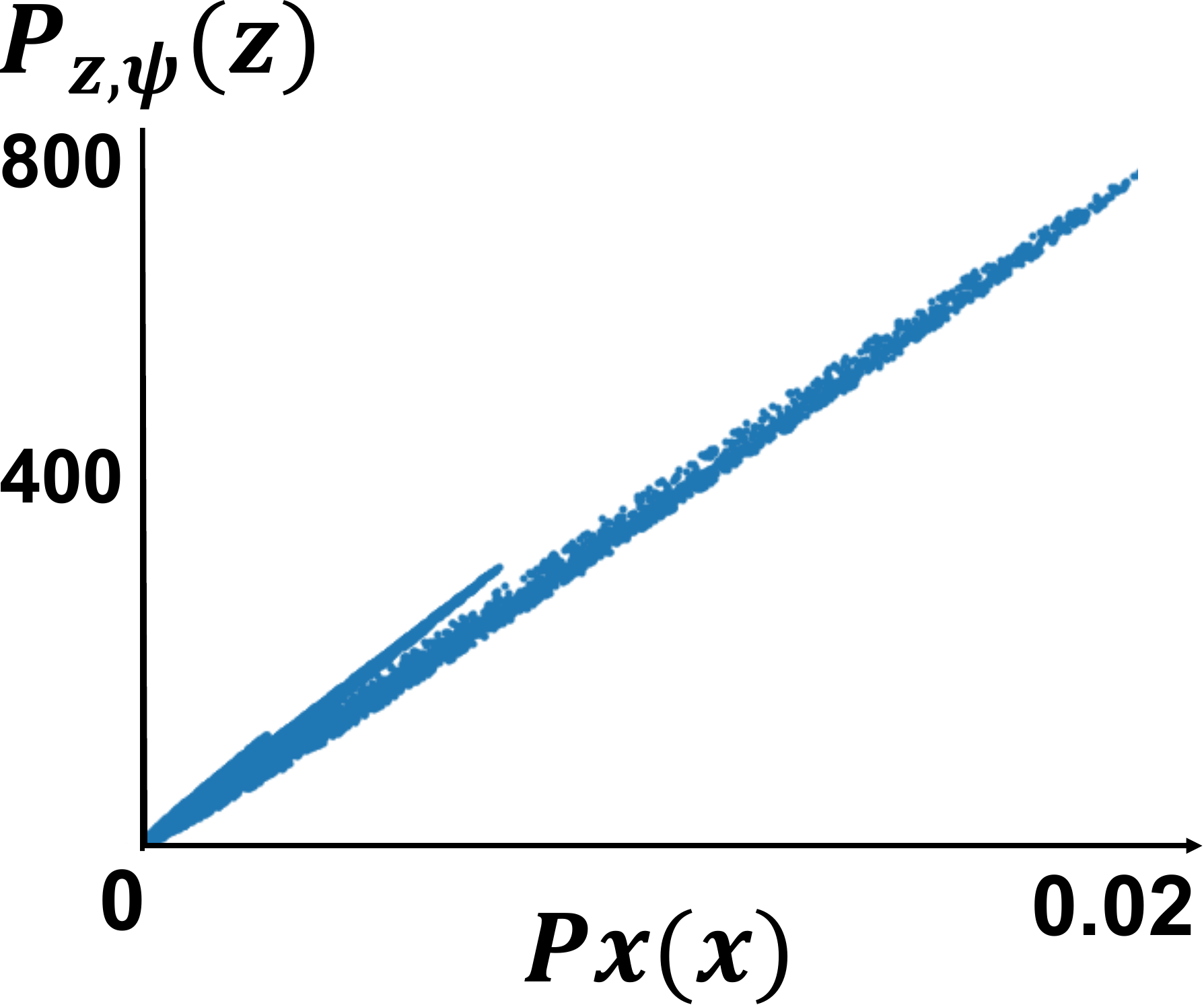}
  \subcaption{Ours (d)}\label{pdf_RaDOGAGA_l2}
 \end{minipage}
 \begin{minipage}[b]{0.32\linewidth}
  \centering
  \includegraphics[keepaspectratio, scale=0.13]
  {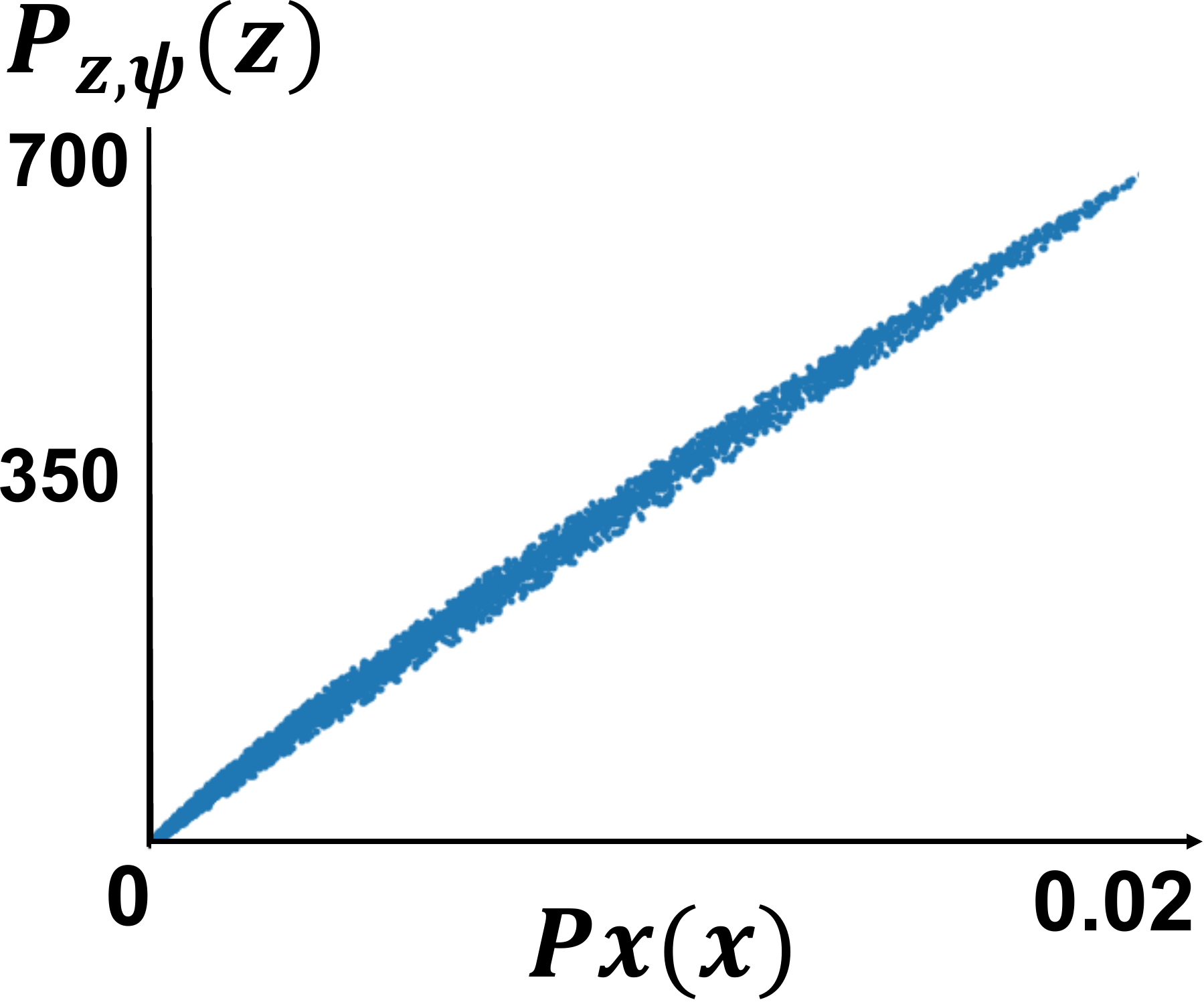}
  \subcaption{Ours (log(d))}\label{pdf_RaDOGAGA_log}
 \end{minipage}
 \caption{Plot of $P_{\vx}(\vx)$ vs $P_{\vz, \psi}(\vz)$. In RaDOGAGA, $P_{\vx}(\vx)$ and $P_{\vz, \psi}(\vz)$ are proportional while we cannot see that in DAGMM.}\label{z_pdf}
\end{figure}

\mred Figure \ref{z_plot} displays the distribution of the input data source $\vs$ and latent variable $\vz$. Although both methods can capture that $\vs$ is generated from three mixture-components, there is a difference in the PDFs. Since the data is generated from GMM, the value of the PDF gets higher as the sample gets closer to the centers of clusters. 
However, in DAGMM, this tendency looks distorted. 
This difference is further demonstrated in Fig.~\ref{z_pdf}, which shows a plot of $P_{\vx}(\vx)$ (horizontal axis) against $P_{\vz, \psi}(\vz)$ (vertical axis). \mblk In our method, $P_{\vx}(\vx)$ and $P_{\vz, \psi}(\vz)$ are almost proportional to each other as in the theory, 
but we cannot observe such a proportionality in DAGMM. 
This difference is also quantitatively obvious. That is, correlation coefficients between $P_{\vx}(\vx)$ and $P_{\vz, \psi}(\vz)$ are 0.882 (DAGMM), 0.997 (RaDOGAGA(d)), and 0.998 (RaDOGAGA(log(d))). 
We can also observe that, in RaDOGAGA(d), there is a slight distortion in its proportionality in the area of $P_{\vx}(\vx) < 0.01$.
When $P_{\vz, \psi}(\vz)$ is sufficiently fitted, $h(d)=\log(d)$ makes $P_{\vx}(\vx)$ and $P_{\vz, \psi}(\vz)$ be proportional more rigidly. 
More details 
are given in Appendix \ref{sec:hd}. 
%
%
%
%
%
%
%
%
\subsection{Anomaly Detection Using Real Data}\label{exp_ano}
\renewcommand{\thefootnote}{\fnsymbol{footnote}}
\renewcommand{\thempfootnote}{\fnsymbol{mpfootnote}}
\begin{table*}[t]
\begin{center}
\renewcommand{\footnoterule}{\empty}
\caption{Average and standard deviations (in brackets) of Precision, Recall and F1}\label{tab:anomaly} 
\begin{minipage}{\textwidth}
\begin{center}
\begin{tabular}{c|l|lll}
\multicolumn{1}{l|}{Dataset} & Methods       & Precision      & Recall         & F1             \\ \hline
\multirow{7}{*}{KDDCup}  & ALAD\footnotemark[1]   & 0.9427 (0.0018) & 0.9577 (0.0018) & 0.9501 (0.0018) \\
                             & INRF\footnotemark[1] & 0.9452 (0.0105) & 0.9600 (0.0113) & 0.9525 (0.0108) \\
                             & GMVAE\footnotemark[1]         & 0.952          & 0.9141         & 0.9326         \\
                             & DAGMM         & 0.9427 (0.0052) & 0.9575 (0.0053) & 0.9500 (0.0052) \\
                             & RaDOGAGA(d)    & 0.9550 (0.0037) & 0.9700 (0.0038) & 0.9624 (0.0038) \\
                             & RaDOGAGA(log(d))   & \bf{0.9563 (0.0042)} & \bf{0.9714 (0.0042)} & \bf{0.9638 (0.0042)} \\ \hline
\multirow{5}{*}{Thyroid}     & GMVAE\footnotemark[1]         &  \bf{0.7105}         & 0.5745         & 0.6353         \\
                             & DAGMM        & 0.4656 (0.0481) & 0.4859 (0.0502) & 0.4755 (0.0491) \\
                             & RaDOGAGA(d)    & 0.6313 (0.0476) & 0.6587 (0.0496) & 0.6447 (0.0486) \\
                             & RaDOGAGA(log(d))   &  0.6562 (0.0572) &  \bf{0.6848 (0.0597)} &  \bf{0.6702 (0.0585)} \\ \hline
\multirow{6}{*}{Arrythmia}   & ALAD\footnotemark[1]          & 0.5000 (0.0208) & 0.5313 (0.0221) & 0.5152 (0.0214) \\
                             & GMVAE\footnotemark[1]         & 0.4375         & 0.4242         & 0.4308         \\
                             & DAGMM        & 0.4985 (0.0389) & 0.5136 (0.0401) & 0.5060 (0.0395) \\
                             & RaDOGAGA(d)    &  \bf{0.5353 (0.0461)} &  \bf{0.5515 (0.0475)} &  \bf{0.5433 (0.0468)} \\
                             & RaDOGAGA(log(d))   & 0.5294 (0.0405) & 0.5455 (0.0418) & 0.5373 (0.0411) \\ \hline
\multirow{3}{*}{KDDCup-rev}  
                             & DAGMM       & 0.9778 (0.0018) & 0.9779 (0.0017) & 0.9779 (0.0018) \\
                             & RaDOGAGA(d)    & 0.9768 (0.0033) & 0.9827 (0.0012) & 0.9797 (0.0015) \\
                             & RaDOGAGA(log(d))   &  \bf{0.9864 (0.0009)} &  \bf{0.9865 (0.0009)} &  \bf{0.9865 (0.0009)}\\ \hline
\end{tabular}
   \footnotetext[1]{Scores are cited from \citet{ALAD} (ALAD), \citet{inclusive} (INRF), and \citet{GMVAE} (GMVAE).}
\end{center}
\end{minipage}
\end{center}
\end{table*}
We here examine whether the clear relationship between $P_{\vx}(\vx)$ and $P_{\vz, \psi}(\vz)$ is useful in anomaly detection in which PDF estimation is the key issue. 
We use four public datasets\footnote[3]{Datasets can be downloaded at \url{https://kdd.ics.uci.edu/} and \url{http://odds.cs.stonybrook.edu}.}: KDDCUP99, Thyroid, Arrhythmia, and KDDCUP-Rev. The (instance number, dimension, anomaly ratio(\%)) of each dataset is (494021, 121, 20),  (3772, 6, 2.5),  (452, 274, 15), and (121597, 121, 20). The details of the datasets are given in Appendix \ref{app_ano}.
\subsubsection{EXPERIMENTAL SETUP}
For a fair comparison with previous works, we follow the setting in \citet{DAGMM}. 
Randomly extracted 50\% of the data were assigned to the training and the rest to the testing. During the training, only normal data were used. During the test, the entropy $R$ for each sample was calculated as the anomaly score, and if the anomaly score is higher than a threshold, it is detected as an anomaly. The threshold is determined by the ratio of the anomaly data in each data set. For example, in KDDCup99, data with $R$ in the top 20 \% is detected as an anomaly. 
As metrics, precision, recall, and F1 score are calculated. We run experiments 20 times for each dataset split by 20 different random seeds. 
%
%
\subsubsection{BASELINE METHODS}
As in the previous section, DAGMM is taken as a baseline.
We also compare the scores of our method with the 
ones reported in previous works conducting 
the same experiments \citep{ALAD, inclusive, GMVAE}.
\subsubsection{CONFIGURATION}
As in \citet{DAGMM}, in addition to the output from the encoder, $\frac{\parallel \vx- \vx' \parallel_{2}}{\parallel \vx \parallel_{2}}$ and $\frac{ \vx \cdot \vx'}{\parallel \vx \parallel_{2}\parallel \vx' \parallel_{2}}$ are concatenated to $\vz$ and sent to EN. Note that $\vz$ is sent to the decoder before concatenation. Other configuration a except for the hyperparameter is same as in the previous experiment. The hyperparameter for each dataset is described in Appendix \ref{app_ano}. The input data are max-min normalized. 
\subsubsection{Results}
Table \ref{tab:anomaly} reports the average scores and standard deviations (in brackets).
Compared to DAGMM, RaDOGAGA has a better performance regardless of types of $h(\cdot)$. 
\mred Note that, our method does not increase model complexity at all. \mblk
Simply introducing the RDO mechanism to the autoencoder is effective for anomaly detection.
Moreover,  
RaDOGAGA achieves the highest performance compared to other 
methods.
%
RaDOGAGA(log(d)) is superior to RaDOGAGA(d) except for Arrhythmia. This result suggests that a much rigid orthonormality can likely bring better performance.

\section{Conclusion}
In this paper, we propose RaDOGAGA which embeds data in a low-dimensional Euclidean space isometrically. 
With RaDOGAGA, the relation of latent variables and data is quantitatively tractable. 
For instance, $P_{\vz, \psi}(\vz)$ obtained by the proposed method is related to $P_{\vx}(\vx)$ in a clear form, e.g., they are proportional when $\mA(\vx)=\mI$. 
Furthermore, thanks to these properties, we achieve a state-of-the-art performance in anomaly detection. 

Although we focused on the PDF estimation as a practical task in this paper, the properties of RaDOGAGA will benefit a variety of applications. 
For instance, data interpolation will be easier because a straight line in the latent space is geodesic in the data space. 
It also may help the unsupervised or semi-supervised learning since the distance of $\vz$ reliably reflects the distance of $\vx$. 
Furthermore, our method will promote disentanglement because, thanks to the orthonormality, PCA-like analysis is possible. 

To capture the essential features of data, 
it is important to fairly evaluate the significance of latent variables.
Because isometric embedding 
ensures this fairness, 
we believe that RaDOGAGA will bring a \emph{Breakthru} for generative analysis. 
As a future work, we explore the usefulness of this method in various tasks mentioned above. 
\section*{Acknowledgement}
We express our gratitude to Naoki Hamada, Ramya Srinivasan,  Kentaro Takemoto, Moyuru Yamada, Tomoya Iwakura, and Hiyori Yoshikawa for constructive discussion. 
\bibliography{example_paper}
\bibliographystyle{icml2020}
\appendix
\clearpage

\onecolumn
\section{How Jacobian Matrix Forms a Constantly Scaled Orthonormal System}
\label{OrthoJacobian}

In this appendix, 
we derive equations corresponding to  Eqs.~(\ref{Ortho}) and (\ref{isometricity}) 
for the case of $M > N$. 
The guiding principle of derivation is the same as in Section \ref{TheoreticalProps}: examining the condition to minimize the expected loss. 
As in Section \ref{TheoreticalProps}, we assume that the encoder and the decoder are trained enough in terms of reconstruction error so that $\vx \simeq \hat{\vx}$ holds 
and the second term
$ \lambda_{1} h \left(D \left(\vx, \hat {\vx} \right)\right)$ 
in  Eq.~(\ref{cost1}) can be ignored.

We assume that the Jacobian matrix $\mJ(\vz) = \partial \vx / \partial  \vz =  \partial g_{\phi}(\vz) / \partial  \vz\in \sR^{M \times N}$ 
is full-rank at every point $\vz \in \sR^{N}$ as in Section \ref{TheoreticalProps}.
Based on Eq.~(\ref{dfn_brevex}), Eq.~(\ref{dfn_hatx}) and Taylor expansion,  the difference $\breve{\vx}-\hat{\vx}$ can be approximated by 
$\acute{\bm \epsilon} =  \sum_{i=1}^N   \epsilon_i (\partial \vx/\partial z_{i}) \in \sR^{M}$.   
As in Section \ref{TheoreticalProps}, the expectation of the third term in Eq.~(\ref{cost1}) is re-written as follows: 
\begin{eqnarray}
\underset{{}^{{{\bm \epsilon} \sim P_{\bm \epsilon}(\bm \epsilon)}}}{E} 
\left[\acute{{\bm \epsilon}}^{\top}  \mA (\bm x)  \acute{{\bm \epsilon}}\right]
= \sigma^2  \sum_{j=1}^{N} \left( \frac{\partial \vx}{\partial z_{j}} \right)^{\top} \mA (\bm x) \left(\frac{\partial \vx}{\partial z_{j}} \right).  \label{cost2_2_2}
\end{eqnarray}
This equation has the same form as Eq.~(\ref{cost2_2}) except the differences in dimensions: $\partial \vx/ \partial z_{j}  \in \sR^{M}$ and $\mA (\bm x)  \in \sR^{M \times M}$ in Eq.~(\ref{cost2_2_2}) while $\partial \vx/ \partial z_{j}  \in \sR^{N}$ and $\mA (\bm x)  \in \sR^{N \times N}$ in Eq.~(\ref{cost2_2}). We have essentially no difference from Section \ref{TheoreticalProps} so far. 

However, from this point, we cannot follow the same way we used in Section \ref{TheoreticalProps} to derive the equation corresponding to Eq.~(\ref{Ortho}), due to the mismatch of $M$ and $N$. 
Yet, as we show below, step-by-step modifications lead us to the same conclusion.

Firstly, note that we can always regard $g_{\phi}$ as a composition function by inserting a smooth invertible function $\rho: \R^{N} \to \R^{N}$ and its inverse as follows: 
\begin{align}
g_{\phi}(\vz) = g_{\phi}( \rho^{-1} (\rho (\vz)))= \tilde{g}_{\phi}( \rho(\vz) ).
\end{align}

Let $\vy \in \sR^{N}$ be an auxiliary variable defined by 
$\vy = \rho (\vz)$.  
Due to the chain rule of differentiation, $\partial \vx / \partial \vz$ can be represented as 
\begin{align}
\frac{\partial \vx}{\partial \vz} = \frac{\partial \vx}{\partial \vy}  \frac{\partial \vy}{\partial \vz} = \mG \mB, 
\end{align}
where we define $\mG$ and $\mB$ as $\mG = \partial \vx / \partial \vy \in \sR^{M \times N}, \mB = (\vb_{1}, \dots, \vb_{N}) = \partial \vz / \partial \vy \in \sR^{N \times N}$.

Since $\vz$ and $\vy$ have the same dimension $N$, the relationship between $P_{\vz}(\vz)$ and $P_{\vy}(\vy)$ is described by $|\det(\mB)|$ (the absolute value of Jacobian determinant), which corresponds to 
the volume change under the function $\rho$, as follows: 
\begin{align}
P_{\vz}(\vz) = |\det(\mB)| P_{\vy}(\vy).
\end{align} 
Thus the expectation of $L$ in Eq.~(\ref{cost1}) can be  approximated as follows: 
\begin{align}
\label{cost2_f_2}
\underset{{}^{{{\bm \epsilon} \sim P_{\bm \epsilon}(\bm \epsilon)}}}{E} 
 \left[L \right]
\simeq - \log( | \det( \mB)| )-\log(P_{\vy}(\vy))) 
+ \lambda_2 \sigma ^2  \left( \sum_{j=1}^{N}  \left(\mG \vb_{j}  \right)^\top  \bm A (\bm x) \left(\mG \vb_{j}  \right) \right). 
\end{align}

To derive the minimization condition of the expected loss, we need further preparations. Let $\tilde{b}_{ij}$ denote the cofactor of matrix $\mB$ with regard to the element $b_{ij}$. We define a vector $\tilde{\vb}_{j}$ as follows:
\begin{align}
\tilde{\vb}_{j} = 
\left( 
\begin{array}{c}
\tilde{b}_{1j} \\
\tilde{b}_{2j} \\
\vdots \\
\tilde{b}_{Nj}
\end{array}
\right).
\end{align}
The following equation is a property of the cofactor~\cite{Cofactor}: 
\begin{align}
\vb_{i}^{\top} \tilde{\vb}_{j} = \sum_{k=1}^{N} b_{ki} \tilde{b}_{kj} = \delta_{ij} \det(\mB). \label{Cofact1}
\end{align}
In addition, since $| \det (\mB) | = (\det(\mB)^{2})^{\frac{1}{2}} = ( (  \sum_{k=1}^{N} b_{kj} \tilde{b}_{kj} )^{2} )^{\frac{1}{2}}$, we have the following result:
\begin{align}
\frac{\partial |\det(\mB)|}{\partial b_{ij}} &= \frac{1}{2} ( (\sum_{k=1}^{N} b_{kj} \tilde{b}_{kj})^{2} )^{-\frac{1}{2}} \cdot 2 (\sum_{k=1}^{N} b_{kj} \tilde{b}_{kj}) \tilde{b}_{ij} 
= \frac{\det(\mB)}{|\det(\mB)|} \tilde{b}_{ij}. 
\end{align}
Therefore, the following equations hold:
\begin{align}
\frac{\partial \log ( |\det(\mB)|)}{\partial b_{ij}}  &= \frac{1}{|\det(\mB)|} \frac{\partial |\det(\mB)|}{\partial b_{ij}} = \frac{1}{|\det(\mB)|} \frac{\det(\mB)}{|\det(\mB)|} \tilde{b}_{ij} = \frac{\det(\mB)}{\det(\mB)^{2}}  \tilde{b}_{ij} = \frac{1}{\det(\mB)} \tilde{b}_{ij},  \\
\frac{\partial \log ( |\det(\mB)|)}{\partial \vb_{j}} &= \frac{1}{\det(\mB)} \tilde{\vb}_{j}. \label{Cofact2}
\end{align}

By differentiating the right hand side of Eq.~(\ref{cost2_f_2}) by $\vb_{j}$ and setting the result to zero, 
the following equation is derived as a condition to minimize the expected  loss:
\begin{align}
\label{difcost_s_mn}
2 \lambda_2 {\sigma}^2  \mG^{\top} \mA(\vx)  \mG \vb_{j}   
=
\frac{1}{\det(\mB)} \tilde{\vb}_{j}.
\end{align}
Here we used Eq.~(\ref{Cofact2}). 
By multiplying $\vb_{i}^{\top}$ to this equation from the left and dividing the result by $2 \lambda_{2} \sigma^{2}$, we have
\begin{align}
\label{basis1}
\left(\mG \ \vb_i \right)^\top \mA(\vx)( \mG \ \vb_j)  & = \frac{1}{2 \lambda_2 {\sigma}^2} \frac{1}{\det(\mB)} \vb_i^\top \tilde{\vb}_j \\
& = \frac{1}{2 \lambda_2 {\sigma}^2} \delta_{ij},  \label{delta_for_Gb}
\end{align}
where the second line follows from Eq.~(\ref{Cofact1}). 
Since 
$\mG \vb_i = (\partial \vx / \partial \vy)(\partial \vy/ \partial z_{i}) = \partial \vx/ \partial z_{i}$ 
and   $\mG \vb_j = (\partial \vx / \partial \vy)(\partial \vy/ \partial z_{j}) = \partial \vx/ \partial z_{j}$, 
we can come back to the expression with the original variables $\vx$ and $\vz$ and reach the following conclusion: 
\begin{align}
\left({\frac{\partial \vx}{\partial z_i}}\right)^\top \bm A(\bm x) \left({\frac{\partial \bm x}{\partial z_j}}\right)
=
\frac{1}{2 \lambda_2 {\sigma}^2} \delta_{ij}. \label{Ortho_NonSquare}
\end{align}
Here the dimensions are different from Eq.~(\ref{Ortho}) ($\partial \vx/ \partial z_{i}, \partial \vx/ \partial z_{j} \in \sR^{M}$ and $\mA(\vx) \in \sR^{M \times M}$) but the meaning is same: 
the columns of the Jacobian matrix of two spaces $ \partial \vx / \partial z_{1}, \ldots ,\partial \vx / \partial z_{N}$ form a constantly-scaled orthonormal system  with respect to the inner product defined by $\mA(\vx)$ at every point.

Now we can derive the second conclusion in the exactly same manner as in Section \ref{TheoreticalProps}, although the dimensions are different ($\vv_{\vx}, \vw_{\vx} \in \sR^{M}$, $\mA(\vx) \in \sR^{M \times M}$ and $\vv_{\vz}, \vw_{\vz} \in \sR^{N}$):  
\begin{align}
\vv_{x}\mA(\vx)\vw_{x} &=\sum_{i=0}^N \sum_{j=0}^N \left(\frac{\partial \vx}{\partial z_{i}}v_{z{i}} \right)^\top \mA(\vx) \left(\frac{\partial \vx}{\partial z_{j}}w_{z{j}} \right) \nonumber \\
& =\frac{1}{2 \lambda_2 {\sigma}^2} \sum_{i=0}^{N}v_{z_{i}}w_{z_{i}} 
 =\frac{1}{2 \lambda_2 {\sigma}^2} \vv_{z} \cdot \vw_{z}, 
\end{align}
which means the map is isometric in the sense of Eq.~(\ref{iso_vx}).  
\section{Product of Singular Values as a Generalization of the Absolute Value of Jacobian Determinant } \label{app_prob}

In this appendix, we show the following two arguments we stated in the last part of Section \ref{TheoreticalProps}:  i) when a region in $\sR^{N}$ is mapped to $\sR^{M}$ by the decoder function, the ratio of the volume of the original region and its corresponding value is equal to the product of singular values of Jacobian matrix, ii) this quantity 
can be expressed by $\mA(\vx)$ under a certain condition. 
The Jacobian matrix $\mJ(\vz) = \partial \vx/\partial \vz = \partial g_{\phi}(\vz) / \partial  \vz \in \sR^{M \times N}$ is assumed to be full-rank as in Section \ref{TheoreticalProps} and Appendix \ref{OrthoJacobian}.

Let's consider the singular value decomposition $\mJ(\vz) = \mU(\vz) {\bm \Sigma }(\vz) \mV(\vz)^{\top}$, where $\mU(\vz) \in \sR^{M \times M}, {\bm \Sigma }(\vz) \in \sR^{M \times N}, \mV(\vz) \in \sR^{N \times N}$.
Note that $\{ \mV_{:,j}(\vz) \}_{j=1}^{N}$ is an orthonormal basis of  $\sR^{N}$ 
and  $\{ \mU_{:,j}(\vz) \}_{j=1}^{M}$ is an orthonormal basis of  $\sR^{M}$ 
with respect to the standard inner product. 

Consider an $N$-dimensional hypercube ${\mathbf c}$ specified by $\{ \varepsilon \mV_{:,j}(\vz) \}_{j=1}^{N} \ (\varepsilon > 0)$  
attached to $\vz \in \sR^{N}$.  
When $\varepsilon$ is small, 
the effect of the decoder function on $\{ \varepsilon \mV_{:,j}(\vz) \}_{j=1}^{N}$ is approximated by $\mJ(\vz) = \mU(\vz) {\bm \Sigma }(\vz) \mV(\vz)^{\top}$ 
and thus the mapped region of ${\mathbf c}$ in $\sR^{M}$ is approximated by a region $\tilde{\mathbf c}$ specified by 
$\{ \varepsilon \mJ(\vz) \mV_{:,j}(\vz) \}_{j=1}^{N} = \{ \epsilon s_{j}(\vz) \mU_{:,j}(\vz)  \}_{j=1}^{N}$, where $s_{1}( \vz) \ge \dots \ge s_{N} ( \vz) > 0$ are the singular values of  $\mJ(\vz)$ (remember full-rank assumption we posed).

Therefore, while the volume of the original hypercube ${\mathbf c}$ is $\varepsilon^{N}$,  the corresponding value of $\tilde{\mathbf c} \in \sR^{M}$ is $\varepsilon^{N} J_{sv}(\vz)$, where we define $J_{sv}(\vz)$ as $J_{sv}(\vz) = s_{1}(\vz) \cdots s_{N}(\vz)$, that is,  the product of the singular values of the Jacobian matrix $\mJ(\vz)$. 
This relationship holds for any $\vz \in \sR^{N}$ and we can take arbitrary small $\varepsilon$. 
Thus, the ratio of the volume of an arbitrary region in $\sR^{N}$ and its corresponding value in $\sR^{M}$ is also $J_{sv}(\vz)$\footnote{Consider covering the original region in $\sR^{N}$ by infinitesimal hypercubes.}. 

Let's move to the second argument. 
Note that Eq.~(\ref{Ortho_NonSquare}) can be rewritten in the following form since $\mJ(\vz) = (\partial \vx/\partial z_{1}, \ldots, \partial \vx/\partial z_{N})$:
\begin{align}
\mJ(\vz)^{\top} \mA(\vx) \mJ(\vz)  = \frac{1}{2 \lambda_2 {\sigma}^2} \mI_{N}. \label{JTAJ_M>N}
\end{align}

Let $0 < \alpha_{1}(\mA(\vx)) \le \dots \le \alpha_{N}(\mA(\vx)) \le \dots \le \alpha_{M}(\mA(\vx))$ be the eigenvalues of $\mA(\vx)$. If the condition 
\begin{align}
[\mO_{(M - N) \times N} \ \mI_{N}] \mU(\vz)^{\top} \mA(\vx) \mU(\vz) 
\left[
\begin{array}{c}
\mI_{N} \\
\mO_{(M - N) \times N}
\end{array}
\right] 
= \mO_{(M - N) \times N} \label{ConditionForM>N}
\end{align}
holds for all $\vz \in \sR^{N}$, the following relation holds for $J_{sv}(\vz)$:
\begin{align}
J_{sv}(\vz) =  \left( \frac{1}{2 \lambda_{2} \sigma^{2}} \right)^{\frac{N}{2}} \biggl( \alpha_{1}(\mA(\vx)) \cdots \alpha_{N}(\mA(\vx)) \biggr)^{-\frac{1}{2}}. \label{ResultsForM>N}
\end{align}
Here $\mO_{(M - N) \times N} \in \sR^{(M - N) \times N}$ denotes the matrix consisting of zeros.

To see this, let us first define $\mS(\vz) \in \sR^{N \times N}$ as $\mS(\vz) = \mathrm{diag}(s_{1}(\vz), \dots, s_{N}(\vz))$. 
Then $ \mJ(\vz) = \mU(\vz)^{\top}[ \mS(\vz) \ \mO_{(M - N) \times N}]^{\top} \mV(\vz)$. 
We obtain the following equation by substituting this expression of $\mJ(\vz)$ to Eq.~(\ref{JTAJ_M>N}):
\begin{align}
\mV(\vz) [ \mS(\vz) \ \mO_{(M - N) \times N} ] \mU(\vz)^{\top} \mA(\vx) \mU(\vz) 
\left[
\begin{array}{c}
\mS(\vz) \\
\mO_{(M - N) \times N}
\end{array}
\right] \mV(\vz)^{\top} = \frac{1}{2 \lambda_2 {\sigma}^2} \mI_{N}. \label{TransformedJTAJ}
\end{align}
Furthermore, we get the following equation by multiplying Eq.~(\ref{TransformedJTAJ}) by $\mS(\vz)^{-1} \mV(\vz)^{\top}$ from the left and $\mV(\vz) \mS(\vz)^{-1}$ from the right: 
\begin{align}
 [ \mI_{N} \ \mO_{(M - N) \times N} ] \mU(\vz)^{\top} \mA(\vx) \mU(\vz) 
\left[
\begin{array}{c}
\mI_{N} \\
\mO_{(M - N) \times N}
\end{array}
\right] = \frac{1}{2 \lambda_2 {\sigma}^2} 
\mS(\vz)^{-2}
\end{align}

This means $\mU(\vz)^{\top} \mA(\vx) \mU(\vz)$ has the following form: 
\begin{align}
\mU(\vz)^{\top} \mA(\vx) \mU(\vz)  
= 
\left[
\begin{array}{cc}
\frac{1}{2 \lambda_2 {\sigma}^2} 
\mS(\vz)^{-2} & \mC \\
\mC^{\top} & \mD
\end{array}
\right],  \label{ExpressionUTAU}
\end{align}
where $\mC \in \sR^{ N \times (M-N)} $ and $\mD \in \sR^{(M-N) \times (M-N)}$.  
Note that the standard basis vectors of $\sR^{N}$, namely, $\ve^{(1)} = [1 \cdots 0]^{\top}, \dots, \ve^{(N)} = [0 \cdots 1]^{\top}\in \sR^{N}$,   are the eigenvectors of $\frac{1}{2 \lambda_2 {\sigma}^2} \mS(\vz)^{-2}$ and corresponding eigenvalues are $ \frac{1}{2 \lambda_2 {\sigma}^2 s_{1}(\vz)^{2}}  < \dots < \frac{1}{2 \lambda_2 {\sigma}^2 s_{N}(\vz)^{2}}$.
According to the expression (\ref{ExpressionUTAU}), the condition (\ref{ConditionForM>N}) means $\mC^{\top} = \mO_{(M - N) \times N}
$, and thus $\mC^{\top} \ve^{(j)} = \mathbf{0}$ for all $j = 1, \dots N$ in this situation. 
Note also that the eigenvalues of $\mU(\vz)^{\top} \mA(\vx) \mU(\vz) $ coincide with those of $\mA(\vx)$. 
Therefore, if Eq.~(\ref{ConditionForM>N}) holds, we have 
\begin{align}
\alpha_{1}(\mA(\vx)) =  \frac{1}{2 \lambda_2 {\sigma}^2 s_{1}(\vz)^{2}}, \dots,  \alpha_{N}(\mA(\vx)) =  \frac{1}{2 \lambda_2 {\sigma}^2 s_{N}(\vz)^{2}}, \label{eigenvalues_singularvalues}
\end{align}
due to the inclusion principle \cite{InclusionPrinciple}.  
Eq.~(\ref{ResultsForM>N}) follows from Eq.~(\ref{eigenvalues_singularvalues}).


As mentioned before, when the metric function is square of L2 norm, $\mA(\vx)$ is the identity matrix $\mI_{M}$. 
In this case, Eq.~(\ref{ConditionForM>N}) holds and we have $J_{sv}(\vz) = (1/2 \lambda_{1} \sigma^{2})^{N/2}$\footnote{This can also be directly confirmed by taking determinants of Eq.~(\ref{JTAJ_M>N}) after substituting $\mA(\vx) = \mI_{M}$.}.
\section{Effect of $h(x)$}\label{sec:hd}
\mmmblu
In this section, the effects of $h(d)$ is discussed. 
By training the encoder and the decoder to be exact inverse functions of each other regarding the input data, the mapping becomes much rigidly isometric. 
Actually, for this purpose, it is important to choose $h(d)$ appropriately depending on metric function. 
\mblk

In this appendix we evaluate the behaviors of encoder and decoder in a one dimensional 
case using simple parametric linear encoder and decoder.
\begin{figure}[h]
	\centering
	\includegraphics[clip, width=10cm]{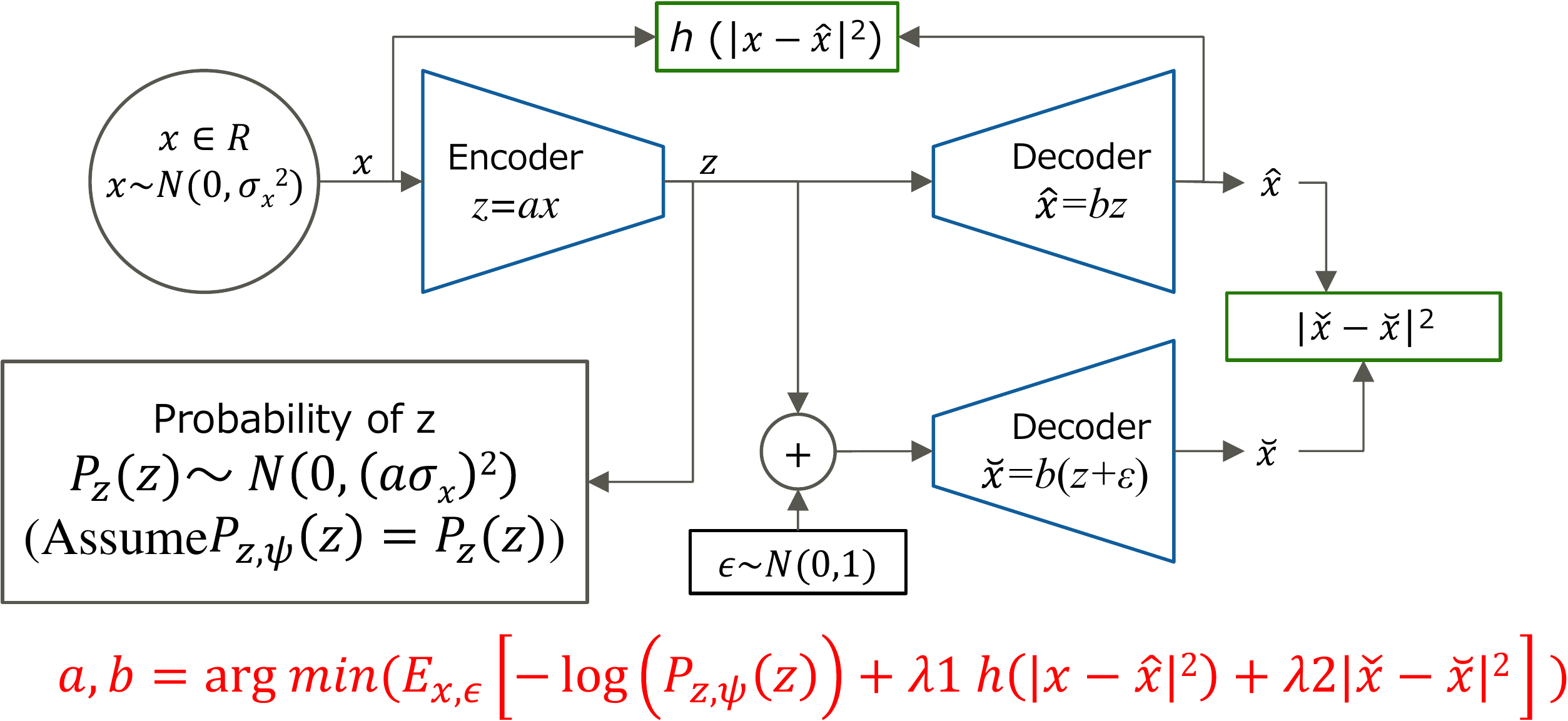}
	\caption{Simple encoder/decoder model to evaluate $h(d)$}
	\label{figSimpleArc}
\end{figure}
Lex $x$ be a one dimensional data with the normal distribution:
\begin{eqnarray}
\label{SimpleModel}
x &\in& \sR, \nonumber \\
x &\sim& \mathscr{N}(0, {\sigma_x}^2). \nonumber
\end{eqnarray}
Lex $z$ be a one dimensional latent variable. Following two linear encoder and decoder are provided with parameter $a$ and $b$:
\begin{eqnarray}
z &=& a  x, \nonumber \\
\hat x &=& b z. \nonumber
\end{eqnarray}
Due to the above relationship, we have 
\begin{align}
P_{z}(z) = \mathscr{N}(0,(a {\sigma_x})^2).
\end{align} 

\mmag Here, square error is used as metrics function $D(x,y)$. \mblk The distribution of noise $ \epsilon $ added to latent variable z is set to $N(0,1)$. Then $\breve x$ is derived by decoding $z + \epsilon $ as:
\begin{eqnarray}
D(x,y) &=& |x-y|^2, \nonumber \\
\epsilon &\sim& \mathscr{N}(0,1), \nonumber \\
\breve x &=& b  (z + \epsilon). \nonumber
\end{eqnarray}
For simplicity, we assume parametric PDF $P_{z,\psi} (z)$ is equal to the real PDF $P(z)$. Because the distribution of latent variable $z$ follows $N(0,(a {\sigma_x})^2)$, the entropy of $z$ can be expressed as follows: 
\begin{eqnarray}
H(z) &=& \int -P_{z}(z)  \log (P_{z}(z)) \mathrm{d} z \nonumber \\
&=& \log (a) + \log ({\sigma_x} \sqrt{2 \pi e}).
\end{eqnarray}
\mblk
Using these notations, Eqs. (\ref{cost1}) and (\ref{getparams}) can be expressed as follows:
\begin{flushleft}
\begin{eqnarray}
\label{SimpleCost1}
L &=& E_{{x}\sim \mathscr{N}(0, {\sigma_x}^2),\ {\epsilon} \sim \mathscr{N}(0,1)}  [-\log P_{z}(z) 
 + \lambda_1 h(|x - \hat x|^2) + \lambda_2  |\hat x, \breve x|^2  ] \nonumber \\ 
&=& \log (a) + \log ({\sigma_x} \sqrt{2 \pi e}) 
  + \lambda_1   E_{{x}\sim N(0, {\sigma_x}^2)} \left [ h(|x - \hat x|^2) \right] + \lambda_2   b^2.
\end{eqnarray}
\end{flushleft}
At first, the case of $h(d)=d$ is examined. By applying $h(d)=d$, Eq. (\ref{SimpleCost1}) can be expanded as follows:
\begin{eqnarray}
\label{SimpleCost2}
L = \log (a) + \log ({\sigma_x} \sqrt{2 \pi e}) + \lambda _1   (a   b - 1)^2   {\sigma _x}^2 + \lambda _2   b^2.
\end{eqnarray}
By solving $ \frac{\partial L}{\partial a} = 0$ and $ \frac{\partial L}{\partial b} = 0$, $a$ and $b$ are derived as follows:
\begin{eqnarray}
a   b &=& \frac{\lambda _1 {\sigma_x} ^2 + 
\sqrt{{\lambda_1}^2 {\sigma_x} ^4 - 2 \lambda_1 {\sigma_x} ^2}}{2 \lambda_1 \sigma_x ^2},  \nonumber \\
a &=& \sqrt {2   \lambda _2}   \left ( \frac{\lambda _1 {\sigma_x} ^2 + 
\sqrt{{\lambda_1}^2 {\sigma_x} ^4 - 2 \lambda_1 {\sigma_x} ^2}}{2 \lambda_1 \sigma_x ^2} \right ), \nonumber \\
b &=& 1 /\sqrt {2   \lambda _2}. \nonumber 
\end{eqnarray}
If $\lambda _1 {\sigma_x} ^2 \gg 1$, these equations are approximated as next:
\begin{eqnarray}
a   b & \simeq & \left (1 - \frac{1}{2 \lambda_1 {\sigma_x} ^2}\right ), \nonumber \\
a &=& {\sqrt {2   \lambda _2}}   \left (1 - \frac{1}{2 \lambda_1 {\sigma_x} ^2}\right ), \nonumber \\
b &=& \nonumber 1 / \sqrt {2   \lambda _2}.
\end{eqnarray}
Here, $a   b$ is not equal to $1$. That is, decoder is not an inverse function of encoder. In this case, the scale of latent space becomes slightly bent in order to minimize entropy function.  \mmag As a result, good fitting of parametric PDF $P_{z}(z) \simeq P_{z,\psi}(z)$ could be realized  while proportional relationship $P_{z}(z) \propto P_{x}(x)$ is relaxed. \mblk

Next, the case of $h(d)=\log (d)$ is examined. By applying $h(d)=\log (d)$ and introducing a minute variable $\Delta$, Eq. (\ref{SimpleCost1}) can be expanded as follows:
\begin{eqnarray}
\label{SimpleCost3}
L = \log (a) + \log ({\sigma_x} \sqrt{2 \pi e}) + \lambda _1   \log \left ( \left(a   b - 1 \right)^2 + \Delta \right) + \lambda _2   b^2.
\end{eqnarray}
By solving $ \frac{\partial Loss}{\partial a} = 0$ and $ \frac{\partial Loss}{\partial b} = 0$ and setting $\Delta \rightarrow 0$, $a$ and $b$ are derived as follows:
\begin{eqnarray}
a   b &=& 1, \nonumber \\
a &=& \sqrt {2   \lambda _2}, \\
b &=& \nonumber 1 /\sqrt {2   \lambda _2}
\end{eqnarray}
Here, $a   b$ is equal to $1$ and decoder becomes an inverse function of encoder regardless of the variance ${\sigma_x}^2$. In this case, good proportional relation $P_{z}(z) \propto P_{x}(x)$ could be realized regardless of the fitting $P_{z, \psi} (z)$ to $P_{z}(z)$.

Considering from these results, there could be a guideline to choose $h(d)$. If the parametric PDF $P_{\vz, \psi} (\vz)$ has enough ability to fit the real distribution $P_{\vz}(\vz)$, $h(d)=\log (d)$ could be better. If not, $h(d)=d$ could be an option.
\section{Expansion of SSIM and BCE to Quadratic Forms}
\label{SSIM_EXP}
In this appendix, it is shown that SSIM and BCE can be approximated in quadratic forms with positive definite matrices except some constants.

Structural similarity (SSIM) \citep{SSIM} is widely used for picture quality metric since it is close to human subjective evaluation. In this appendix, we show $(1-SSIM)$ can be approximated to a quadratic form such as Eq.(\ref{SecondaryForm}).

Eq. (\ref{SSIM_base}) is a SSIM value between cropped pictures $\bm x$ and $\bm y$ with a $W \times W$ window: 
\begin{equation}
\label{SSIM_base}
SSIM_{W \times W}(\bm x, \bm y) = \frac{2 \mu_x \mu_y}{{\mu_x}^2 + {\mu_y}^2}  \frac{2 {\sigma_{xy}}}{{\sigma _x}^2 + {\sigma _y}^2}.
\end{equation}
In order to calculate SSIM index for entire pictures, this window is shifted in a whole picture and all of SSIM values are averaged. 
If $\left (1 - SSIM_{W \times W} \left (\bm x, \bm y \right) \right)$ is expressed in quadratic form, the average for a picture $(1 - SSIM_{picture})$ can be also expressed in quadratic form.

Let $\Delta \bm x$ be a minute displacement of $\bm x$. Then SSIM between $\bm x$ and $\bm x + \Delta \bm x$ can be expressed as follows:
\begin{eqnarray}
\label{SSIM_Approx}
SSIM_{W \times W}(\bm x, \bm x + \Delta \bm x)
= 1-\frac{{\mu_{\Delta \bm x}}^2}{2{\mu_x}^2} - \frac{{\sigma_{\Delta \bm x}}^2}{2{\sigma _x}^2} + O \left( \left ( |\Delta \bm x|/|\bm x| \right) ^3 \right)
\end{eqnarray}

Then ${\mu_{\Delta \bm x}}^2$ and ${\sigma_{\Delta \bm x}}^2$ can be expressed as follows:
\mmred
\begin{align}
\label{MuDeltaX} 
{\mu_{\Delta \bm x}}^2 &= {\Delta \bm x}^\top \mW_{m} {\Delta \bm x},  \\
{\sigma_{\Delta \bm x}}^2 &= {\Delta \bm x}^\top \mW_{v}  {\Delta \bm x}, 
\end{align}
where 
\begin{eqnarray}
\bm W_{m} = \frac{1}{W^2} 
 \left(
  \begin{array}{cccc}
      1 & 1 & \ldots & 1 \\
      1 & 1 & \ldots & 1 \\
      \vdots & \vdots & \ddots & \vdots \\
      1 & 1 & \ldots & 1 \\
    \end{array}
  \right), 
\bm W_{v}
= \frac{1}{W^2} 
  \left(
    \begin{array}{cccc}
      W-1 & -1 & \ldots & -1 \\
      -1 & W-1 & \ldots & -1 \\
      \vdots & \vdots & \ddots & \vdots \\
      -1 & -1 & \ldots & W-1 \\
    \end{array}
  \right).  
\end{eqnarray}
\mblk
\mred
It should noted that matrix $\mW_{m}$ is positive definite and matrix $\mW_{v}$ is positive semidefinite. As a result, $\left (1 - SSIM_{W \times W} \left (\bm x, \bm y \right) \right)$ can be expressed in the following quadratic form with positive definite matrix: \mblk
\mmred
\begin{eqnarray}
\label{SSIM_Approx2}
1 - SSIM_{W \times W}(\bm x, \bm x + \Delta \bm x)
\simeq {\Delta \bm x}^\top \ \bm \left ( \frac{1}{2{\mu_x}^2} \mW_{m} + \frac{1}{2{\sigma_x}^2} \mW_{v} \right )  {\Delta \bm x}.
\end{eqnarray}
\mblk
%
Binary cross entropy (BCE) is also a reconstruction loss function widely used in VAE \citep{VAE}. BCE is defined as follows: 
\begin{equation}
\label{BCE}
BCE(\bm x, \bm y) = \sum_{i=1}^{M}( - x_i \log (y_i) -(1 - x_i ) \log(1- y_i) ).
\end{equation}
%

BCE can be also approximated by a quadratic form with positive definite matrix.  
Let $\Delta \bm x$ be a small displacement of $\bm x$ and $\Delta x_i$ be its $i$-th component.  Then BCE between $\bm x$ and $\bm x + \Delta \bm x$ can be expanded as follows:
%
\begin{align}
\label{BCE2}
BCE(\bm x, \bm x + \Delta \bm x)
&= \sum_{i}( - x_i \log (x_i + \Delta x_i) -(1 - x_i ) \log(1- x_i - \Delta x_i) ) \nonumber \\ 
&= \sum_{i}\left( - x_i \log \left(x_i \left(1 + \frac{ \Delta x_i}{x_i} \right) \right) 
     - \left(1 - x_i \right) \log \left( \left(1- x_i \right) \left(1 - \frac{\Delta x_i}{1- x_i} \right) \right) \right) \nonumber \\ 
&=  \sum_{i}\left( - x_i \log \left(1 + \frac{ \Delta x_i}{x_i} \right) -\left(1 - x_i \right) \log \left(1 - \frac{\Delta x_i}{1- x_i} \right) \right) \nonumber \\
&\ \ \ \ \ \ \ \ + \sum_{i}( - x_i \log (x_i) -(1 - x_i ) \log(1- x_i) ). 
\end{align}
%
Here, the second term of the last equation is constant depending on $\bm x$.  The first term of the last equation is further expanded as follows by using Maclaurin expansion of logarithm:
%
\begin{align}
\label{BCE3}
&\ \ \ \ \  \sum_{i} \Biggl(  - x_i \left(\frac{ \Delta x_i}{x_i}   -  \frac{{\Delta x_i}^2}{2 {x_i}^2} \right) 
 - \left(1 - x_i \right) \left( - \frac{\Delta x_i}{1- x_i} -  \frac{{\Delta x_i}^2}{2 \left(1- x_i \right)^2}\right) + O \left({\Delta x_i}^3 \right) \Biggr) \nonumber \\
&= \sum_{i} \left( \frac{1}{2} \left(\frac{1}{x_i}+\frac{1}{1-x_i} \right){{\Delta x_i}^2} +  O \left({\Delta x_i}^3 \right) \right). 
\end{align}
Then, let a matrix $\bm A(\bm x)$ be defined as follows:
%
\begin{align}\label{ax_bce}
\bm A(\bm x) =  
  \left(
    \begin{array}{ccc}
      \frac{1}{2}\left(\frac{1}{x_1}+\frac{1}{1-x_1}\right) & 0 & \ldots \\
      0 & \frac{1}{2}\left(\frac{1}{x_2}+\frac{1}{1-x_2}\right) & \ldots \\
      \vdots & \vdots & \ddots  \\
    \end{array}
  \right). \nonumber \\
\end{align} 
%
Obviously $\bm A(\bm x)$ is a positive definite matrix.  As a result, BCE between $\bm x$ and $\bm x + \Delta \bm x$ can be approximated by a quadratic form with $\bm x$ depending constant offset as follows: 
%
\begin{align}
BCE(\bm x, \bm x + \Delta \bm x)
\simeq \Delta \bm x^\top \bm A(\bm x) \Delta \bm x 
  + \sum_{i}( - x_i \log (x_i) -(1 - x_i ) \log(1- x_i) ).
\label{bce_tyl}
\end{align}
Note that BCE is typically used for binary data. In this case, the second term in Eq.~(\ref{bce_tyl}) is always 0. 
\section{``Continuous PCA" Feature of Isometric Embedding for Riemannian Manifold}
\label{EX_PCA}
In this section, we explain 
that the isometric embedding realized by 
RaDOGAGA has a continuous PCA feature 
when the following factorized probability density model is used:
\begin{equation}
\label{Fact1}
P_{\bm z, \psi}(\bm z) = \prod_{i=1}^N P_{zi, \psi}(z_i).
\end{equation}
Here, our definition of ``continuous PCA" is the following.  1) Mutual information between latent variables are minimum and likely to be uncorrelated to each other: 2) Energy of latent space is concentrated to several principal components, and the importance of each component can be determined: 3) These features are held for all subspace of a manifold and subspace is continuously connected. 

Next we explain the reason why these feature is acquired. %
As explained in Appendix \ref{OrthoJacobian}, all column vectors of Jacobian matrix of decoder from latent space to data space have the same norm and all combinations of pairwise vectors are orthogonal. In other words, when constant value is multiplied, the resulting vectors are orthonormal. Because encoder is a inverse function of decoder ideally, each row vector of encoder's Jacobian matrix should be the same as column vector of decoder under the ideal condition.  Here, $f_{ortho,\theta}(\bm x)$ and $g_{ortho, \phi}(\bm z_\theta)$ are defined as encoder and decoder with these feature. Because the latent variables depend on encoder parameter $\theta$, latent variable is described as $\bm z_\theta =f_{ortho,\theta}(\bm x)$, and its PDF is defined as $P_{\vz, \theta}(\bm z_\theta)$. 
PDFs of latent space and data space have the following relation where $J_{sv}(\vz_{\theta})$ is the product of the singular values of $\mJ(\vz_{\theta})$ which is a Jacobian matrix between two spaces as explained in Section \ref{TheoreticalProps} and Appendix \ref{app_prob}. 
\begin{equation}
\label{Prop21}
P_{\vz, \theta}(\bm z_\theta)  = J_{sv}(\vz_{\theta})  P_{\vx}(\bm x) \propto 
\biggl( \prod_{j=1}^{N} \alpha_{j}(\mA(\vx)) \biggr)^{-\frac{1}{2}} P_{\vx}(\vx).
\end{equation}
As described before, $P_{\vz, \psi}(\bm z)$ is a parametric PDF of the latent space to be optimized with parameter $\psi$.

By applying the result of 
Eqs.~(\ref{cost2_f_2}) and (\ref{basis1}), Eq. (\ref{cost1}) can be transformed as Eq. (\ref{cost5}) where $\hat {\bm x} = g_{ortho, \phi}(f_{ortho, \theta}(\bm x))$.
\mmred
\begin{eqnarray}
\label{cost5}
L_{ortho} = - \log \left({P_{\vz, \psi}({\bm z_\theta})}\right)+
\lambda_1  h\left(D({\bm x},\hat {\bm x}) \right) + N/2. \nonumber \\
s.t.\ \ {\left(\frac{\partial g_{ortho,\phi}(\bm z_\theta)}{\partial z_{\theta_i}} \right)}^\top \mA(\vx) \left(\frac{\partial g_{ortho,\phi}(\bm z_\theta)}{\partial z_{\theta_j}} \right) = \frac{1}{2 \lambda \sigma ^2} \delta_{ij}.
\end{eqnarray} 
\mblk
Here, the third term of the right side is constant, this term can be removed from the cost function as follows:
\mmag
\begin{eqnarray}
L'_{ortho} = - \log \left({P_{\vz, \psi}({\bm z_\theta})}\right)+ 
\lambda_1 h\left(D({\bm x},\hat {\bm x}) \right).
\end{eqnarray} 
\mblk
Then the parameters of network and PDF are obtained according to the following equation:
\begin{eqnarray}
\label{cost6}
\theta, \phi, \psi = \argmin_{\theta, \phi, \psi} (E_{{\bm x}\sim P_{\vx}({\bm x})}[L'_{ortho}]).
\end{eqnarray} 
$E_{{\bm x}\sim P_{\vx}({\bm x})}[L'_{ortho} ]$ in Eq.~(\ref{cost6}) can be transformed as the next:
\begin{align}
\label{cost7R}
E_{{\bm x}\sim P_{\vx}({\bm x})}[L'_{ortho}] 
&= \int  P_{\vx}(\bm x) \left( - \log \left({P_{\vz, \psi}({\bm z_\theta})}\right)+\lambda_1 \ h\left(D({\bm x},\hat {\bm x}) \right)\right){\mathrm{d}  \bm x} \nonumber \\
&= \int \left (P_{\vz, \theta} (\bm z_\theta) J_{sv}(\vz_{\theta})^{-1} \right)  \left( - \log \left({P_{\vz, \psi}({\bm z_\theta})}\right)\right) \ 
J_{sv}(\vz_{\theta}) \mathrm{d}z_\theta
 + \lambda_1 \int P_{\vx} (\bm x) \ h\left(D({\bm x},\hat {\bm x}) \right) \mathrm{d} \bm x .
\end{align}
%
At first, the first term of the third formula in Eq.(\ref{cost7R}) is examined. Let $\mathrm{d} \bm z_{\theta /i}$ be a differential of $(N-1)$ dimensional latent variables where i-th axis $z_{\theta i}$ is removed from the latent variable $\bm z_\theta$. Then a marginal distribution of $z_{\theta i}$ can be derived from the next equation:
\begin{eqnarray}
\label{MarginalZ}
P_{z, \theta i}(z_{\theta i})=\int P_{\vz, \theta}(\bm z_\theta) \mathrm{d} \bm z_{\theta /i}.
\end {eqnarray}
By using Eqs.(\ref{Fact1}) and (\ref{MarginalZ}), the first term of the third formula in Eq. (\ref{cost7R}) can be expanded as:
\begin{eqnarray}
\label {EntLatent}
\int P_{\vz, \theta} (\bm z_{\theta}) \left( - \log \left({P_{\vz, \psi}({\bm z_{\theta}})}\right)\right)\mathrm{d} \bm z_{\theta} 
&=& \int P_{\vz, \theta} (\bm z_{\theta}) \  \left( - \log \left(\frac{\prod_{i=1}^N P_{z_i, \psi}({z_{\theta i}})}{\prod_{i=1}^N P_{z, \theta i}({z_{\theta i}})}\right)\right)\mathrm{d} \bm z_{\theta} \nonumber \\
&\ & \ + 
\int P_{\vz, \theta} (\bm z_{\theta}) \  \left( - \log \left({\prod_{i=1}^N P_{z, \theta i}({z_{\theta i}})}\right)\right)\mathrm{d} \bm z_{\theta} \nonumber \\
&=& \sum_{i=1}^N \int \left (\int P_{\vz, \theta} (\bm z_{\theta}) \mathrm{d} \bm z_{\theta /i} \right ) \left( - \log \left(\frac{P_{z_i, \psi}({z_{\theta i}})}{P_{z, \theta i}({z_{\theta i}})}\right)\right)\mathrm{d} z_{\theta i} 
\nonumber \\
&\ & \ +
\sum_{i=1}^N \int \left (\int P_{\vz, \theta} (\bm z_{\theta}) \mathrm{d} \bm z_{\theta /i} \right ) \  \left( - \log \left(P_{z, \theta i}({z_{\theta i}})\right)\right)\mathrm{d} z_{\theta i} \nonumber \\
&=& \sum_{i=1}^N D_{KL}(P_{z, \theta i}({z_{\theta i}}) \| P_{z_i, \psi}({z_{\theta i}})) + \sum_{i=1}^N H(z_{\theta i}).
\end{eqnarray} 
%
Here $H(X)$ denotes the entropy of a variable $X$. The first term of the third formula is KL-divergence between marginal probability $P_{z, \theta i}({z_{\theta i}})$ and factorized parametric probability $P_{z_i, \psi}({z_{\theta i}})$. 
The second term of the third formula can be further transformed using mutual information between latent variables $I(\bm z_\theta)$ and equation (\ref{Prop21}). 
\begin{align}
\sum_{i=1}^N H(z_{\theta i}) &= H(\bm z_\theta) + I(\bm z_\theta) \simeq -\int J_{sv}(\vz_{\theta}) P_{\vx}(\vx) \log (J_{sv}(\vz_{\theta}) P_{\vx}(\vx)) J_{sv}(\vz_{\theta})^{-1} \mathrm{d} \vx  + I(\bm z_\theta) \nonumber \\
&= H(\bm x) - \int P_{\vx}(\vx) \log \left( \left( \frac{1}{2 \lambda_{2} \sigma^{2}} \right)^{\frac{N}{2}} \biggl( \prod_{j=1}^{N} \alpha_{j}(\mA(\vx)) \biggr)^{-\frac{1}{2}} \right) \mathrm{d} \vx  + I(\bm z_\theta)
\label{ent}
\end{align}
At second, the second term of the third formula in Eq.~(\ref{cost7R}) is examined. 
When $\bm x$ and $\hat {\bm x}$ are close, the following equation holds. 
\mmag
\begin{eqnarray}
%
D(\bm x, \hat{\bm x}) \simeq (\vx-\hat{\vx})^\top  \mA(\vx) (\vx-\hat{\vx}). 
\end{eqnarray}
\mblk
Note that with given distribution $\vx \sim P_{\vx}({\bm x})$, 
the first and the second term in the right side of Eq.~(\ref{ent}) are fixed value. 
%
Therefore, by using these expansions, Eq.(\ref{cost7R}) can be expressed as:
\begin{eqnarray}
\label{cost8R}
E_{{\bm x}\sim P_{\vx}({\bm x})}[L'_{ortho}] & \simeq &
\sum_{i=1}^N D_{KL}(P_{z, \theta i}({z_{\theta i}}) \| P_{z_i, \psi}({z_{\theta i}})) \nonumber \\
&\ & +I(\bm z _\theta)
+E_{\bm x} \left[ (\vx-\hat{\vx})^\top  \mA(\vx) (\vx-\hat{\vx})  \right]
+ \mathrm{Const.}
\end{eqnarray} 
\mmmblu
Here, the real space $\sR^{M}$ is divided into a plurality of small subspace partitioning $\Omega \bm x_{1},\ \Omega \bm x_{2}, \ \cdots$.  
Note that $\sR^{M}$ is an inner product space endowed with metric tensor $\mA(\vx)$. 
Let $\Omega \bm z_1,\ \Omega \bm z_2, \ \cdots$ be the division space of the latent space ${\bm z} \in {\bm R}^N$ corresponding to $\Omega \bm x$. 

Then Eq. (\ref{cost8R}) can be rewritten as: 
\begin{eqnarray}
\label{DivSpace}
E_{{\bm x}\sim P_{\vx}({\bm x})}[L'_{ortho}] & \simeq & \sum_{i=1}^N D_{KL}(P_{z, \theta i}({z_{\theta i}}) \| P_{z_i, \psi}({z_{\theta i}})) \nonumber \\
&\ & + 
\sum_{k} \left ( I(\bm z _\theta \in \Omega \bm z_{\theta k})
+E_{\bm x \in \Omega \bm x_{k}} \left[ (\vx-\hat{\vx})^\top  \mA(\vx) (\vx-\hat{\vx}) \right] \right ) + \mathrm{Const.}
\end{eqnarray} 
%
For each subspace partitioning, Jacobian matrix for the transformation from $\Omega \bm x_{k}$ to $\Omega \bm z_{\theta k}$ forms constantly scaled orthonormal system with respect to $\mA(\vx)$.  
\mblk
According to Karhunen-Lo\`eve Theory \citep{KLTBook}, the orthonormal basis which minimize both mutual information and reconstruction error leads to be Karhunen-Lo\`eve transform (KLT). It is noted that the basis of KLT is equivalent to PCA orthonormal basis. 

As a result, when Eq. (\ref{DivSpace}) is minimized, Jacobi matrix from $\Omega \bm x_{k}$ to $\Omega \bm z_{\theta k}$ for each subspace partitioning should be KLT/PCA. 
Consequently, the same feature as PCA will be realized such as the determination of principal components etc. 

From these considerations, we conclude that RaDOGAGA has a ``continuous PCA" feature. 
This is experimentally shown in Section \ref{exp_iso} and Appendix \ref{ap_exp_pca}. 
\begin{figure}[h]
	\centering
	\includegraphics[clip, width=12cm]{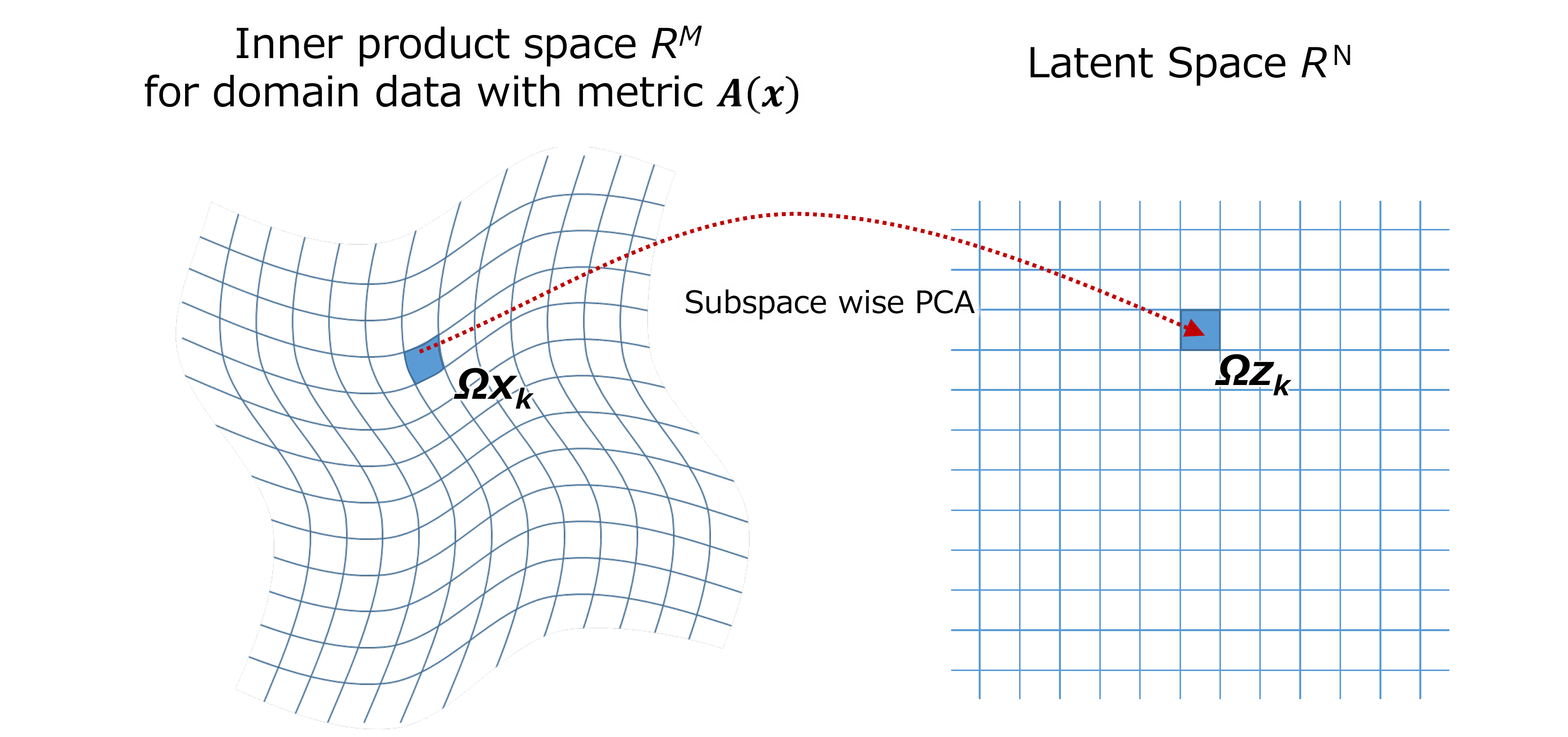}
	\caption{Continuous KLT (PCA) Mapping from input domain to latent space. For all small subspace partitioning $\Omega \bm x_{k}$ domain space (which is inner product space with metric tensor $\mA(\vx)$), mapping from $\Omega \bm x_{k}$ to $\Omega \bm z_{\theta k}$ can be regarded as  PCA
}
	\label{figKLT2}
\end{figure}
\section{Detail and Expansion Result of Experiment in Section \ref{exp_iso}}
\label{app_iso}
In this section, we will provide further detail and a result of the complemental experiment regarding section \ref{exp_iso}. 
\subsection{GDN Activation}
GDN activation function~\citep{GDN} 
is known to suitable for image compression. 
For implementation, we use a TensorFlow library\footnote[6]{ \url{https://github.com/tensorflow/compression/tree/master/docs/api\_ docs/python/tfc}}. 
\subsection{Other Training Information}
The batch size is 64. The iteration number is 500,000. We use NVIDIA Tesla V100 (SXM2).\\
For RaDOGAGA, since our implementation was done based on the source code for image compression, entropy rate is calculated as $-\log((P_{\vz,\psi}(\vz))/(M\log 2)$, meaning bit per pixel. 
In addition, for RaDOGAGA, the second term of Eq.~(7) is always MSE in this experiment. This is because we found that the training with $1-SSIM$ as the reconstruction loss is likely to diverge at the beginning step of the training. Therefore, we tried to start training with MSE and then fine-tuned with $1-SSIM$.  Eventually, the result is almost the same as the case without finetuning. Therefore, to simplify the training process, we do not usually finetune. 
%
\subsection{Generation of $\vv_{z}$ and $\vw_{z}$}
To evaluate the isometricity of the mapping, it is necessary to prepare random tangent vector $\vv_{z}$ and $\vw_{z}$ with a scattered interior angle. 
We generate two different tangent vectors $\vv_{z}=\{ v_{z1}, v_{z2}, \ldots, v_{zn} \}$ and $\vw_{z}=\{ w_{z1}, w_{z2}, \ldots, w_{zn} \}$ in the following manner. 
First, we prepare $\vv'_{z} \in \sR^{N}$ as $\{1.0, 0.0,\ldots 0.0 \}$. 
Then, we sample $\bm{\alpha} = \{ \alpha_{1}, \alpha_{2}, \ldots, \alpha_{n-1} \}$ $\left(\alpha_{1 \cdots n-2} \sim U \left(0, \pi \right), \alpha_{n-1} \sim U \left(0, 2\pi \right) \right)$ to set $\vw'$ as the conversion of polar coordinate $\{r, \bm{\alpha} \} \in \sR^{N}$ to rectangular coordinates, where $r=1$. 
Thus, the distribution of interior angle of $\vv'_{z}$ and $\vw'_{z}$ also obey $\alpha_{1} \sim U \left(0, \pi \right)$. 
Next, we randomly rotate the plane $\sR^{N}$ in which interior angle of $\vv'_{z}$ and $\vw'_{z}$ is $\alpha_{1}$ 
in the following way~\cite{Rotation}
and obtain $\vv_{z}$ and $\vw_{z}$.
\begin{align}        
\bm{\rho} = - \frac{\cos \alpha_{1}} {\sin \alpha_{1}}\vv'_{z} + \frac{1}{\sin \alpha_{1}}\vw'_{z}, \ \ \ \ \  
\bm{\tau} = \vv'_{z}  \nonumber, 
\end{align}
then, 
\begin{align}        
\left(\begin{array}{c}\vv_{z}\\ \vw_{z}\end{array}\right)=
\begin{bmatrix}-\sin\omega & \cos\omega \\ \cos\omega\sin\alpha_{1}-\sin\omega\cos\alpha_{1} &  \sin\omega\sin\alpha_{1}+\cos\omega\cos\alpha_{1} \end{bmatrix}
\left(\begin{array}{c}\bm{\rho}\\ \bm{\tau}\end{array}\right),
\end{align}
where $\omega \sim U \left(0, 2\pi \right)$ is the rotation angle of the plane. 
Note that since this is the rotation of the plane, the interior angle between $\vv$ and $\vw$ is kept to $\alpha_{1}$. 
Finally, we normalize the norm of  $\vv_{z}$ and $\vv_{z}$ to be 0.01. 

\subsection{Experiment with MNIST Dataset and BCE}
Besides of the experiment in main paper, we conducted an experiment with MNIST dataset~\citep{lecun1998gradient}\footnote[8]{\url{http://yann.lecun.com/exdb/mnist/}} which contains handwritten digits binary images with the image size of $28 \times 28$. 
We use 60,000 samples in the training split. 
The metric function is $BCE$, where $\mA(\vx)$ is approximated as Eq. (\ref{ax_bce}). 
Autoencoder consists of  FC layers with sizes of 1000, 1000, 128, 1000, and 1000. 
We attach $softplus$ as activation function except for the last of the encoder and the decoder. 
In this experiment, we modify the form of the cost function of beta-VAE as
\begin{eqnarray}
\label{cost_VAE2}
L = -L_{kl} + \lambda_1 h \left(D \left({\bm x},{\hat {\bm x}}\right)\right) + \lambda_2  D \left({\hat {\bm x}}, {\breve {\bm x}}\right),
\end{eqnarray} 
where $\hat{\vx}$ is the output of the decoder without noise, and $\breve {\vx}$ is the output of the decoder with the noise of reparameterization trick. 
We set $(\lambda_{1}, \lambda_{2})$ as (10, 1) for beta-VAE and (0.01, 0.01) for RaDOGAGA. 
Optimization is done with Adam optimizer with learning late $1\times10^{-4}$ for beta-VAE $1\times10^{-5}$ for RaDOGAGA. 
The batch size is 256 and the training iteration is 30,000. 
These parameters are determined to make the $PSNR=20\log_{10}\left( \frac{MAX_{\vx}^2}{MSE} \right)$, where $MAX_{\vx}=255$, between input and reconstruction image approximately 25 dB. 

\begin{figure}[h]
 \begin{minipage}[b]{0.49\linewidth}
  \centering
  \includegraphics[keepaspectratio, scale=0.20]
  {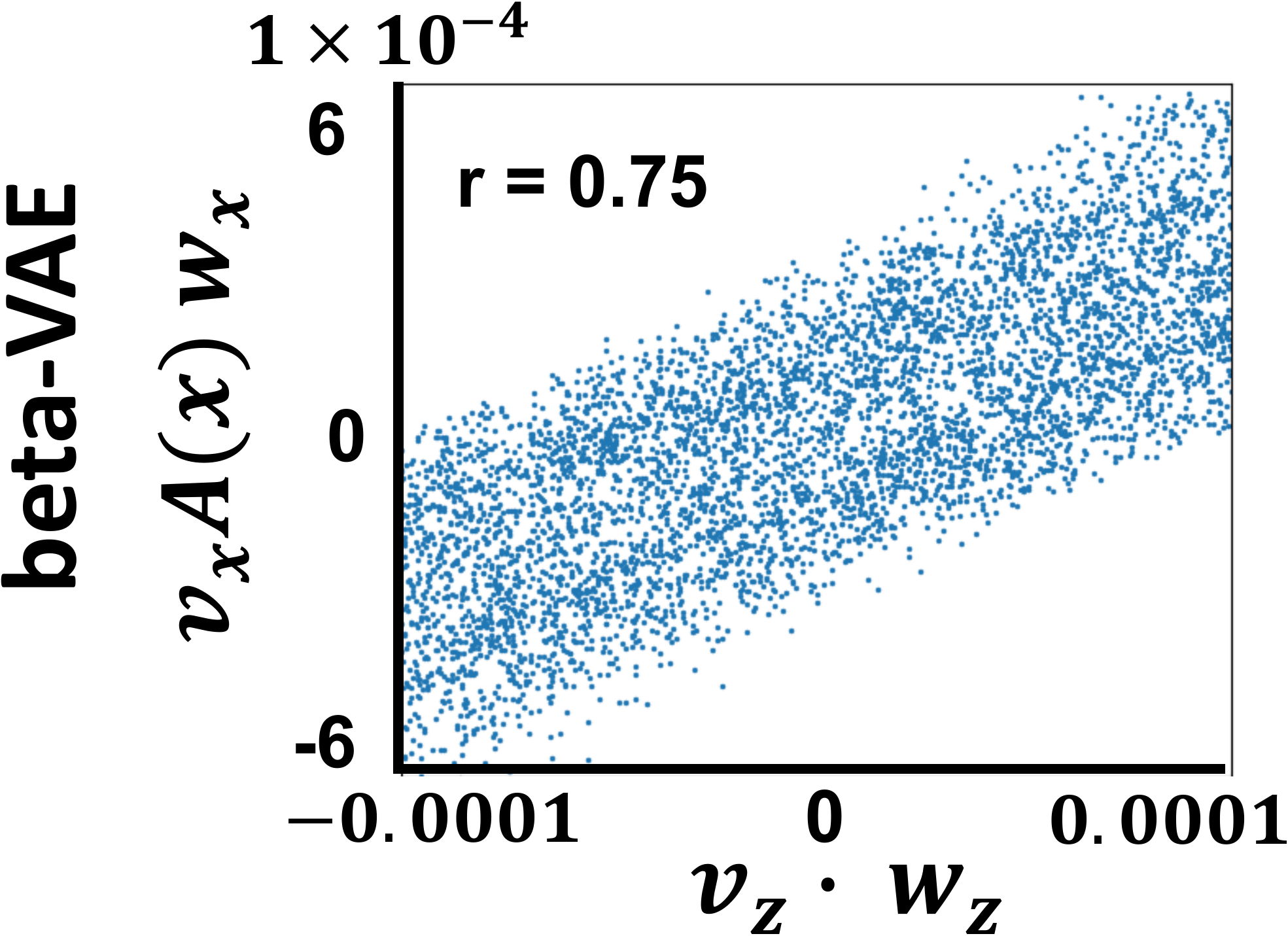}
  \subcaption{beta-VAE}
 \end{minipage}
 \begin{minipage}[b]{0.49\linewidth}
  \centering
  \includegraphics[keepaspectratio, scale=0.20]
  {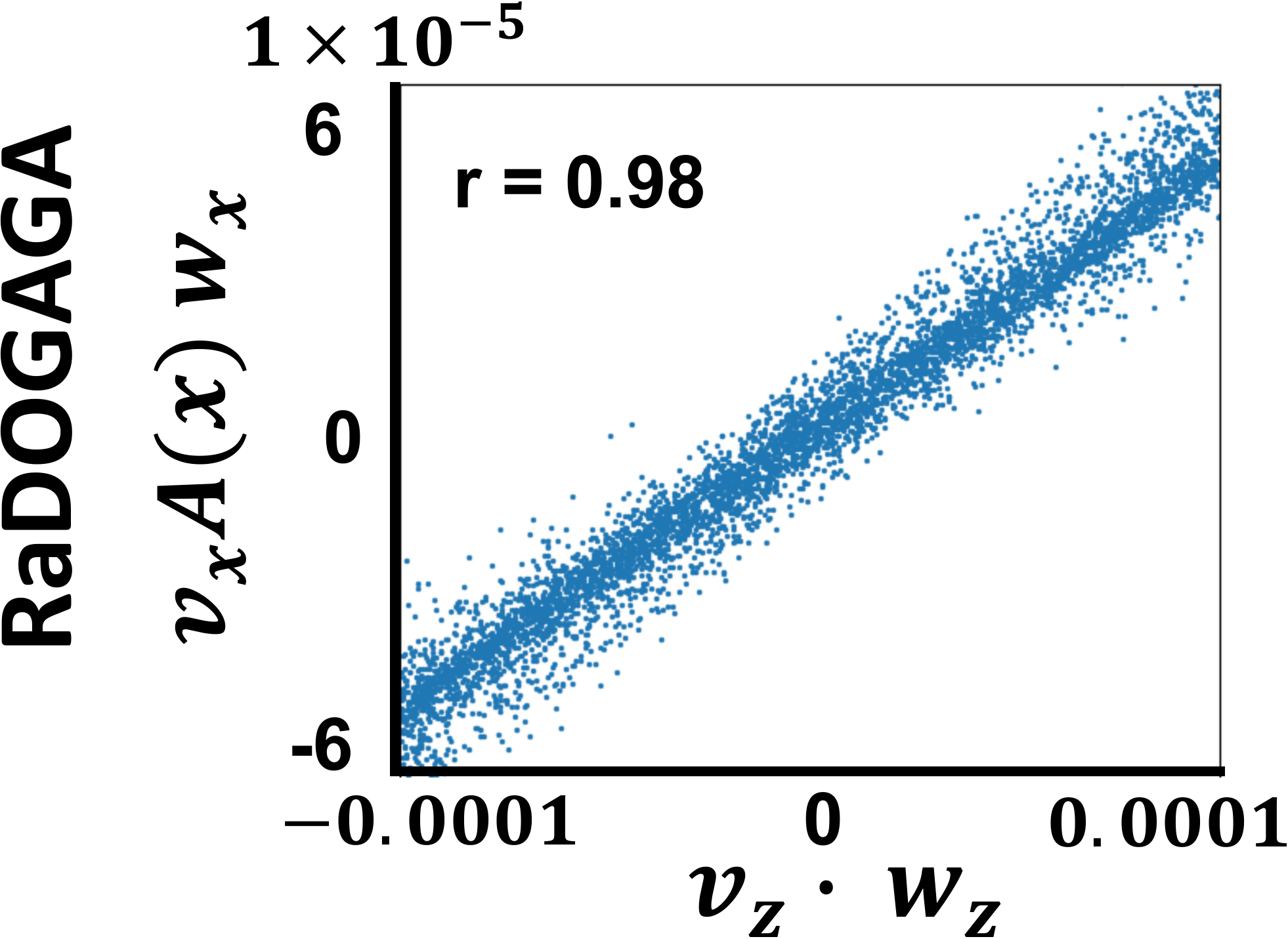}
  \subcaption{RaDOGAGA}
 \end{minipage}
 \caption{Plot of $\vv_z \cdot \vw_z$ (horizontal axis) and ${\vv_x}^\top \bm{A}(\vx) \vw_x$ (vertical axis). In beta-VAE (left), the correlation is week while in our method (right) we can observe proportionality. }
\label{iso_bce}
\end{figure}

Figure \ref{iso_bce} depicts the result. We can observe that map of RaDOGAGA is isometric as well even for the case the metric function is $BCE$. 
Consequently, even if the metric function is complicated one, the impact of the latent variable on the metric function is tractable. 
We expect this feature promotes further improving of metric learning, data interpolation, and so on. 
%

\subsection{Isometricity of Encoder Side}\label{ap_exp_enc}
\begin{figure}[h]
 \begin{minipage}[b]{0.50\linewidth}
  \centering
  \includegraphics[keepaspectratio, scale=0.20]
  {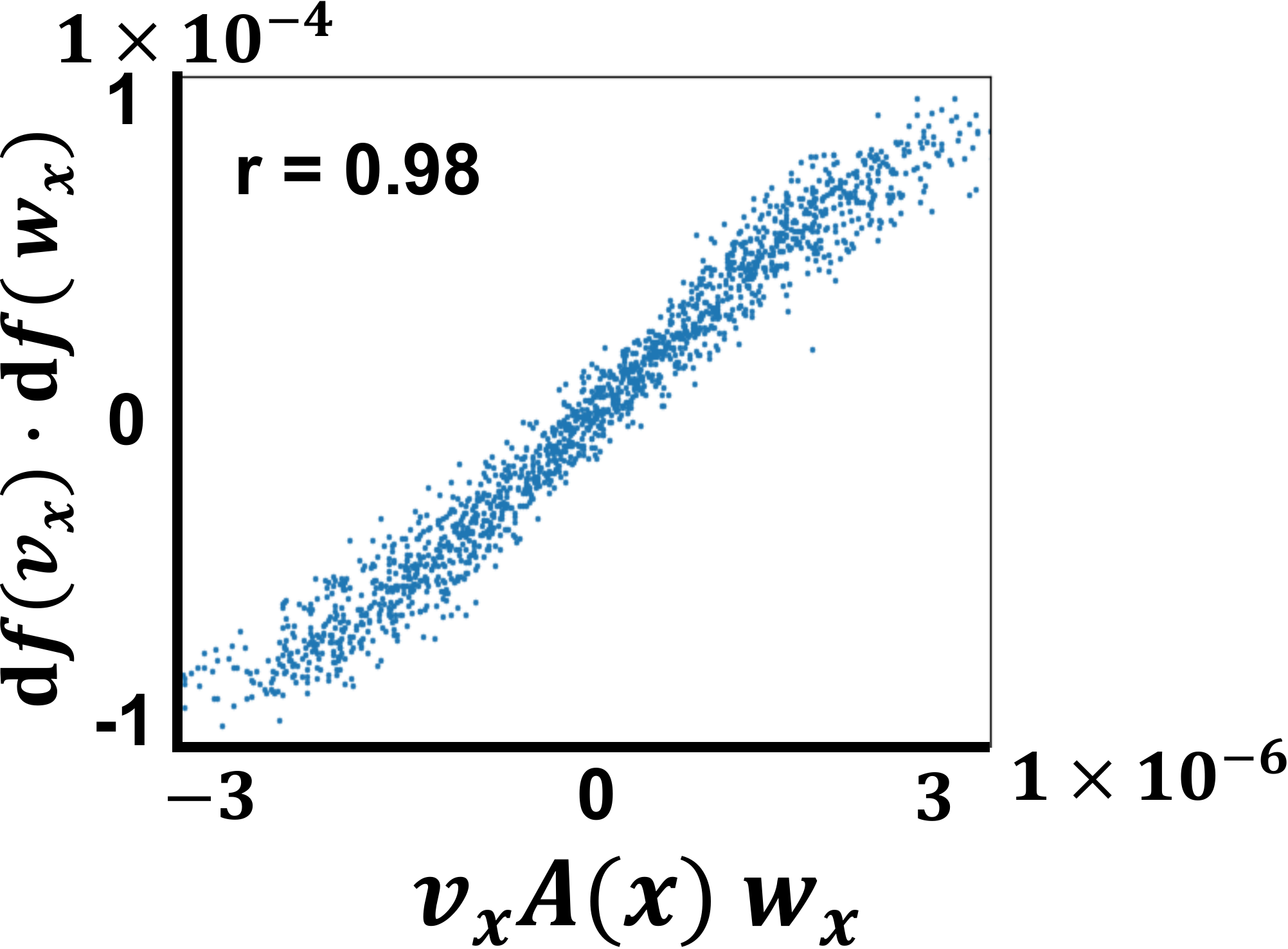}
  \subcaption{Variance of $\vz$}
 \end{minipage}
 \begin{minipage}[b]{0.50\linewidth}
  \centering
  \includegraphics[keepaspectratio, scale=0.20]
  {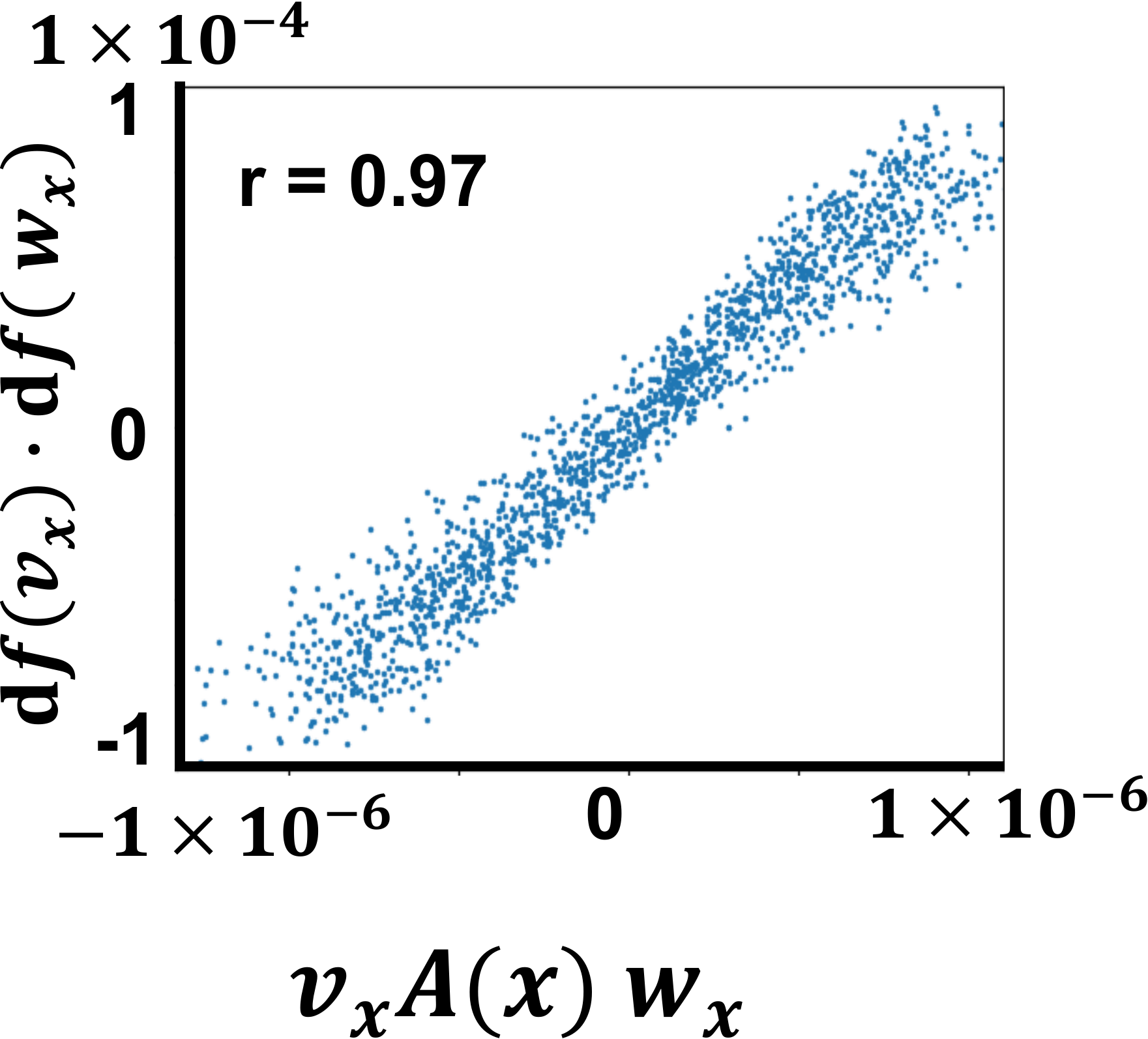}
  \subcaption{}
 \end{minipage}
 \caption{ ${\vv_x}^\top \bm{A}(\vx) \vw_x$ vs $\dif f(\vv_x) \cdot \dif f(\vw_x)$.
 The mapping by encoder is also isometric. }
\label{fig:emb}
\end{figure}

In Section \ref{exp_iso}, we showed the isometricty of decoder side because it is common to analyse the behavior of latent variables by observing the decoder output such as latent traverse. 
We also clarify that the embedding by encoder $f$ keep isometric. Given two tangent vector $\vv_{x}$ and $\vw_{x}$, ${\vv_{x}}^\top \mA(\vx) \vw_{x}$ is compared to $\dif f(\vv_{x}) \cdot \dif f(\vv_{x})$.  $\dif f(\vw_{x})$ is also approximated by $f(\vx + \vv_x) - f(\vx)$. As Fig.~\ref{fig:emb} shows, the embedding to the latent space is isometric. Consequently, it is experimentally supported that our method enables to embed data in Euclidean space isometrically. The result of the same experiment for the case of the metric is $1-SSIM$ is provided in Appendix \ref{app_iso}. 

\subsection{Additional Latent Traverse}\label{ap_exp_pca}
\begin{figure}[h]
 \begin{minipage}[b]{0.50\linewidth}
  \centering
  \includegraphics[keepaspectratio, scale=0.30]
  {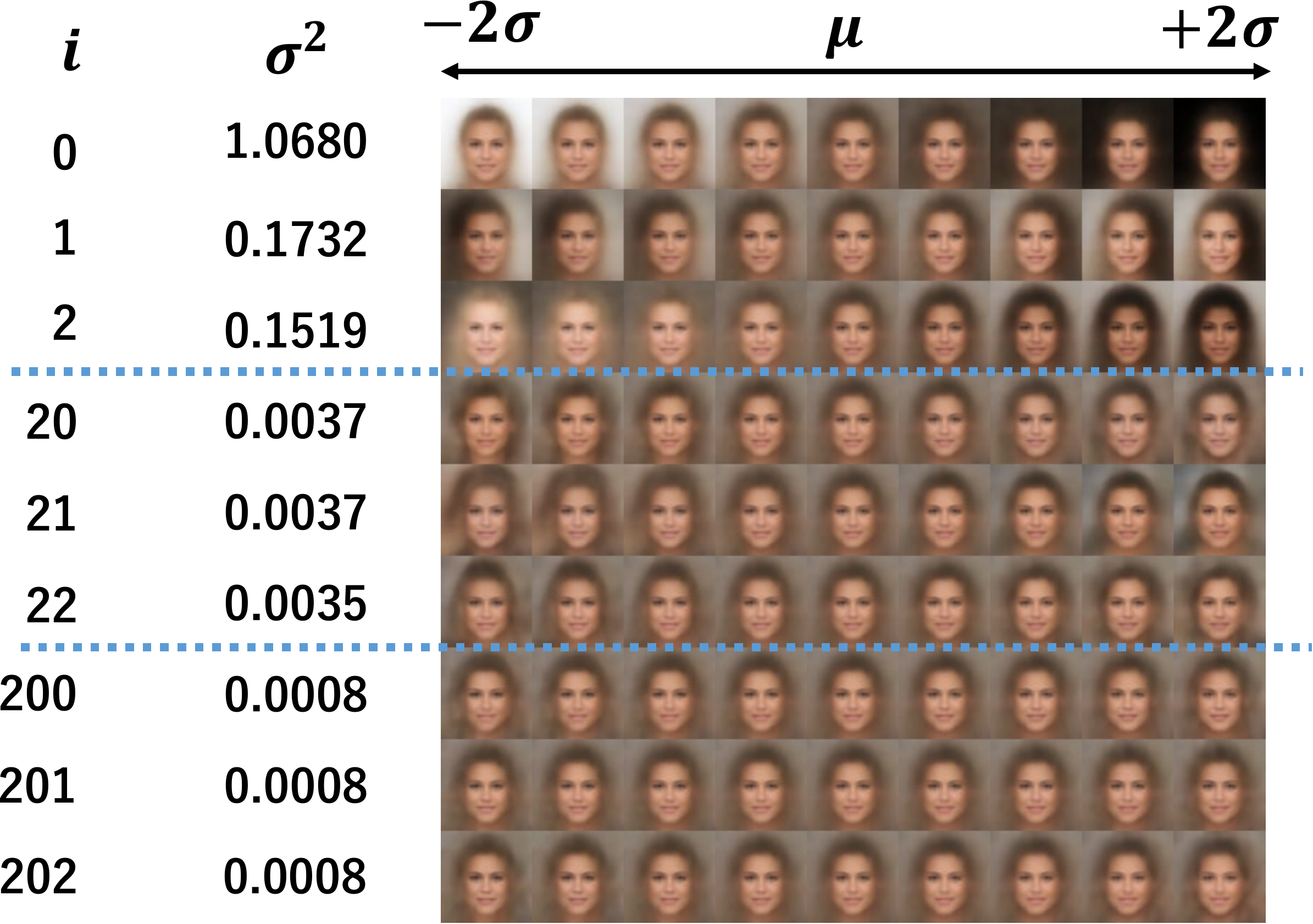}
  \subcaption{$MSE$}
	\label{fig:trv_mse}
 \end{minipage}
 \begin{minipage}[b]{0.50\linewidth}
  \centering
  \includegraphics[keepaspectratio, scale=0.30]
  {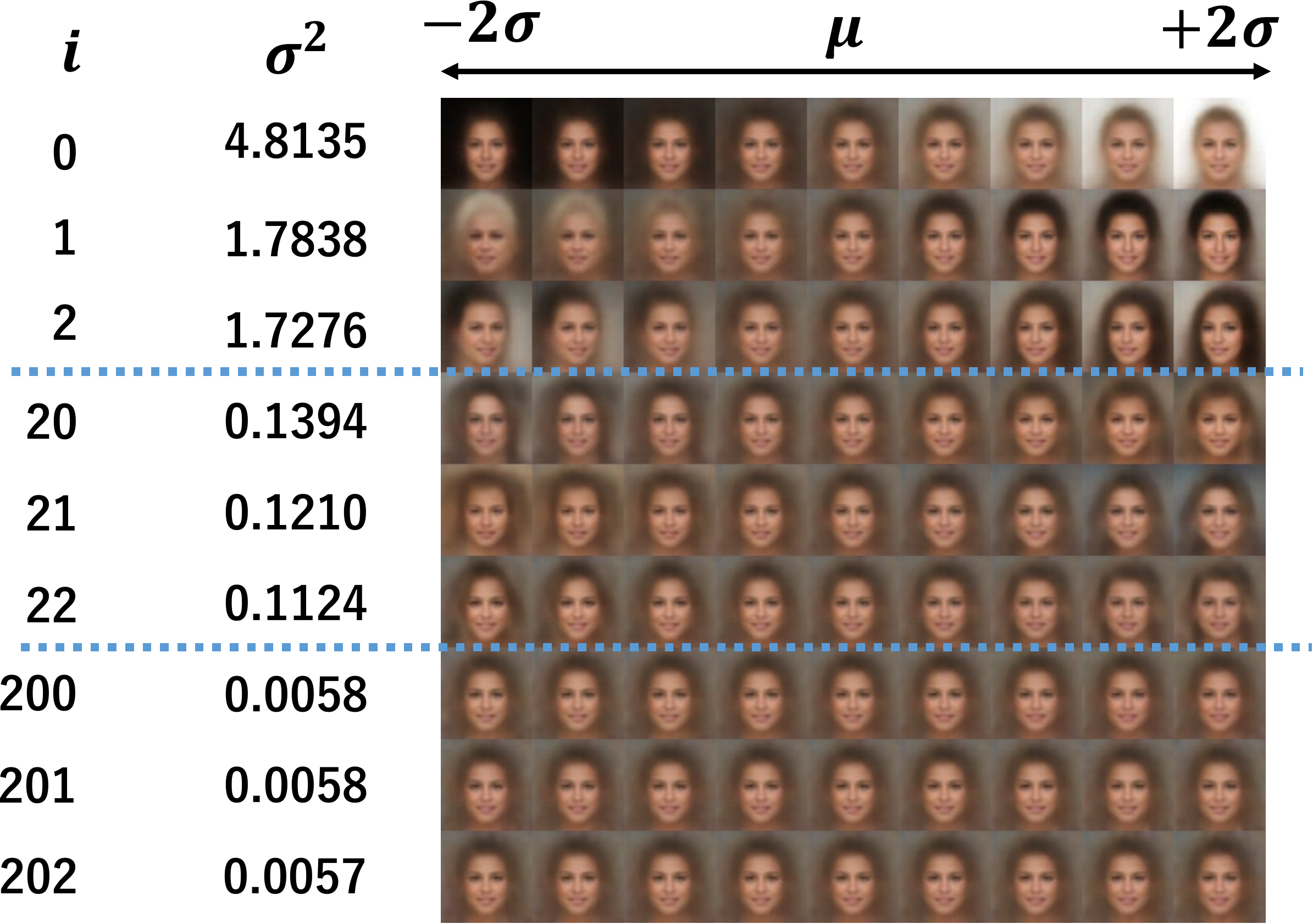}
  \subcaption{$1-SSIM$}
	\label{fig:trv_ssim}
 \end{minipage}
 \caption{Latent space traversal of z. For the top-3 variables, pictures look significantly different. 
In the middle range($z_{20}$, $z_{21}$, $z_{22}$ ), the difference is smaller than the upper three but still observable. 
For the bottom three, there is almost no difference.
}\label{fig:trv}
\end{figure}

In Section \ref{exp_iso}, the latent traverse for variables with the top 9 variances was provided. 
To further clarify whether the variance is corresponding to visual impact, the latent traverse of RaDOGAGA for $z_{0}$, $z_{1}$, $z_{2}$, $z_{20}$, $z_{21}$, $z_{22}$, $z_{200}$, $z_{201}$, and $z_{202}$ are shown in Fig.~\ref{fig:trv}. 
Apparently, a latent traverse with a larger $\sigma$ makes a bigger difference in the image. When the $\sigma^2$ gets close to 0, there is almost no visual difference. 
Accordingly, the behavior as continuous PCA is clarified throughout the entire variables. 
\section{Detail of the Experiment in Section \ref{exp_ano}}
\label{app_ano}
In this section, we provide further detail of experiment in Section \ref{exp_ano}.
\subsection{Datasets}
We describe the detail of following four public datasets:

\textbf{KDDCUP99 \citep{UCI2019}} The KDDCUP99 10 percent dataset from the UCI repository is a dataset for cyber-attack detection.
This dataset consists of 494,021 instances and contains 34 continuous features and 7 categorical ones. We use one hot representation to encode the categorical features, and eventually obtain a dataset with features of 121 dimensions.
Since the dataset contains only 20\% of instances labeled -normal- and the rest labeled as -attacks-, -normal- instances are used as anomalies, since they are in a minority group. 

\textbf{Thyroid \citep{UCI2019}} This dataset contains 3,772 data sample with 6-dimensional feature from patients and can be divided in three classes: normal (not hypothyroid), hyperfunction, and subnormal functioning. We treat the hyperfunction class (2.5\%) as an anomaly and rest two classes as normal.

\textbf{Arrhythmia \citep{UCI2019}} This is dataset to detect cardiac arrhythmia containing 452 data sample with 274-dimensional feature. We treat minor classes (3, 4, 5, 7, 8, 9, 14, and 15, accounting for 15\% of the total) as anomalies, and the others are treated as normal. 

\textbf{KDDCUP-Rev \citep{UCI2019}} To treat “normal” instances as majority in the KDDCUP dataset, we keep all “normal” instances and randomly pick up “attack” instances so that they compose 20\% of the dataset. In the end, the number of instance is 121,597. 

Data is max-min normalized toward dimension through the entire dataset.
\subsection{Hyperparameter and Training Detail}
Hyperparameter for RaDOGAGA is described in Table \ref{tab:hyper1}. First and second column is number of neurons. 
($\lambda_{1}$,  $\lambda_{2}$) is determined experimentally. 
For DAGMM, the number of neuron is the same as Table \ref{tab:hyper1}. 
We set ($\lambda_{1}$,  $\lambda_{2}$) as (0.1, 0.005) referring \citet{DAGMM} except for Thyroid. 
Only for Thyroid, ($\lambda_{1}$,  $\lambda_{2}$) is (0.1, 0.0001) since (0.1, 0.005) does not work well with our implementation.  
Optimization is done by Adam optimizer with learning rate $1\times10^{-4}$ for all dataset. The batch size is 1024 for all dataset. The epoch number is 100, 20000, 10000, and 100 respectively. We save and the test models by every 1/10 epochs and early stop is applied. For this experiment, we use GeForce GTX 1080. 
\begin{table*}[h]
\caption{Hyper parameter for RaDOGAGA}
\label{tab:hyper1}
\begin{center}
\begin{tabular}{l|l|l|l|l|l|l}
Dataset      & Autoencoder      & EN &$\lambda_{1}(d)$ & $\lambda_{2}(d)$ & $\lambda_{1}((log(d))$ & $\lambda_{2}(log(d))$   \\ \hline
KDDCup99   & 60, 30, 8, 30, 60 & 10, 4 &100 & 1000 &10 & 100\\ \hline
Thyroid    & 30, 24, 6, 24, 30 & 10, 2 &10000 & 1000 &100 & 1000\\ \hline
Arrhythmia & 10, 4, 10         & 10, 2&1000 & 1000 &1000 & 100 \\ \hline
KDDCup-rev & 60, 30, 8, 30, 60 & 10, 2 &100 & 100 &100 & 100\\ \hline
\end{tabular}
\end{center}
\end{table*}
\subsection{Experiment with different network size}
In addition to experiment in main page, we also conducted experiment with same network size as in \citet{DAGMM} with parameters in Table \ref{hyper2}
\begin{table*}[h]
\caption{Hyper parameter for RaDOGAGA(same network size as in \citet{DAGMM}}
\label{hyper2}
\begin{center}
\begin{tabular}{l|l|l|l|l|l|l}
Dataset      & Autoencoder      & EN &$\lambda_{1}(d)$ & $\lambda_{2}(d)$ & $\lambda_{1}((log(d))$ & $\lambda_{2}(log(d))$   \\ \hline
KDDCup99   & 60, 30, 1, 30, 60 & 10, 4  &100 & 100 &100 & 1000\\ \hline
Thyroid    & 12, 4, 1, 4, 12 & 10, 2 & 1000 &10000 & 100 &10000 \\ \hline
Arrhythmia & 10, 2, 10         & 10, 2 &1000 & 100 &1000& 100 \\ \hline
KDDCup-rev & 60, 30, 1, 30, 60 & 10, 2 &100 & 100 &100& 1000\\ \hline
\end{tabular}
\end{center}
\end{table*}

Now, we provide results of setting in Table \ref{hyper2}. 
In Table \ref{tab:anomaly_2}, RaDOGAGA- and DAGMM- are results of them and DAGMM is result cited from \citet{DAGMM}.
Even with this network size, our method has boost from baseline in all dataset. 
\begin{table*}[h]
\renewcommand{\footnoterule}{\empty}
\caption{Average and standard deviations (in brackets) of Precision, Recall and F1}
\label{tab:anomaly_2} 
\begin{minipage}{\textwidth}
\begin{center}
\begin{tabular}{c|l|lll}
\multicolumn{1}{l|}{Dataset} & Methods       & Precision      & Recall         & F1             \\ \hline
\multirow{4}{*}{KDDCup} 
                             & DAGMM         & 0.9297         & 0.9442         & 0.9369         \\
                             & DAGMM-        & 0.9338 (0.0051) & 0.9484 (0.0052) & 0.9410 (0.0051) \\
                             & RaDOGAGA-(L2)     & 0.9455 (0.0016) & 0.9608 (0.0018) & 0.9531 (0.0017) \\
                             & RaDOGAGA-(log)    & 0.9370 (0.0024) & 0.9517 (0.0025) & 0.9443 (0.0024) \\ \hline
\multirow{4}{*}{Thyroid}     
                             & DAGMM         & 0.4766         & 0.4834         & 0.4782         \\
                             & DAGMM-       & 0.4635 (0.1054) & 0.4837 (0.1100) & 0.4734 (0.1076) \\
                             & RaDOGAGA-(L2)     & 0.5729 (0.0449) & 0.5978 (0.0469) & 0.5851 (0.0459) \\
                             & RaDOGAGA-(log)    & 0.5729 (0.0398) & 0.5978 (0.0415) & 0.5851 (0.0406) \\ \hline
\multirow{4}{*}{Arrythmia}   
                             & DAGMM         & 0.4909         & 0.5078         & 0.4983         \\
                             & DAGMM-        & 0.4721 (0.0451) & 0.4864 (0.0464) & 0.4791 (0.0457) \\
                             & RaDOGAGA-(L2)     & 0.4897 (0.0477) & 0.5045 (0.0491) & 0.4970 (0.0484) \\
                             & RaDOGAGA-(log)    & 0.5044 (0.0364) & 0.5197 (0.0375) & 0.5119 (0.0369) \\ \hline
\multirow{4}{*}{KDDCup-rev}  
									& DAGMM         & 0.937          & 0.939          & 0.938          \\
                             & DAGMM-        & 0.9491 (0.0163) & 0.9498 (0.0158) & 0.9494 (0.0160) \\
                             & RaDOGAGA-(L2)     & 0.9761 (0.0057) & 0.9761 (0.0056) & 0.9761 (0.0057) \\
                             & RaDOGAGA-(log)    & 0.9791 (0.0036) & 0.9799 (0.0035) & 0.9795 (0.0036) \\
\end{tabular}
\end{center}
\end{minipage}
\end{table*}
\clearpage
\end{document}